\documentclass[10pt,twocolumn,letterpaper]{article}

\usepackage{cvpr}              % To produce the CAMERA-READY version
\usepackage[table]{xcolor}
\definecolor{oursblue}{RGB}{232,243,255}

\definecolor{cvprblue}{rgb}{0.21,0.49,0.74}
\usepackage[pagebackref,breaklinks,colorlinks,allcolors=cvprblue]{hyperref}
\usepackage{adjustbox}
\usepackage{rotating}
\usepackage{placeins}
\usepackage{float}
\usepackage{graphicx}
\usepackage{xcolor}
\usepackage{tikz}
\usepackage{caption}
\usepackage{capt-of}
\usepackage{arydshln}

\newcommand{\mocapgridimage}[1]{%
    \begin{minipage}{0.18\linewidth}
        \centering
        \includegraphics[
            width=0.98\linewidth
        ]{figures/qualitative_mocap_main_overlay_v2/#1.pdf}%
    \end{minipage}%
}
\newcommand{\mocapgridrow}[2]{%
    \noindent
    \begin{minipage}{0.05\linewidth}
        \begin{sideways}\scriptsize\textbf{\shortstack{#1}}\end{sideways}
    \end{minipage}%
    \mocapgridimage{#2_000035}%
    \mocapgridimage{#2_000063}%
    \mocapgridimage{#2_000115}%
    \mocapgridimage{#2_000122}%
    \mocapgridimage{#2_000179}%
    \par
}
\newcommand{\dexycbgridimage}[1]{%
    \begin{minipage}{0.18\linewidth}
        \centering
        \includegraphics[
            width=0.92\linewidth
        ]{figures/supplementary_materials/DexYCB_20201002-subject-08_20201002_112114_836212060125/#1.pdf}%
    \end{minipage}%
}
\newcommand{\dexycbgridrow}[2]{%
    \noindent
    \begin{minipage}{0.05\linewidth}
        \begin{sideways}\scriptsize\textbf{\shortstack{#1}}\end{sideways}
    \end{minipage}%
    \dexycbgridimage{#2_000030}%
    \dexycbgridimage{#2_000042}%
    \dexycbgridimage{#2_000047}%
    \dexycbgridimage{#2_000054}%
    \dexycbgridimage{#2_000067}%
    \par
}
\newcommand{\hogridimage}[1]{%
    \begin{minipage}{0.18\linewidth}
        \centering
        \includegraphics[
            width=0.98\linewidth
        ]{figures/supplementary_materials/HO3D_AP10/#1.pdf}%
    \end{minipage}%
}
\newcommand{\hogridrow}[2]{%
    \noindent
    \begin{minipage}{0.05\linewidth}
        \begin{sideways}\scriptsize\textbf{\shortstack{#1}}\end{sideways}
    \end{minipage}%
    \hogridimage{#2_000015}%
    \hogridimage{#2_000101}%
    \hogridimage{#2_000135}%
    \hogridimage{#2_000337}%
    \hogridimage{#2_000480}%
    \par
}
\newcommand{\randomgridimage}[1]{%
    \begin{minipage}{0.18\linewidth}
        \centering
        \includegraphics[
            width=0.98\linewidth
        ]{figures/supplementary_materials/random/#1.pdf}%
    \end{minipage}%
}
\newcommand{\randomgridrow}[2]{%
    \noindent
    \begin{minipage}{0.05\linewidth}
        \begin{sideways}\scriptsize\textbf{\shortstack{#1}}\end{sideways}
    \end{minipage}%
    \randomgridimage{#2_000060}%
    \randomgridimage{#2_000011}%
    \randomgridimage{#2_000073}%
    \randomgridimage{#2_000064}%
    \randomgridimage{#2_000131}%
    \par
}


\def\confName{CVPR}
\def\confYear{2027}

\title{Tool Pose Estimation and Tracking Needs a Helping Hand}

\author{Kuan Wei Huang\\
Cornell University\\
{\tt\small kwhuang@cs.cornell.edu}
\and
Bharath Hariharan\\
Cornell University\\
{\tt\small bharathh@cs.cornell.edu}
}

\begin{document}
% \maketitle
\raggedbottom

% \begin{strip}
%   \centering
%   % \includegraphics[width=\textwidth]{figures/teaser_v1.pdf}
%   % \par\vspace{3em}
%   % \includegraphics[width=\textwidth]{figures/teaser_v2.pdf}
%   % \par\vspace{3em}
%   \includegraphics[width=\textwidth]{figures/teaser_v4.pdf}
%   \captionof{figure}{Given an in-the-wild egocentric video (a), our approach (c) leverages the orange-outlined hand mask to outperform state-of-the-art method~\cite{foundationposewen2024} (d) in object pose estimation. (b-d) shows the pose and rendered mesh outline for the ground-truth, our method, and a state-of-the-art method, respectively. }
% \end{strip}
\twocolumn[{%
    \renewcommand{\twocolumn}[1][]{#1}%
    \maketitle
    {%
        \centering
        \includegraphics[width=\textwidth]{figures/teaser.pdf}
        \captionof{figure}{For a frame in an egocentric video (a), our approach (c) leverages the hand mask (outlined in orange) to achieve more accurate object pose estimation than prior work~\cite{foundationposewen2024} (d). Panels (b--d) show the poses and corresponding rendered mesh outlines for the ground-truth, our method and the prior art, respectively.}
        \label{fig:teaser}%
    }%
    \vspace{15pt}
}]

\begin{figure*}[!t]
    \centering
    \includegraphics[
        width=0.90\textwidth,
        trim=0 25 370 23,
        clip
    ]{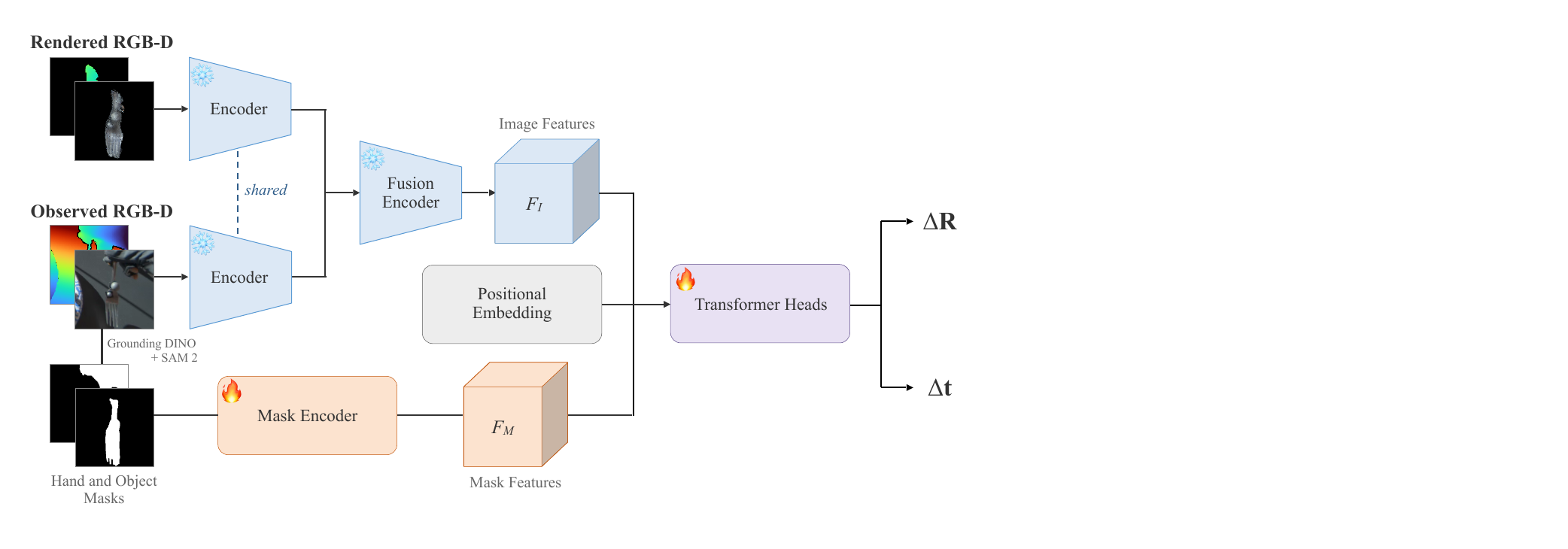}
    \caption{\textbf{Mask-augmented refinement network architecture.} Building on FoundationPose~\cite{foundationposewen2024}, our refinement network encodes the rendered and observed RGB-D crops with frozen encoders. Hand and object masks are encoded by a lightweight trainable convolutional branch and added with the fused RGB-D features. The network then adds a positional embedding to the resulting feature map and uses two transformer heads to predict rotation and translation updates, $\Delta\mathbf{R}$ and $\Delta\mathbf{t}$.}
    % \caption{\textbf{Mask-augmented refinement network architecture.} Building on FoundationPose~\cite{foundationposewen2024}, our refinement network processes the rendered and observed RGB-D crops through a frozen shared encoder, followed by a frozen fusion encoder. Hand and object masks are encoded by a lightweight trainable convolutional branch and added with the fused RGB-D features. The network then adds a positional embedding to the resulting feature map and uses two transformer heads to predict rotation and translation updates, $\Delta\mathbf{R}$ and $\Delta\mathbf{t}$.}
    \label{fig:method-refiner-architecture}
\end{figure*}

\begin{abstract}
Estimating the pose of a hand-held tool is particularly challenging: the hand occludes much of the tool, and the visible geometry of the tool is often insufficient to determine its 6-DoF pose due to symmetry and planar structures. Existing pose estimation methods primarily rely on the object's visual appearance and geometry, while the relationship between the hand and tool remains underutilized. We address this problem by turning the hand from a source of occlusion into a source of information for object pose estimation. Given an RGB-D sequence, our method refines the object pose through an iterative render-and-compare process that leverages object and hand masks as additional inputs, and is supervised by a loss that keeps the object's projected silhouette within the hand and object regions. We evaluate our approach on existing object pose estimation benchmarks and a newly collected motion-capture dataset of tool use. Our method consistently outperforms existing approaches across a broad range of metrics, demonstrating robust temporal consistency on challenging tool-use sequences. Our project website is available at: \url{https://tool-pose-tracking.github.io/}.
% Our project website is available at: \url{https://toolposeestimation.github.io/}.
\end{abstract}    
\section{Introduction}
\label{sec:intro}

Consider Figure~\ref{fig:teaser}. A hand grips a fork, leaving the tines visible and the middle of the handle hidden.  Hand-held tools like forks are common in everyday activities. To understand how the tool is being used, we want to estimate its 6-DoF pose over time, tracking both where it moves and how it turns. However, we find that existing estimators and trackers struggle when the hand hides much of the tool. 

Tool use is a common scenario in egocentric datasets~\cite{perrett2025hdepic,Damen2021EPIC-KITCHENS,Grauman2022Ego4D,Hoque2025egodex}, but they lack ground-truth object poses.  Object pose benchmarks~\cite{chao2021dexycb,hampali2020ho3dv3,Liu_2022_hoi4d,liu2024taco,banerjee2025hot3d,YangCVPR2022OakInk,Kwon_2021_h2o} cover diverse grasping and manipulation scenarios, leaving hand-held tools underexplored. To bridge this gap, we create MoCap Tools, a benchmark for tool-use pose estimation and tracking. Our experiments reveal that accurately recovering tool poses in these scenarios remain challenging. Since tools are often thin, symmetric, or textureless, the small portion visible around the fingers provides too little information to determine the full pose. This becomes a problem for state-of-the-art render-and-compare methods~\cite{labbe2022megapose,foundationposewen2024,ponimatkin2025freepose,li2018deepim}, as even at the correct pose, a render of the full tool can differ substantially from the occluded observation.

To mitigate this challenge, consider Figure~\ref{fig:teaser} again: although we cannot see the complete fork, we know that it should continue behind the fingers.
Thus, our key insight is that \emph{the hand mask provides key information about the tool precisely where its appearance is missing.} The object mask identifies the pixels that the projected tool should cover, while the hand mask indicates where the hidden portion may extend. The hand's motion also provides a cue for tracking the tool across frames, and the shape of the grasp gives clues about the tool's orientation. Building on these insights, we introduce per-frame object and hand masks as additional inputs to the pose estimator. To further leverage the masks, we also introduce a novel differentiable silhouette loss that encourages projected tool points to cover the visible object region while penalizing projections outside the union of both masks. 

To validate our approach, we evaluate on existing object pose benchmarks~\cite{chao2021dexycb,hampali2020ho3dv3} and MoCap Tools. Our method outperforms existing approaches on most metrics, achieving the lowest translation error and the lowest 90th-percentile rotation and translation errors across all three datasets. Ablation studies highlight the complementary roles of the hand and object masks and show that the silhouette loss further improves performance on most metrics over mask conditioning alone.

In sum, our contributions are:
\begin{enumerate}
    \item A mask-augmented render-and-compare network for 6-DoF object pose estimation and tracking.
    \item A differentiable silhouette loss using hand and object masks to improve pose refinement under occlusion.
    \item A motion-capture validation dataset focused on egocentric tool use.
    \item An evaluation showing improved tracking robustness on established object pose estimation benchmarks and challenging tool-use sequences.
\end{enumerate}

\section{Related Work}
\label{sec:rel_work}
% be sure to add dima's group's works HopFormer, Rohit

% \subsection{Using hand as additional cues}
% \kwh{should we include ComPose and say the code is not yet released?} 

% \subsection{Object pose estimation}
% recent concurrent also uses the hand for pose estiamtion but uses the hand joints, which we find and joints are noisy, joints don't give information about occlusion, our ablations show that using masks is more useful than joints. hand joints don't provide occlusion patterns

% 4 main ways: render-and-compare, direct regression, ICP, correspondence based

\subsection{Object Pose Estimation}
Single-frame object pose estimation recovers an object's 3D rotation and translation from an RGB or RGB-D observation. 
Earlier learning-based methods directly regress pose from RGB~\cite{xiang2018posecnn} and RGB-D~\cite{wang2019densefusion}. GDR-Net~\cite{Wang_2021_GDRN} later incorporate dense geometric features into RGB pose regression. To handle occlusion, correspondence-based methods like PV-Net~\cite{peng2019pvnet} lets visible pixels vote for hidden keypoints, while other methods establish correspondences through 3D keypoint voting or template matching~\cite{He_2020_pvn3d,ornek2024foundpose}. When matches are ambiguous, render-and-compare methods~\cite{labbe2022megapose,Moon_2024_genflow,li2018deepim,labbe2020cosypose} refine pose hypotheses against the observation. FoundationPose~\cite{foundationposewen2024} uses this strategy for RGB-D registration, and Any6D~\cite{lee2025any6d} extends it to objects specified by a single RGB-D reference image. 

When a grasp hides much of the object, the hand provides additional evidence. UniHOPE~\cite{Wang_2025_unihope} uses grasp-aware feature fusion, while HOPformer~\cite{bansal2026hopformer} attends to WiLoR~\cite{potamias2024wilor} hand features to jointly estimate both hand poses and the object pose. In contrast, our approach uses hand masks rather than hand pose estimates, which we find to be noisy. As the dominant occluder, hand masks directly capture hand-object occlusion patterns not explicitly represented by hand poses, constraining the possible location of occluded object while providing cues about the object's orientation.

\subsection{Object Pose Tracking}
After estimating an initial pose, object pose tracking leverages temporal information to estimate an object's motion across frames. 6-PACK~\cite{wang20196pack} estimates motion by matching 3D keypoints across frames, while BundleSDF~\cite{bundlesdfwen2023} jointly tracks the object and reconstructs its geometry. FoundationPose~\cite{foundationposewen2024} initializes tracking from the previous frame's pose and iteratively refines it by comparing rendred views with the current RGB-D observation. FreePose~\cite{ponimatkin2025freepose} similarly tracks with a 3D model, but retrieves approximate meshes for objects in the RGB videos. 

Despite these advances, tracking hand-held tools remains challenging because persistent hand occlusion limits visual evidence and can cause pose estimates to drift. To address this challenge, ROHIT~\cite{Zhu_2026_rohit} leverages grasp types to propagate object poses. A recent concurrent work ComPose~\cite{shin2026compose} adaptively combines hand-joint motion and object geometry. Unlike these methods, our approach leverages motion cues from the per-frame hand masks, achieving the lowest translation errors across all evaluated metrics.

\section{Method}
\label{sec:method}

% We build on FoundationPose, a model-based method for 6-DoF object pose estimation and tracking and adapt it to hand-held tools, where much of the object is occluded by the hand. We first provide a background on FoundationPose in Section~\ref{subsec:prelim-fp}. We then describe how to inject the mask effectively in Section~\ref{subsec:mask-conditioned-refinement} and the fine-tuning process in~\ref{subsec:fine-tuning}.
% \bh{Long sentence that says many things. Break it up into two: "Model-based object trackers refine the pose by... This is challenging because... This challenge is especially severe for hand-held tools..}
Model-based object trackers estimate pose by comparing a rendering of the object in the predicted pose with the observed image. This is difficult when objects have geometric symmetry or exhibit motion blur from fast camera or object movement. These challenges are further exacerbated for hand-held tools, where a significant part of the object is covered by the hand and the tools are often flat or textureless.
% Model-based object trackers that refine the pose by comparing a rendering of the object with the observed image must decide, from the image alone, which pixels belong to the object. This decision becomes especially challenging for hand-held tools, where a significant part of the object is covered by the hand.
We address these challenges by incorporating object and hand masks to provide additional constraints on the predicted object pose. We implement this approach in a render-and-compare pipeline building on~\cite{foundationposewen2024}. We first review FoundationPose~\cite{foundationposewen2024} in Section~\ref{subsec:prelim-fp}, then introduce our mask-conditioned approach in Section~\ref{subsec:mask-conditioned-refinement}, and describe the fine-tuning process in Section~\ref{subsec:fine-tuning} .

\subsection{Background: A Render-and-Compare Pipeline}
\label{subsec:prelim-fp}

%\subsubsection{Inputs and Outputs}
\label{subsubsec:fp-inputs-outputs}
FoundationPose takes an RGB-D video with known camera intrinsics, a textured object mesh, and an object mask for the first frame. 
% \bh{Say something about where the textured object mesh might come from.} 
The textured mesh is either provided directly as a CAD model or reconstructed using BundleSDF~\cite{bundlesdfwen2023}. 
The outputs are the object poses $\mathbf{T}_i=(\mathbf{R}_i,\mathbf{t}_i)\in \mathrm{SE} (3)$ in camera coordinates for every frame $i$.

%\subsubsection{Method Overview}
\label{subsubsec:method-overview}
FoundationPose works by generating candidate 6D object poses, rendering the object from each candidate viewpoint, and using neural networks to iteratively refine and score the poses based on how well the rendered RGB-D images match the real observation. The process is different for the initial frame and for subsequent frames.

For the initial frame, FoundationPose generates multiple pose hypotheses by sampling different 3D rotations around the object. 
The 3D translation is the same for all hypotheses, and is obtained by back-projecting the center of the object mask's 2D bounding box and moving it to the median depth of the masked pixels. For each candidate pose, it renders the object from the corresponding camera viewpoint to produce an RGB-D image. Both the rendered and observation RGB-D are fed into a refinement network (Section~\ref{subsubsec:fp-refinement-network}), which updates each hypothesis for five iterations. A scoring network (Section~\ref{subsubsec:fp-scoring-network}) takes all refined hypotheses and both rendered and observation RGB-D and ranks all refined hypotheses jointly. The top-ranked hypothesis is selected as the pose estimate. This concludes the registration. 

For subsequent frames, FoundationPose only uses the refinement network.
The refinement network takes the estimated pose from the previous frame and runs two iterations of refinement. The scoring network and the object mask are not used after the first frame, and there is no re-initialization. Thus, pose tracking accuracy on all later frames is determined by the refinement network alone. 

% \subsubsection{Registration (first frame)}
% \label{subsubsec:fp-registration}
% Registration starts with 252 rotation hypotheses, from 42 viewpoints on an icosphere times 6 in-plane rotations in $60^\circ$ increments. All hypotheses share one translation, obtained by back-projecting the center of the object mask's 2D bounding box to the median depth of the masked pixels. A refinement network (Section~\ref{subsubsec:fp-refinement-network}) updates each hypothesis for five iterations. A scoring network (Section~\ref{subsubsec:fp-scoring-network}) ranks all refined hypotheses jointly, and the top-ranked hypothesis is selected as the pose estimate.

% \subsubsection{Tracking}
% \label{subsubsec:fp-tracking}
% For every subsequent frame, the refinement network runs on the previous frame's estimate for two iterations. The scoring network and the object mask are not used after the first frame, and there is no re-initialization. Thus, tracking accuracy is determined by the refinement network alone. 

\subsubsection{Refinement Network}
\label{subsubsec:fp-refinement-network}
The goal of the refinement network is to iteratively predict rotation and translation updates to best match the rendered RGB-D image with the observed image. 

Let $\mathbf{T}_h=(\mathbf{R}_h,\mathbf{t}_h)$ be a candidate pose. Initially, both RGB-D images are cropped to the same square window, centered on the projection of $\mathbf{t}_h$ and spanning $1.2\times$ the mesh diameter $d$ at the depth of $\mathbf{t}_h$, and resized to $160\times160$. Each crop has six channels: RGB scaled to $[0,1]$, and the $XYZ$ map normalized by subtracting $\mathbf{t}_h$ and dividing by the mesh radius $d/2$. Pixels with invalid depth are set to zero. A shared CNN encoder is applied to the rendered and the observed crop. Their features are then concatenated and passed through a fusion encoder, giving a map $F \in \mathbb{R}^{512\times20\times20}$. A sinusoidal positional embedding is added to the 400 patch tokens, and two separate heads, each a transformer encoder layer followed by a linear layer, predict a rotation update $\Delta\mathbf{R} \in \mathrm{SO}(3)$ and a translation update $\Delta\mathbf{t} \in \mathbb{R}^3$. The refined pose is 
\begin{equation}
    \mathbf{T}'_h = \big(\Delta\mathbf{R}\,\mathbf{R}_h,\; \mathbf{t}_h + \Delta\mathbf{t}\big).
    \label{eq:decode}
\end{equation}
During training, the network is supervised by the loss:
\begin{equation}
    \mathcal{L}_{\mathrm{refine}} = 
    \lambda_t \left\| \Delta \mathbf{t} - \Delta \mathbf{t}_{gt} \right\|_2 + \lambda_R \left\| \Delta \mathbf{R} - \Delta \mathbf{R}_{gt} \right\|_2, 
\end{equation}
where $\Delta\mathbf{t}_{gt}$ and $\Delta\mathbf{R}_{gt}$ are ground-truth updates, rotations are compared as axis-angle vectors, and $\lambda_t=1$ and $\lambda_R=0.1$ are the weights listed in the configuration file of the released GitHub repository.

\subsubsection{Scoring Network}
\label{subsubsec:fp-scoring-network}
The scoring network is used only for the first frame to jointly rank all initially generated poses. It takes the same input pairs as the refinement network, but with a crop ratio of 1.1. Its encoders share the same architecture as those of the refinement network. Self-attention is applied to the 400 tokens, followed by average pooling to obtain a 512-dimensional feature vector per hypothesis. Multi-head attention then compares features across all hypotheses and outputs a score per hypothesis. The highest scoring hypothesis becomes the selected pose estimate.

\subsection{Mask-conditioned Refinement}
\label{subsec:mask-conditioned-refinement}
%\subsubsection{Motivation}
%\kwh{would "Existing Challenges" be a better section header?}
% \kwh{TODO: identify why prior art fails especially in the tool setting and why using hand masks is the right way to address this challenge}
% Most of the state-of-the-art pose estimation pipelines follow some kind of render-and-compare approach. This can be especially challenging for estimating the pose of a hand-held tool as the hand often occludes a significant amount of the tool, leaving minimal visible areas, as seen in Figure~\ref{fig:}. In the case of the approach mentioned in Section~\ref{subsec:prelim-fp}, only the initial frame sees the object mask. When you pair minimal visible cues and the lack of object tracks, this results in 
% FP struggles with tool pose estimation and tracking because without object masks every frame, the tracks can easily drift. And when tracks drift, the rotation goes too.
% As we see in the ablation studies, even if we just provide the object masks every frame, FP still struggles because the tool is mostly occluded by the hand. If you ask me to estimate the pose of a tool, I would struggle too. I would naturally rely on the hand. 
\begin{figure}[t]
    \centering
    \includegraphics[
        width=\columnwidth,
        trim={0 0 0 100pt},
        clip
    ]{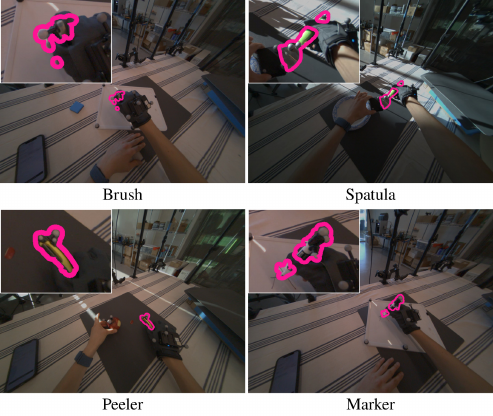}
    \caption{A random sample of object masks from MoCap Tools shows that only small portions of the objects are visible, posing a major challenge for render-and-match methods.}
    \label{fig:method-motivation}
\end{figure}

As discussed above, FoundationPose, like many other pose estimators, renders the object at a hypothesized pose, compares with the observation, and update the hypothesized pose.
Intuitively, we expect the model to update the hypothesized pose so that the rendering matches the observation.
This works well when the object is fully visible.
However, hand-held tools frequently appear heavily occluded by the hand (Figure~\ref{fig:method-motivation}).
In this case, even if the hypothesized pose is correct, the unoccluded rendered object will not match the heavily occluded observation, making the render-and-compare pipeline prone to error.

One might argue that a learnt refinement network, e.g. FoundationPose, can be trained to be resilient to heavy occlusion given enough training data. However, if the learnt model must generalize, it needs to see a huge diversity of occlusion patterns on a wide variety of objects: training data that we simply do not have. 

We posit that a more data-efficient alternative is to provide more information to the refinement network.
At the very least, a mask identifying the visible object pixels should help the network identify precisely which pixels it is trying to match.
Indeed, FoundationPose does use the object mask, but only for the initial frame; as such, on later frames with heavy occlusion, it quickly loses track, never to recover.

However, the object mask is not enough, because the visible part of the object may be small and ambiguous. This is especially true of tools, which tend to be not only very heavily occluded but also small, symmetric, and featureless. For example, a butter knife is thin, silver, and long, and a tiny silver sliver does not tell us where the rest of the knife is. 

Our key insight is that in such ambiguous situations, the \emph{hand silhouette} provides a wealth of information.
First, the hand is the dominant occluder, and knowing the occluder's mask can help the model conjecture where the object's occluded pixels might be.
Second, the hand is causing object motion, so knowing where the hand is allows us to track the object better.
Third, the hand is holding the object, and the hand silhouette provides cues about hand pose, which in turn should provide cues about object orientation.

Based on these insights above, we propose an approach that incorporates object and hand masks as inputs to the render-and-compare pipeline.

%Most of the state-of-the-art pose estimation pipelines follow some form of render-and-compare strategy. However, this approach can be especially challenging when estimating the pose of hand-held tools, as the hand often occludes a significant portion of the tool, leaving only a small fraction of its surface visible, as seen in Figure~\ref{fig:method-motivation}. This challenge is exacerbated in the approach described in Section~\ref{subsec:prelim-fp}, which utilizes the object mask only in the initial frame. The combination of an already limited visual evidence in the initial frame and the absence of object masks in subsequent frames can often lead to tracking failures. Once tracking is lost, pose estimates become unreliable as the rendered object is no longer aligned meaningfully with the observed object image. Even when object masks are available throughout the sequence, accurately inferring the pose of a heavily occluded tool remains challenging. As a human, I would naturally also leverage the hand as a source of contextual information. This observation motivates the key insight of our work: the hand provides valuable complementary cues for estimating and tracking the pose of hand-held tools.\kwh{bold the previous sentence on key insight for emphasis?} Building on this insight, we introduce an approach that jointly leverages hand and object masks to improve tool pose estimation and tracking.  

\subsubsection{Mask-augmented Refinement Network}
We leave the scoring network of FoundationPose as-is and only change the refinement network. The scoring network is only used in the first frame, while the refinement network is used in every frame.

% \bh{Say how we get the masks.} 
We produce object and hand masks with Grounding DINO~\cite{liu2023groundingdino} and SAM 2~\cite{ravi2024sam2}. For each frame, we take the binary object mask and the binary hand mask, and crop both masks $m$ with the same pose-dependent crop used for the observed image described in Section~\ref{subsubsec:fp-refinement-network}. This results in a $2\times160\times160$ input that is pixel-aligned with the refiner's image crop. A three-layer convolutional branch $S$ maps the masks to a $512\times20\times20$ feature map. This matches the shape of the fusion image encoder output $F_I$, so we add the two and leave the rest of the refiner unchanged. The last layer of $S$ is zero-initialized, so the network gives exactly the pretrained refiner's output before fine-tuning. 

%\subsubsection{Scoring Network}
%We keep the pretrained scoring network unchanged. We find empirically that first-frame selection performance worsened after fine-tuning. 

\subsubsection{Silhouette Loss}
Beyond conditioning on hand and object masks, we propose a silhouette loss to supervise the refined pose. Intuitively, the object rendered at a hypothesized pose should cover the visible object pixels and not extend beyond the two masks, which we formulate as $\mathcal{L}_{\mathrm{cov}}$ and $\mathcal{L}_{\mathrm{out}}$, respectively. 

To compute this silhouette loss, $\mathcal{L}_{\mathrm{sil}}$,  while preserving differentiability, we project sampled object mesh points using the refined pose. Specifically, we sample a fixed set of $P=512$ points $\{\mathbf{x}_k\}$ on the mesh surface and project them using the refined pose $\mathbf{T}'_h$ into the crop of hypothesis $\mathbf{T}_h$ to obtain pixel coordinates $\mathbf{u}_k$. Since $\mathbf{T}'_h$ is computed from the network outputs by Equation~\eqref{eq:decode}, the $\mathbf{u}_k$ are differentiable with respect to those outputs. Within the crop, let $O$ be the object mask and $U$ the union of the object and hand masks, both binarized at 0.5.

The coverage loss, $\mathcal{L}_{\mathrm{cov}}$, encourages the projected object to cover all visible object regions. We uniformly sample $Q=512$ pixels $\mathbf{p}_q$ from $O$ with replacement and compute the distance from each sampled pixel to its nearest projected mesh point: 
\begin{equation}
    \mathcal{L}_{\mathrm{cov}}=\frac{1}{Q}\sum_{q=1}^{Q} \min_k\lVert\mathbf{p}_q-\mathbf{u}_k\rVert_2,\\
\end{equation}

The outside loss, $\mathcal{L}_{\mathrm{out}}$, penalizes projected mesh points that fall outside $U$. Let $D_U$ be the Euclidean distance transform of $U$, which measures the distance in pixels for each pixel to the nearest pixel of $U$ (zero inside $U$), we define:
\begin{equation}
    \mathcal{L}_{\mathrm{out}}=
    \frac{1}{P}\sum_k\Biggl[
    D_U\bigl(\pi(\mathbf{u}_k)\bigr) + \bigl\lVert\mathbf{u}_k-\pi(\mathbf{u}_k)\bigr\rVert_2\Biggr]
\end{equation}
where $\pi$ clamps a point to the crop boundaries. The first term penalizes points outside $U$ but within the crop, while the latter additionally penalizes points projected beyond the crop boundaries. 

We combine the two complementary losses, $\mathcal{L}_{\mathrm{cov}}$ and $\mathcal{L}_{\mathrm{out}}$, and normalize by their crop size $W=160$ to obtain: 
\begin{equation}
    \mathcal{L}_{\mathrm{sil}}=\frac{\mathcal{L}_{\mathrm{cov}}+\mathcal{L}_{\mathrm{out}}}{W}.
\end{equation}
The final loss becomes 
\begin{equation}
    \mathcal{L} = \mathcal{L}_{\mathrm{refine}} + 
    \lambda_{sil} \mathcal{L}_{\mathrm{sil}}, 
\end{equation}
with $\lambda_{sil}=0.5$, which is set empirically.

\subsection{Implementation details}
\label{subsec:fine-tuning}

\subsubsection{Training Data}
\label{subsubsec:method-training-finetuning-data}
We initialize our network with FoundationPose and fine-tune on HOI4D's~\cite{Liu_2022_hoi4d} training set, because it contains a large and diverse number of videos with accurate ground-truth object poses. We randomly select 5\% of the clips (66 in total) for validation. For each clip, we segment the object and hand with Grounding DINO~\cite{liu2023groundingdino} and SAM 2~\cite{ravi2024sam2}. We reconstruct the object mesh with BundleSDF~\cite{bundlesdfwen2023}. This follows our test-time evaluation setup, where meshes are also reconstructed from input clips. We express the HOI4D ground-truth object poses in the reconstructed mesh's coordinate frame using a transformation that aligns the two on the first frame. 

%\subsubsection{Training Pairs}
Following~\cite{foundationposewen2024}, for every frame, we generate five hypotheses by perturbing the ground-truth pose. More specifically, we sample a random rotation axis and an angle uniformly from $[0^\circ,20^\circ]$. Translation offsets are sampled uniformly within $2\,cm$ along the $X$- and $Y$-axes and $5\,cm$ along the $Z$-axis. 

\subsubsection{Inference}
At test time, we run the registration and tracking outlined in Section~\ref{subsec:prelim-fp} with the mask-conditioned refiner in both registration and tracking stages. 

\subsubsection{Checkpoint Selection} 
We track the 66 HOI4D validation clips for every saved epoch and select the epoch with the lowest relative rotation and translation errors. We measure relative error in two ways: between consecutive frames $(i, i{+}1)$ and between the initial frame and each subsequent frame $(0, i)$. For a frame pair $(i,j)$, the relative pose is $\mathbf{T}_j\mathbf{T}_i^{-1}$, which we compute for both the prediction and the ground truth in camera coordinates. The rotation error is the geodesic angle between their rotations, and the translation error is the Euclidean distance between their translations. For each clip, we take the root mean square of each error over all pairs, then average over clips. We select the epoch~1 checkpoint, because it has the lowest overall error across all four relative metrics among the first 10 epochs. 
\section{Evaluation}
\label{sec:eval}
% Metric notes
% per frame consec trans RMSE is capped by 1000mm, 
% per frame init trans RMSE is capped by 2000mm, 

\newcommand{\posecell}[3][\relax]{%
    \posecellparts{#1}{#2}#3\relax
}
\def\posecellparts#1#2#3.#4\relax{%
    \rlap{
        \phantom{\textbf{#2}}%
        \settowidth{\dimen0}{#1{#3}}%
        \kern-\dimen0 #1{#3.#4}%
    }%
    \phantom{\textbf{#2.00}}%
}

\begin{table*}[!t]
\centering
\footnotesize
\setlength{\tabcolsep}{2pt}
\renewcommand{\arraystretch}{0.95}
\begin{adjustbox}{max width=\textwidth}
\begin{tabular}{@{}ll*{6}{c}@{\hspace{6pt}}*{6}{c}@{}}
\toprule
& & \multicolumn{6}{c}{\textbf{Consecutive ($i$ to $i+1$)}}
  & \multicolumn{6}{c}{\textbf{Initial-to-subsequent ($0$ to $i$)}} \\
\cmidrule(lr){3-8}\cmidrule(lr){9-14}
& & \multicolumn{2}{c}{Mean ($\downarrow$)}
  & \multicolumn{2}{c}{P90 ($\downarrow$)}
  & \multicolumn{2}{c}{AUC ($\uparrow$)}
  & \multicolumn{2}{c}{Mean ($\downarrow$)}
  & \multicolumn{2}{c}{P90 ($\downarrow$)}
  & \multicolumn{2}{c}{AUC ($\uparrow$)} \\
\cmidrule(lr){3-4}\cmidrule(lr){5-6}\cmidrule(lr){7-8}
\cmidrule(lr){9-10}\cmidrule(lr){11-12}\cmidrule(lr){13-14}
Dataset & Method
  & RRE & RTE & RRE & RTE & ADD & ADD-S
  & RRE & RTE & RRE & RTE & ADD & ADD-S \\
\midrule

DexYCB & MegaPose
  & \posecell{000}{8.89} & \posecell{0000}{43.9}
  & \posecell{000}{21.33} & \posecell{0000}{82.5}
  & \posecell{00}{73.40} & \posecell{00}{83.59}
  & \posecell{000}{63.08} & \posecell{0000}{243.5}
  & \posecell{000}{154.51} & \posecell{0000}{630.4}
  & \posecell{00}{34.03} & \posecell{00}{44.65} \\
& FP (pretrained)
  & \posecell{000}{3.07} & \posecell{0000}{8.3}
  & \posecell{000}{7.24} & \posecell{0000}{24.1}
  & \posecell{00}{91.31} & \posecell{00}{95.12}
  & \posecell{000}{25.17} & \posecell{0000}{77.6}
  & \posecell{000}{90.24} & \posecell{0000}{327.8}
  & \posecell{00}{68.97} & \posecell{00}{74.67} \\
& FP (fine-tuned)
  & \posecell{000}{2.24} & \posecell{0000}{6.0}
  & \posecell{000}{5.26} & \posecell{0000}{17.1}
  & \posecell{00}{93.60} & \posecell{00}{96.32}
  & \posecell{000}{22.68} & \posecell{0000}{58.5}
  & \posecell{000}{80.95} & \posecell{0000}{200.1}
  & \posecell{00}{72.54} & \posecell{00}{79.73} \\
& FoundPose
  & \posecell{000}{72.26} & \posecell{0000}{388.9}
  & \posecell{000}{173.76} & \posecell{0000}{1000.0}
  & \posecell{00}{17.69} & \posecell{00}{30.89}
  & \posecell{000}{92.38} & \posecell{0000}{622.1}
  & \posecell{000}{174.17} & \posecell{0000}{1925.3}
  & \posecell{00}{10.02} & \posecell{00}{23.96} \\
& FreePose
  & \posecell[\textbf]{000}{2.11} & \posecell{0000}{32.6}
  & \posecell{000}{5.81} & \posecell{0000}{57.9}
  & \posecell{00}{80.96} & \posecell{00}{87.67}
  & \posecell{000}{31.06} & \posecell{0000}{254.3}
  & \posecell{000}{96.40} & \posecell{0000}{697.1}
  & \posecell{00}{34.83} & \posecell{00}{46.33} \\
& UniHOPE
  & \posecell{000}{53.08} & \posecell{0000}{367.2}
  & \posecell{000}{149.95} & \posecell{0000}{1000.0}
  & \posecell{00}{23.34} & \posecell{00}{30.63}
  & \posecell{000}{108.43} & \posecell{0000}{775.0}
  & \posecell{000}{170.76} & \posecell{0000}{2000.0}
  & \posecell{00}{9.17} & \posecell{00}{11.41} \\
& Ours (w/o sil. loss)
  & \posecell[\underline]{000}{2.20} & \posecell[\underline]{0000}{4.8}
  & \posecell[\underline]{000}{5.00} & \posecell[\underline]{0000}{10.7}
  & \posecell[\underline]{00}{94.77} & \posecell[\underline]{00}{96.86}
  & \posecell[\underline]{000}{19.74} & \posecell[\underline]{0000}{42.6}
  & \posecell[\underline]{000}{64.07} & \posecell[\underline]{0000}{60.7}
  & \posecell[\underline]{00}{77.38} & \posecell[\underline]{00}{85.78} \\
\rowcolor{oursblue}
& Ours (w/ sil. loss)
  & \posecell[\underline]{000}{2.20} & \posecell[\textbf]{0000}{3.9}
  & \posecell[\textbf]{000}{4.97} & \posecell[\textbf]{0000}{8.3}
  & \posecell[\textbf]{00}{95.59} & \posecell[\textbf]{00}{97.26}
  & \posecell[\textbf]{000}{18.78} & \posecell[\textbf]{0000}{28.5}
  & \posecell[\textbf]{000}{56.81} & \posecell[\textbf]{0000}{40.6}
  & \posecell[\textbf]{00}{79.43} & \posecell[\textbf]{00}{88.47} \\
\midrule

HO3Dv3 & MegaPose
  & \posecell{000}{2.49} & \posecell{0000}{16.0}
  & \posecell{000}{5.49} & \posecell{0000}{6.9}
  & \posecell{00}{95.15} & \posecell{00}{96.54}
  & \posecell{000}{27.87} & \posecell{0000}{141.6}
  & \posecell{000}{123.72} & \posecell{0000}{237.0}
  & \posecell{00}{75.75} & \posecell{00}{81.17} \\
& FP (pretrained)
  & \posecell{000}{1.25} & \posecell[\underline]{0000}{1.4}
  & \posecell{000}{2.43} & \posecell{0000}{2.0}
  & \posecell[\underline]{00}{98.18} & \posecell[\underline]{00}{98.65}
  & \posecell[\textbf]{000}{16.55} & \posecell[\underline]{0000}{14.1}
  & \posecell{000}{86.40} & \posecell{0000}{27.9}
  & \posecell[\underline]{00}{82.24} & \posecell[\underline]{00}{91.01} \\
& FP (fine-tuned)
  & \posecell[\textbf]{000}{0.94} & \posecell{0000}{10.7}
  & \posecell[\underline]{000}{1.93} & \posecell{0000}{2.0}
  & \posecell{00}{93.57} & \posecell{00}{94.71}
  & \posecell{000}{18.28} & \posecell{0000}{128.5}
  & \posecell[\underline]{000}{82.28} & \posecell[\underline]{0000}{26.3}
  & \posecell{00}{80.39} & \posecell{00}{88.18} \\
& FoundPose
  & \posecell{000}{33.24} & \posecell{0000}{94.1}
  & \posecell{000}{136.92} & \posecell{0000}{139.4}
  & \posecell{00}{59.38} & \posecell{00}{75.91}
  & \posecell{000}{61.94} & \posecell{0000}{252.7}
  & \posecell{000}{174.82} & \posecell{0000}{1024.0}
  & \posecell{00}{40.63} & \posecell{00}{64.20} \\
& FreePose
  & \posecell{000}{1.62} & \posecell{0000}{4.3}
  & \posecell{000}{2.80} & \posecell{0000}{9.8}
  & \posecell{00}{95.41} & \posecell{00}{97.18}
  & \posecell{000}{88.94} & \posecell{0000}{86.4}
  & \posecell{000}{148.97} & \posecell{0000}{200.7}
  & \posecell{00}{17.78} & \posecell{00}{53.52} \\
& UniHOPE
  & \posecell{000}{30.66} & \posecell{0000}{191.7}
  & \posecell{000}{122.34} & \posecell{0000}{842.9}
  & \posecell{00}{51.66} & \posecell{00}{60.42}
  & \posecell{000}{63.76} & \posecell{0000}{358.5}
  & \posecell{000}{164.23} & \posecell{0000}{1152.1}
  & \posecell{00}{16.02} & \posecell{00}{28.84} \\
& Ours (w/o sil. loss)
  & \posecell[\underline]{000}{0.99} & \posecell{0000}{5.2}
  & \posecell{000}{1.94} & \posecell[\underline]{0000}{1.9}
  & \posecell{00}{95.58} & \posecell{00}{96.37}
  & \posecell[\underline]{000}{17.39} & \posecell{0000}{81.0}
  & \posecell{000}{86.32} & \posecell{0000}{27.8}
  & \posecell{00}{81.39} & \posecell{00}{90.70} \\
\rowcolor{oursblue}
& Ours (w/ sil. loss)
  & \posecell[\underline]{000}{0.99} & \posecell[\textbf]{0000}{1.0}
  & \posecell[\textbf]{000}{1.87} & \posecell[\textbf]{0000}{1.8}
  & \posecell[\textbf]{00}{98.63} & \posecell[\textbf]{00}{98.85}
  & \posecell{000}{18.52} & \posecell[\textbf]{0000}{10.1}
  & \posecell[\textbf]{000}{80.20} & \posecell[\textbf]{0000}{23.7}
  & \posecell[\textbf]{00}{82.56} & \posecell[\textbf]{00}{93.91} \\
\midrule

MoCap Tools & MegaPose
  & \posecell{000}{15.84} & \posecell{0000}{77.6}
  & \posecell{000}{38.89} & \posecell{0000}{147.2} & -- & --
  & \posecell{000}{98.62} & \posecell{0000}{300.5}
  & \posecell{000}{169.48} & \posecell{0000}{743.2} & -- & -- \\
& FP (pretrained)
  & \posecell{000}{3.68} & \posecell{0000}{5.8}
  & \posecell{000}{7.83} & \posecell{0000}{14.6} & -- & --
  & \posecell{000}{44.26} & \posecell{0000}{55.0}
  & \posecell{000}{149.98} & \posecell{0000}{176.6} & -- & -- \\
& FP (fine-tuned)
  & \posecell{000}{2.58} & \posecell[\underline]{0000}{3.7}
  & \posecell{000}{5.67} & \posecell[\underline]{0000}{7.8} & -- & --
  & \posecell[\underline]{000}{42.04} & \posecell[\underline]{0000}{32.5}
  & \posecell[\underline]{000}{147.48} & \posecell[\underline]{0000}{101.8} & -- & -- \\
& FoundPose
  & \posecell{000}{89.66} & \posecell{0000}{441.7}
  & \posecell{000}{175.78} & \posecell{0000}{899.6} & -- & --
  & \posecell{000}{109.76} & \posecell{0000}{636.3}
  & \posecell{000}{176.08} & \posecell{0000}{1158.8} & -- & -- \\
& FreePose
  & \posecell[\textbf]{000}{2.36} & \posecell{0000}{12.9}
  & \posecell[\underline]{000}{5.49} & \posecell{0000}{29.6} & -- & --
  & \posecell{000}{79.58} & \posecell{0000}{306.1}
  & \posecell{000}{158.01} & \posecell{0000}{780.1} & -- & -- \\
& UniHOPE
  & \posecell{000}{42.80} & \posecell{0000}{339.4}
  & \posecell{000}{142.13} & \posecell{0000}{1000.0} & -- & --
  & \posecell{000}{116.24} & \posecell{0000}{1325.0}
  & \posecell{000}{170.64} & \posecell{0000}{2000.0} & -- & -- \\
& Ours (w/o sil. loss)
  & \posecell{000}{2.93} & \posecell{0000}{25.5}
  & \posecell{000}{7.05} & \posecell{0000}{14.1} & -- & --
  & \posecell{000}{48.73} & \posecell{0000}{118.0}
  & \posecell{000}{162.15} & \posecell{0000}{119.1} & -- & -- \\
\rowcolor{oursblue}
& Ours (w/ sil. loss)
  & \posecell[\underline]{000}{2.38} & \posecell[\textbf]{0000}{3.2}
  & \posecell[\textbf]{000}{5.07} & \posecell[\textbf]{0000}{6.6} & -- & --
  & \posecell[\textbf]{000}{31.56} & \posecell[\textbf]{0000}{27.6}
  & \posecell[\textbf]{000}{82.77} & \posecell[\textbf]{0000}{80.9} & -- & -- \\
\bottomrule
\end{tabular}
\end{adjustbox}
\caption{Pose estimation results across all temporal intervals and benchmarks for all methods. RRE is in degrees, RTE is in mm, and ADD and ADD-S are in percent. FP denotes FoundationPose. Bold values indicate best performance and underlined values are second-best.}
\label{tab:pose_estimation}
\end{table*}

\subsection{Datasets}
\label{subsec:eval-datasets}
We evaluate our method on two established RGB-D object pose benchmarks: DexYCB~\cite{chao2021dexycb} and HO3Dv3~\cite{hampali2020ho3dv3}. For DexYCB, we use the S3 split designed to evaluate performance for unseen objects. These datasets cover a wide range of hand-object interaction scenarios, object categories, and scene conditions, allowing us to assess the robustness and generalization of our method.

To further quantify our method, we create MoCap Tools, an egocentric tool-use validation dataset captured with a motion capture (MoCap). The dataset consists of 11 manipulation sequences featuring 11 everyday tools interacting with various target objects. We capture egocentric RGB streams using Meta Project Aria Gen 2 glasses and obtain depth maps by applying FoundationStereo~\cite{wen2025foundationstereo} to stereo image pairs from the Aria glasses SLAM cameras. We record ground-truth 6-DoF tool poses using a synchronized OptiTrack setup, with markers attached to both the tools and the headset. To register the MoCap measurements to the egocentric camera coordinate frame, we calibrate the rigid transformation between the headset's marker and its RGB camera. Lastly, we temporally synchronize the MoCap and egocentric streams by estimating their clock offset from their angular velocity measurements.

\subsection{Baseline Methods and Setup}
We compare against five pose estimation and tracking methods, including two variants of FoundationPose~\cite{foundationposewen2024}. These include single-image methods, like MegaPose~\cite{labbe2022megapose}, FoundPose~\cite{ornek2024foundpose}, and UniHOPE~\cite{Wang_2025_unihope}, as well as sequence-based methods, like FoundationPose~\cite{foundationposewen2024} and FreePose~\cite{ponimatkin2025freepose}. For FoundationPose, we evaluate both the original model with pretrained weights and a fine-tuned variant trained on data described in Section~\ref{subsubsec:method-training-finetuning-data}. We refer to these two methods in Table~\ref{tab:pose_estimation} as FoundationPose (pretrained) and FoundationPose (fine-tuned), respectively. Unlike FoundationPose, our approach takes both object and hand masks as inputs. We also evaluate the performance of our approach with and without the silhouette loss.

To ensure a fair comparison, we supply all methods with the same RGB-D observations, camera intrinsics, and object meshes. For single-image methods, we perform inference independently on each frame and aggregate the pose estimates. For UniHOPE, we use the publicly available pretrained weights corresponding to DexYCB and HO3D when evaluating on their respective benchmarks. We evaluate UniHOPE on MoCap Tools with its HO3D pretrained weights due to greater similarity in object categories. We reconstruct all object meshes using BundleSDF~\cite{bundlesdfwen2023}.

% \FloatBarrier
\begin{figure*}[!t]
    \centering
    \begin{minipage}{0.95\textwidth}
        \centering
        \noindent
        \hspace*{0.05\linewidth}%
        \begin{minipage}{0.90\linewidth}
            \centering
            \begin{tikzpicture}
                \draw[->, line width=1.2pt]
                    (0,0) -- node[midway, above] {\small time} (0.9\linewidth,0);
            \end{tikzpicture}%
        \end{minipage}\par
        \vspace{0.1cm}
        \mocapgridrow{Ours\\(w/ silhouette loss)}{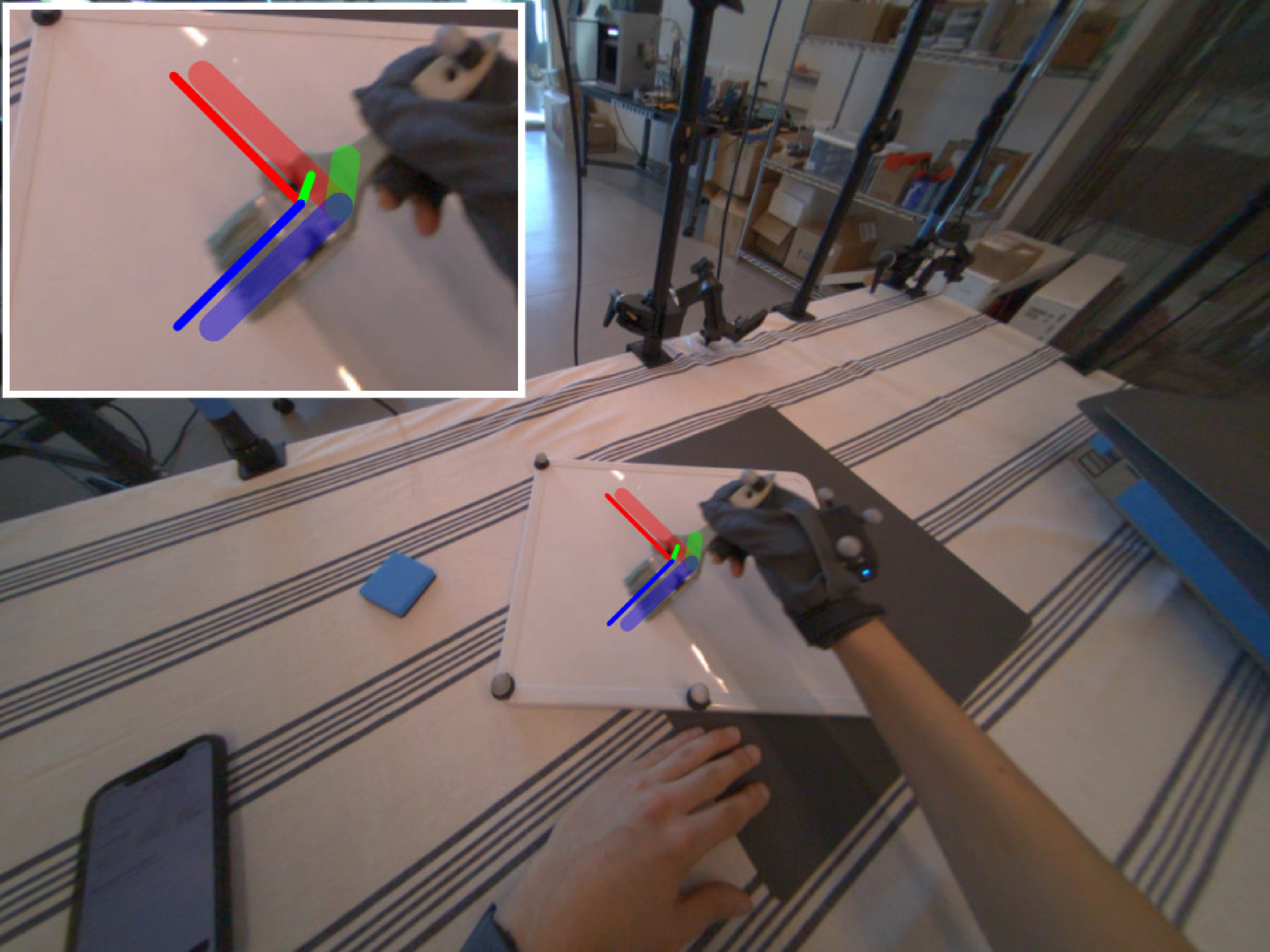}
        \vspace{0.02cm}
    
        \mocapgridrow{FoundationPose\\Pretrained}{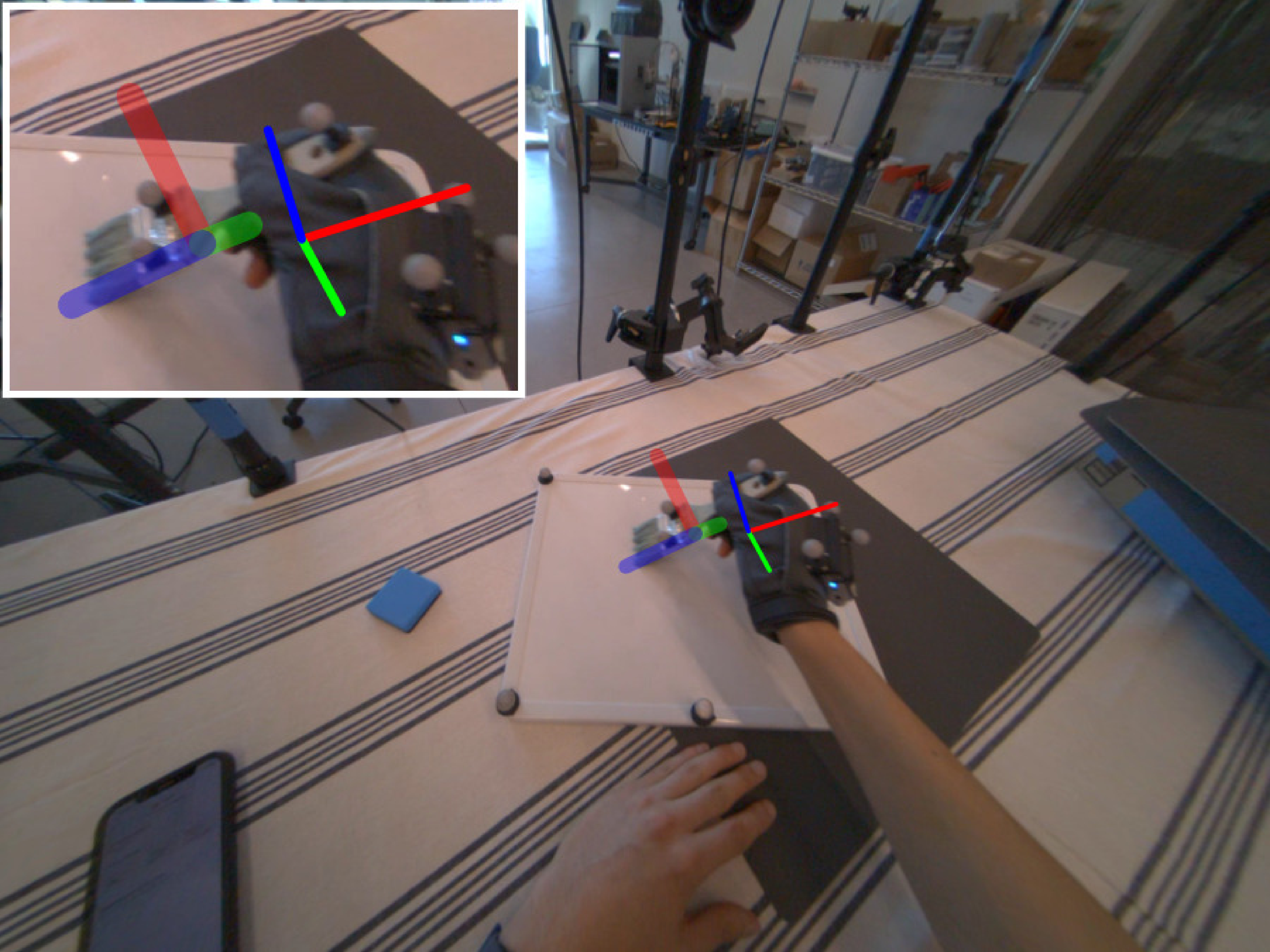}
        \vspace{0.02cm}
    
        \mocapgridrow{FoundationPose\\Fine-tuned}{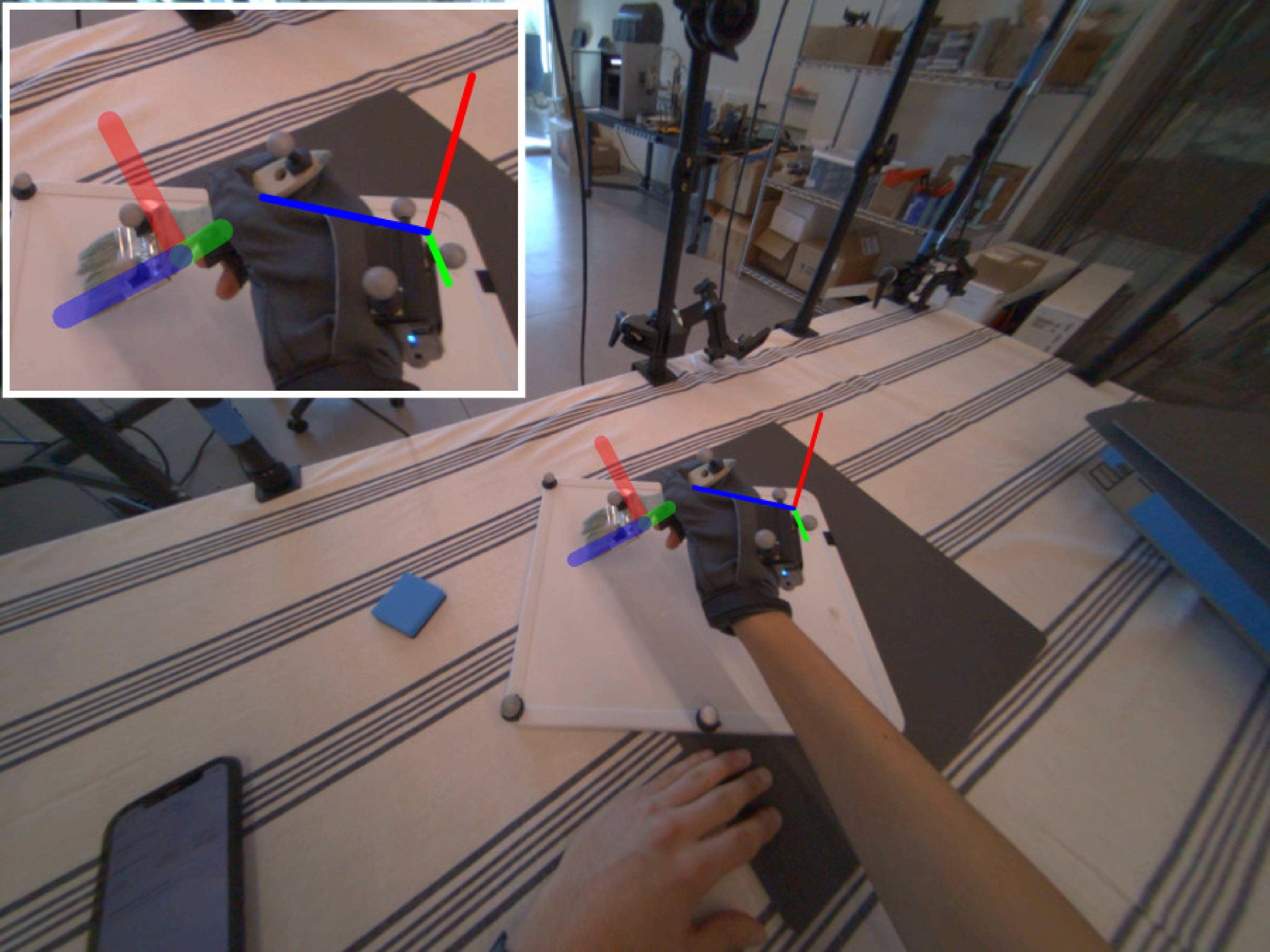}
        \vspace{0.02cm}
    
        \mocapgridrow{MegaPose}{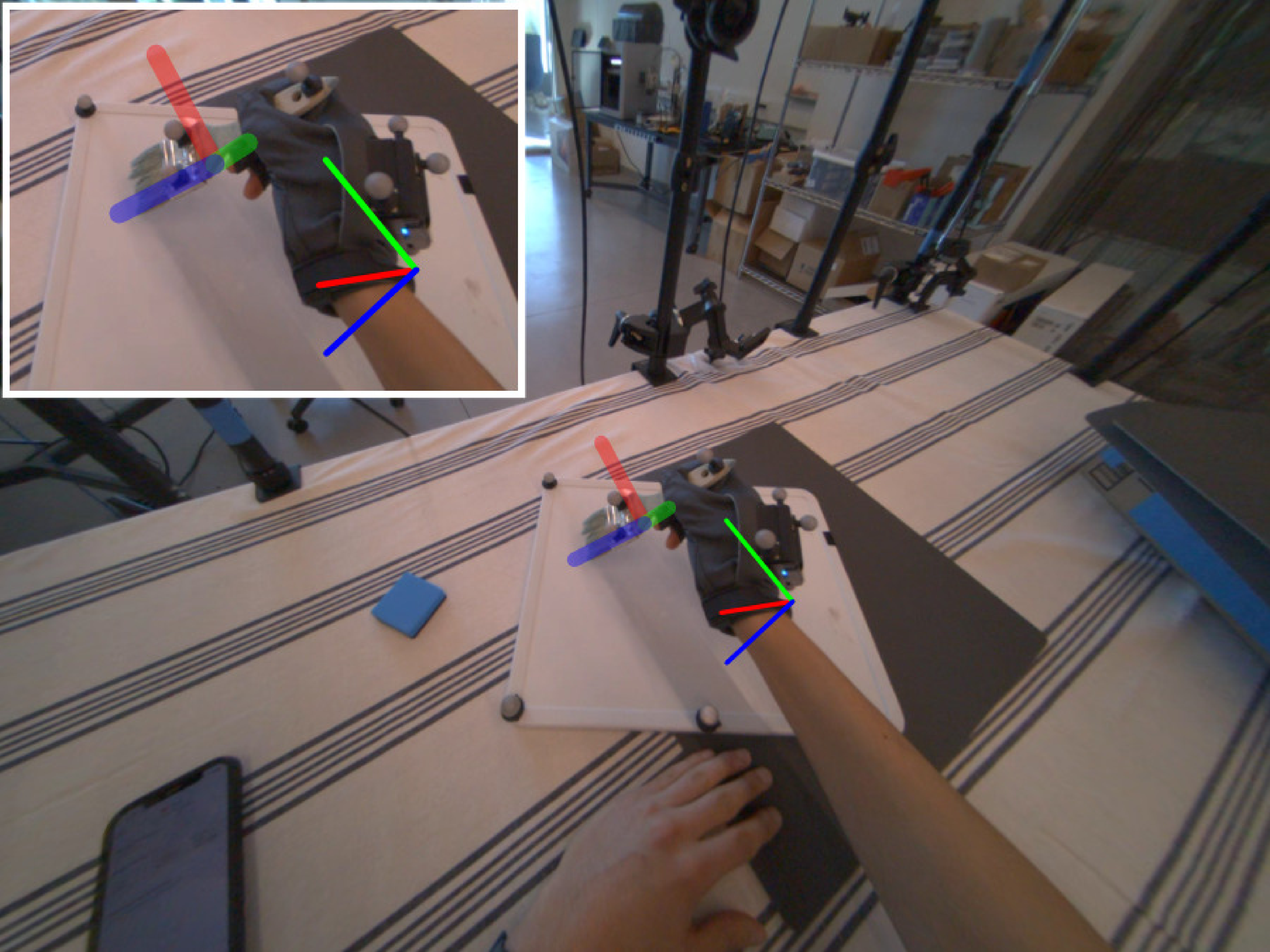}
        \vspace{0.02cm}
    
        \mocapgridrow{FoundPose}{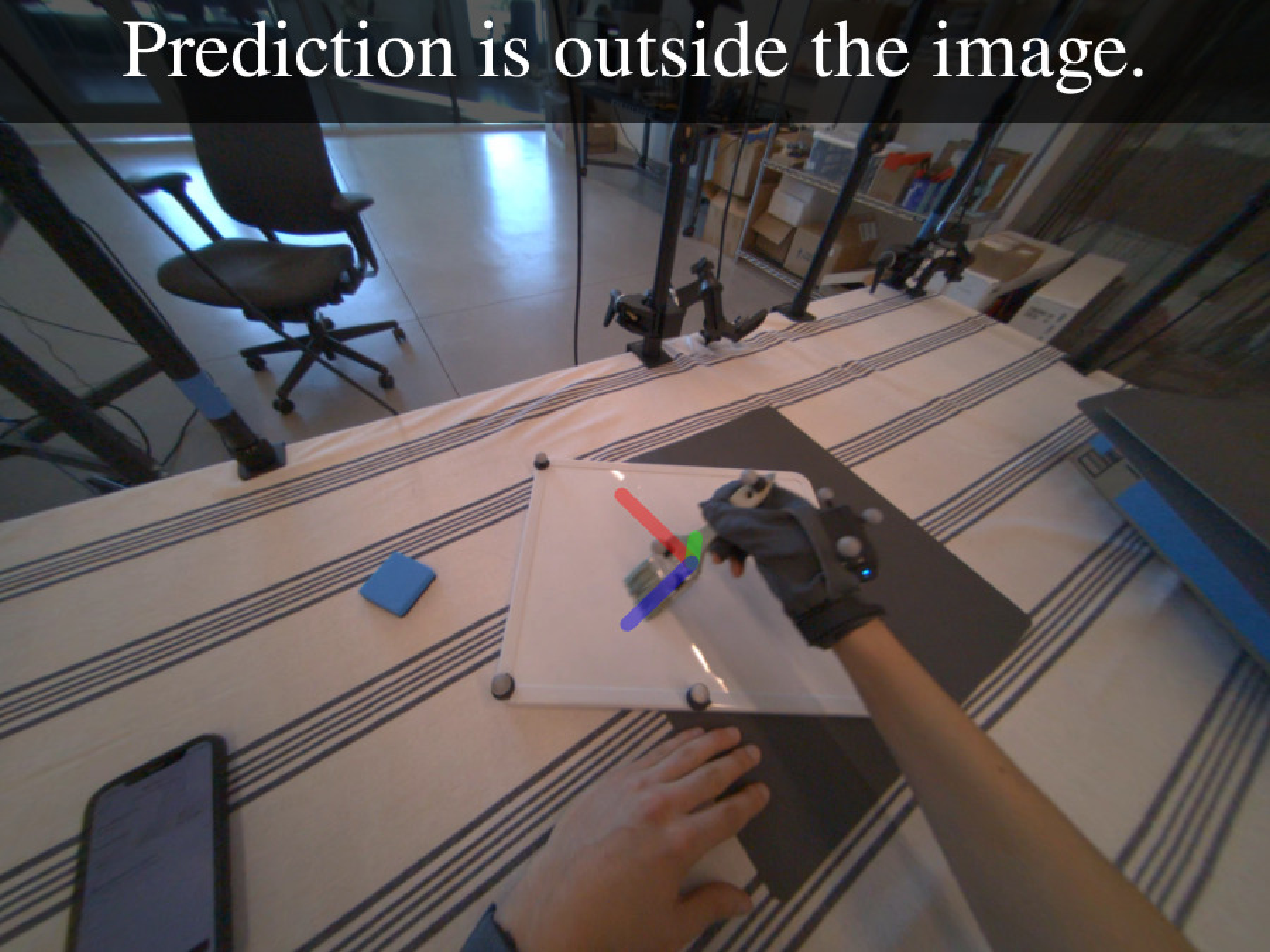}
        \vspace{0.02cm}
    
        \mocapgridrow{FreePose}{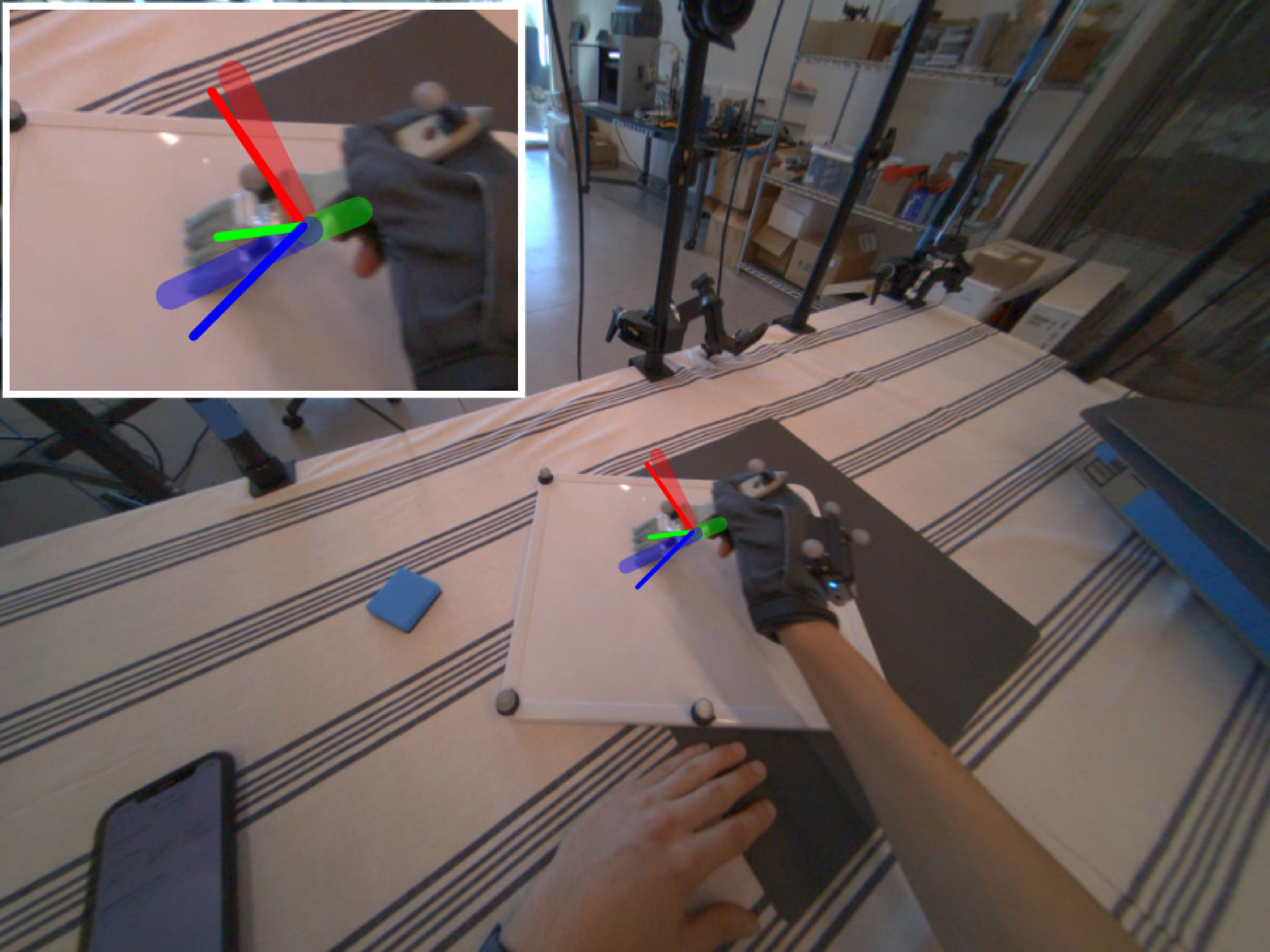}
        \vspace{0.02cm}
    
        \mocapgridrow{UniHOPE}{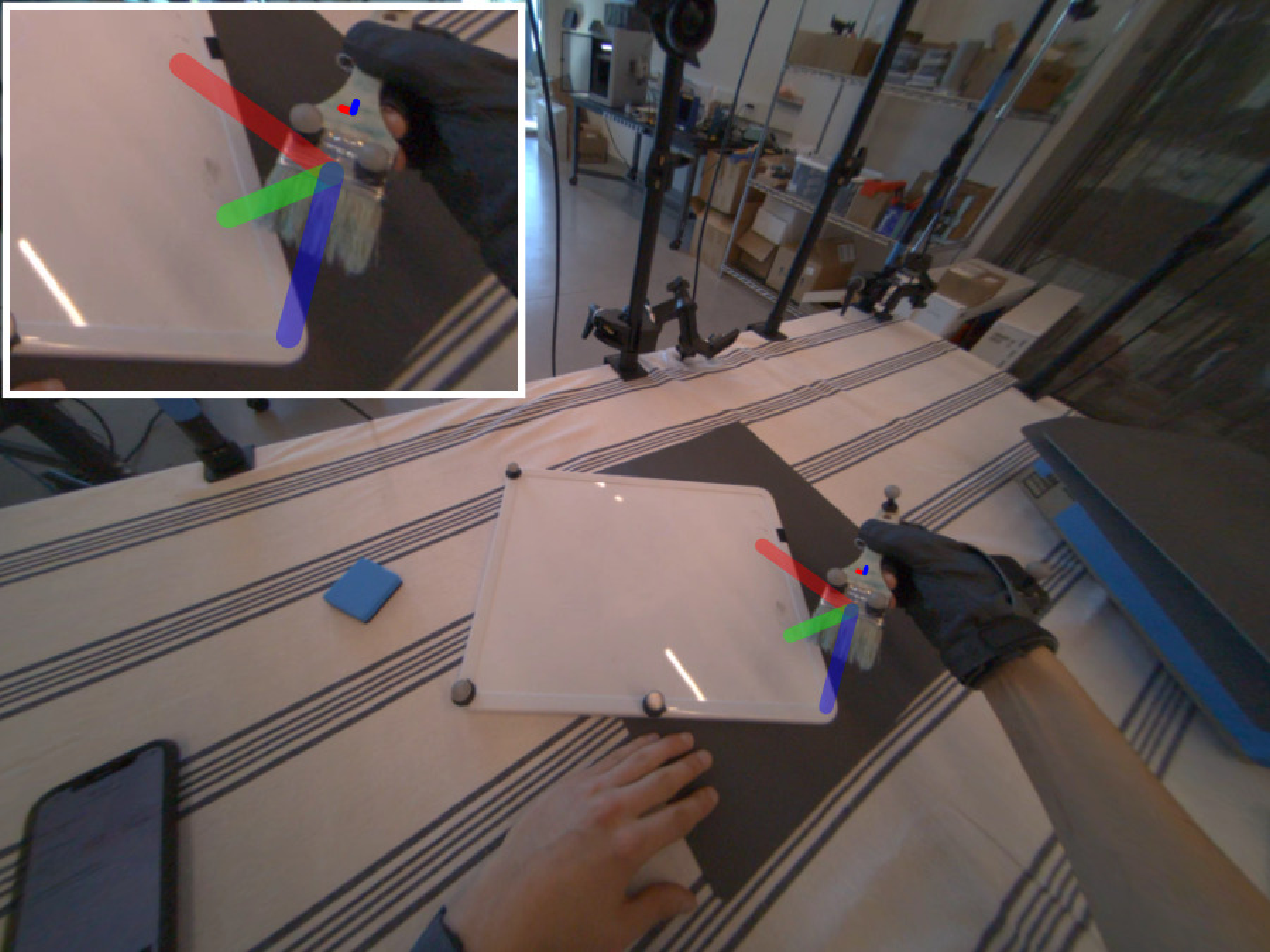}
        \vspace{0.02cm}         

    \end{minipage}
    \caption{Qualitative results on MoCap Tools across a video sequence. The saturated RGB coordinate axes indicate predicted poses, while the thicker, semi-transparent axes represent ground-truth poses. Zoomed-in views of the predicted and ground-truth poses are shown in the top-left corner of each keyframe. }
    \label{fig:evaluation-qualitative}
\end{figure*}

\subsection{Metrics}
\label{subsec:eval-metrics}
We evaluate relative pose estimation performance over two temporal intervals: 
consecutive frames ($i$ to $i+1$) and initial frame to each subsequent frame ($0$ to $i$). The former measures local frame-to-frame motion accuracy, while the latter captures accumulated pose drift over time. We assess performance along three axes: accuracy, robustness, and geometric alignment. Relative Rotation Error (RRE) and Relative Translation Error (RTE) measure pose accuracy, and we report the per-clip mean error. The 90th-percentile error (P90) captures the worst-performing tail of the error distribution across all pooled frames. Average Distance of Model Points (ADD) computes the mean distance between corresponding model points transformed by the predicted and ground-truth poses. Its symmetry-aware variant (ADD-S) accounts for object symmetries by measuring the mean nearest-neighbor distance. To limit the influence of extreme outliers, we cap per-clip translation errors at $1000\,\mathrm{mm}$ for consecutive-frame evaluation and $2000\,\mathrm{mm}$ for initial-to-subsequent-frame evaluation before taking the mean.

\subsection{Pose Estimation and Tracking}
\label{subsec:eval-pose-estimation-tracking}
\subsubsection{Quantitative Analysis}
\label{subsubsec:eval-quantitative}
We present quantitative comparisons with state-of-the-art methods in Table~\ref{tab:pose_estimation}. Our method consistently outperforms all evaluated baselines across datasets, metrics, and temporal intervals, except for four mean RRE metrics. These results demonstrate robust pose estimation and tracking across diverse objects, manipulations, and scenes for both short-term motion and long-range tracking consistency. ADD/ADD-S are not reported for MoCap Tools, because it has no object models registered to the ground-truth frame.

Our method achieves the lowest RTE across all datasets and temporal intervals, outperforming methods with per-frame object masks, demonstrating more accurate object tracking. It also has the lowest P90 errors across all datasets and temporal intervals, indicating robustness against large pose deviations. In the initial-to-subsequent setting, our method reduces rotation and translation P90 errors by 19.2\% and 21.1\% on average across the three datasets, respectively, compared to the second-best baseline.

For mean RRE, our method achieves the best performance in all but four dataset-temporal interval combinations. Specifically, it ranks a close second across all three datasets in the consecutive-frame setting, within $0.1^\circ$ of the best method and third in HO3Dv3 in the initial-to-subsequent-frame setting. The comparatively smaller gains may stem from dataset-specific biases introduced by training on a single dataset. Training on more diverse datasets may further improve generalization and rotational accuracy.

\subsubsection{Qualitative Analysis}
\label{subsubsec:eval-qualitative}
Figure~\ref{fig:evaluation-qualitative} presents a qualitative comparison with existing methods on a hand-held brush sequence from MoCap Tools. We visualize the predicted and ground-truth object poses with RGB axes, where saturated axes represent predicted and wider, more transparent axes indicate ground-truth. 

Existing methods struggle to maintain accurate object poses under severe hand occlusion. Single-image methods---MegaPose, FoundPose, and UniHOPE---show substantial rotational and translational errors on all selected frames. FreePose achieves more robust object tracking by leveraging Grounding DINO for object detection and SAM 2 for mask propagation, but struggles with rotation estimation. Both the pretrained and fine-tuned versions of FoundationPose exhibit tracking drift starting from the third frame. 

In contrast, our method maintains more consistent alignment with ground-truth poses across selected frames, demonstrating the effectiveness of mask conditioning and silhouette loss. 
% In contrast, our method maintains consistent object poses across the selected frames. Our predictions overall exhibit more robust and consistent pose alignment with the ground-truth, demonstrating the effectiveness of incorporating both hand and object masks and our silhouette loss. 

\begin{table}[!tb]
\centering
\small
\setlength{\tabcolsep}{3pt}
\renewcommand{\arraystretch}{0.95}
\setlength{\aboverulesep}{0.25ex}
\setlength{\belowrulesep}{0.35ex}

\begin{adjustbox}{
  max width=\columnwidth,
  max totalheight=0.95\textheight,
  keepaspectratio
}
% \begin{tabular}{llccccc}
\begin{tabular}{lcccc}

\toprule
  & \multicolumn{2}{c}{Mean ($\downarrow$)}
  & \multicolumn{2}{c}{P90 ($\downarrow$)} \\
  % & Weighted pose err. ($\downarrow$) \\
\cmidrule(lr){2-3}
\cmidrule(lr){4-5}
  Variant
  & RRE ($^\circ$) & RTE (mm)
  & RRE ($^\circ$) & RTE (mm) \\

\midrule

% DexYCB
% & Pretrained
% & 32.36 & 354.8 & 90.24 & 976.4 & 673.9 \\
% & No additional inputs
% & 29.02 & 338.6 & 80.95 & 921.3 & 615.3 \\
% & + hand masks
% & 27.41 & \underline{315.6} & 76.26 & 842.7 & 578.7 \\
% & + object masks
% & \textbf{23.33} & \textbf{271.3} & \textbf{55.17}
% & \underline{674.4} & \textbf{493.9} \\
% & + both masks (hand and object)
% & \underline{23.78} & \textbf{271.3} & \underline{56.81}
% & \textbf{652.6} & \underline{500.8} \\
% & + both masks + wrist pose
% & 30.15 & 362.8 & 86.47 & 949.7 & 649.7 \\
% & + both masks + hand pose
% & 31.01 & 380.0 & 91.67 & 1033.0 & 672.2 \\
% \midrule

% HO3Dv3
% & Pretrained
% & \textbf{18.96} & \textbf{150.5} & 86.40
% & \underline{666.0} & \textbf{364.7} \\
% & No additional inputs
% & 21.68 & 417.7 & 82.28 & 855.9 & 591.6 \\
% & + hand masks
% & \underline{21.02} & 1067.0 & 90.70 & 894.7 & 1204.2 \\
% & + object masks
% & 21.18 & 373.1 & 85.15 & 844.7 & 543.5 \\
% & + both masks (hand and object)
% & \underline{21.02} & \underline{151.1} & \textbf{80.20}
% & \textbf{625.8} & \underline{398.9} \\
% & + both masks + wrist pose
% & 22.17 & 822.4 & \underline{82.26} & 795.7 & 982.3 \\
% & + both masks + hand pose
% & 26.57 & 1473.3 & 92.52 & 854.9 & 1710.6 \\
% \midrule

Pretrained
& \posecell{0}{44.26} & \posecell{000}{55.0} & \posecell{00}{149.98} & \posecell{000}{176.6} \\
No additional inputs
& \posecell{0}{42.04} & \posecell{000}{32.5} & \posecell{00}{147.48} & \posecell{000}{101.8} \\
+ hand masks
& \posecell{0}{39.11} & \posecell[\textbf]{000}{27.0} & \posecell{00}{149.01} & \posecell{000}{92.1} \\
+ object masks
& \posecell{0}{52.25} & \posecell{000}{27.7} & \posecell{00}{160.98} & \posecell[\underline]{000}{84.0} \\
\rowcolor{oursblue}
+ both masks (hand and object)
& \posecell[\textbf]{0}{31.56} & \posecell[\underline]{000}{27.6} & \posecell[\textbf]{00}{82.77}
& \posecell[\textbf]{000}{80.9} \\
+ both masks + wrist poses
& \posecell[\underline]{0}{33.78} & \posecell{000}{33.1} & \posecell{00}{116.36}
& \posecell{000}{110.5} \\% & \underline{761.1} \\
+ both masks + hand poses
& \posecell{0}{36.23} & \posecell{000}{30.1} & \posecell[\underline]{00}{104.77} & \posecell{000}{99.2} \\% & 803.8 \\
\bottomrule
\end{tabular}
\end{adjustbox}

\caption{Ablation study evaluating the impact of training with different combinations of hand masks, object masks, wrist poses, and hand poses, using silhouette loss under the MoCap Tools initial-to-subsequent setting.}
\label{tab:pose_ablation}
\end{table}

\subsection{Ablation Studies}
\label{subsec:eval-ablations}

\begin{figure}[!tb]
  \centering
  \includegraphics[width=\columnwidth]{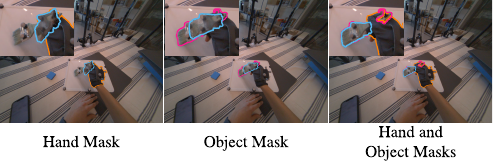}
  \caption{Object mesh outlines for three input configurations: hand mask only, object mask only, and both masks.}
  \label{fig:eval-ablation-masks}
\end{figure}

\subsubsection{Masks}
\label{subsubsec:ablation-masks}
% both masks give good results, so we want to see the effect of each mask
% table 2 shows quantitatively, ...
% intuitively, this makes sense because . this is shown in fig4
We first investigate the individual contributions of the hand and object masks. To isolate their effects, we train two variants in which the refiner receives either only the hand masks or the object masks, with all other settings unchanged. Table~\ref{tab:pose_ablation} shows that each mask individually improves the translation accuracy, with either minimal improved or degraded rotation performance. Combining both masks yields lower errors on all but one metric, validating that they provide complementary information.

This behavior can be understood through their roles in the silhouette loss, as illustrated in Figure~\ref{fig:eval-ablation-masks}. When only hand masks are used, the object region is empty, making the $\mathcal{L}_{\mathrm{cov}}$ term zero and $\mathcal{L}_{\mathrm{out}}$ pushes points to lie within the hand mask (left). Conversely, when only object masks are used, $\mathcal{L}_{\mathrm{out}}$ incorrectly penalizes object points that project outside the visible object mask, including points occluded by the hand (middle). Using both masks addresses these limitations by ensuring the projected points cover the visible region while being inside the masks (right).

\subsubsection{Silhouette Loss}
\label{subsubsec:ablation-silhouette Loss}
% given the importance of both masks, we next analyze the importance of silhouette loss
% table 1 shows the importance of silhouette loss
Given the complementary roles of the hand and object masks, we next investigate the contribution of the silhouette loss. Table~\ref{tab:pose_estimation} shows that incorporating the silhouette loss improves or maintains performance on all but one metric. This suggests that explicitly constraining the projected object geometry using hand and object masks provides useful supervision beyond the refiner's existing training objective. 

\begin{figure}[!tb]
  \centering
  \includegraphics[width=\columnwidth]{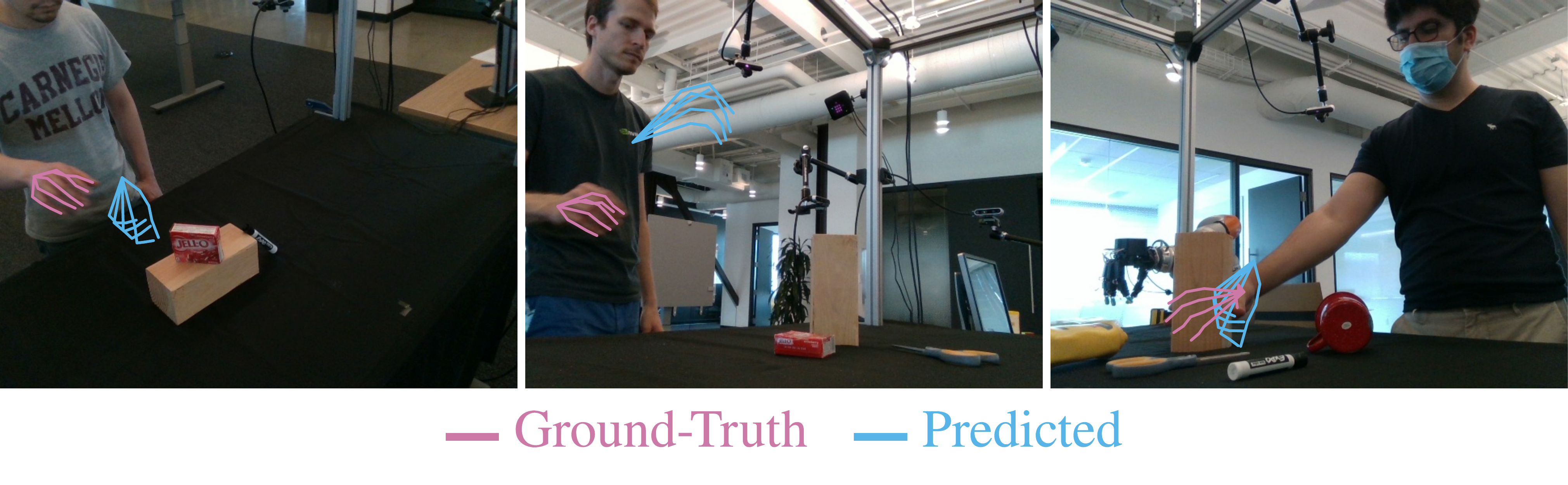}
  \caption{A state-of-the-art hand pose estimator~\cite{zhu2026wuji_mint} struggles to consistently predict accurate wrist and hand poses, which could degrade model performance when used as inputs.}
  \label{fig:eval-ablation-wrist-hand-pose}
\end{figure}

\subsubsection{Wrist and Hand Poses}
\label{subsubsec:ablation-wrist-hand-poses}
% we also explored adding other inputs on top of both masks - wrist and hand poses but find worse performance
% talk about how those are fed in
% talk about results from table 2. 
We also explore whether explicit hand geometry provides additional benefits beyond hand and object masks. Specifically, we evaluate two variants that augment the refiner's input with wrist pose or hand joint positions while retaining both hand and object masks. Both variants use hand pose predictions from MINT~\cite{zhu2026wuji_mint}, a state-of-the-art hand pose estimator. For each object pose hypothesis, we transform the predicted wrist pose or the 21 hand joints into the hypothesized object coordinate frame. A lightweight MLP then encodes this information and adds it to every refiner feature token. 

As shown in Table~\ref{tab:pose_ablation}, neither variant improves the baseline using both masks, suggesting that hand and object masks offer a simpler yet more effective signal. Additionally, we find that MINT~\cite{zhu2026wuji_mint} produces inconsistent wrist and hand pose predictions, potentially introducing noise that limits the benefits of incorporating explicit hand geometry.

\section{Conclusion}
\label{sec:conclusion}
% key insight
% describe new approach
% describe the new silhouette loss
% describe how our method performs against SOTA 
In this work, we show that hand masks, traditionally used to identify occlusions, can also provide valuable cues for 6-DoF pose estimation and tracking of hand-held tools. Building on this insight, we propose a render-and-compare approach conditioned on hand and object masks. We further introduce a novel differentiable silhouette loss that encourages the projected object to cover the visible regions in the observed image while remaining within the union of the masks. Experiments on DexYCB, HO3Dv3, and our newly collected MoCap Tools dataset demonstrate improved accuracy, robustness, and temporal consistency. One limitation of our approach is its reliance on depth inputs, which may be unreliable near thin objects and occlusion boundaries.

% \paragraph{Limitations.} Our reliance on RGB-D inputs may limit robustness in scenarios where depth estimation is unreliable, especially near thin objects and occlusion boundaries. 

\section{LLM Usage Statement}
We used LLM in editing or refining author-produced content in this paper and research. 
{
    \FloatBarrier
    \small
    \bibliographystyle{ieeenat_fullname}
    \bibliography{main}

@String(CVPR= {IEEE Conf. Comput. Vis. Pattern Recog.})

@String(ICCV= {Int. Conf. Comput. Vis.})

@String(ECCV= {Eur. Conf. Comput. Vis.})

@String(CVPR  = {CVPR})

@String(ICCV  = {ICCV})

@String(ECCV  = {ECCV})

@ARTICLE{Damen2021EPIC-KITCHENS,
   title={The EPIC-KITCHENS Dataset: Collection, Challenges and Baselines},
   author={Damen, Dima and Doughty, Hazel and Farinella, Giovanni Maria  and Fidler, Sanja and 
           Furnari, Antonino and Kazakos, Evangelos and Moltisanti, Davide and Munro, Jonathan 
           and Perrett, Toby and Price, Will and Wray, Michael},
   journal={IEEE Transactions on Pattern Analysis and Machine Intelligence (TPAMI)},
   year={2021},
   volume={43},
   number={11},
   pages={4125-4141},
   doi={10.1109/TPAMI.2020.2991965}
}

@InProceedings{perrett2025hdepic,
  author    = {Perrett, Toby and Darkhalil, Ahmad and Sinha, Saptarshi and Emara, Omar and Pollard, Sam and Parida, Kranti and Liu, Kaiting and Gatti, Prajwal and Bansal, Siddhant and Flanagan, Kevin and Chalk, Jacob and Zhu, Zhifan and Guerrier, Rhodri and Abdelazim, Fahd and Zhu, Bin and Moltisanti, Davide and Wray, Michael and Doughty, Hazel and Damen, Dima},
  title     = {HD-EPIC: A Highly-Detailed Egocentric Video Dataset},
  booktitle = {Proceedings of the IEEE/CVF Conference on Computer Vision and Pattern Recognition (CVPR)},
  year      = {2025},
  month     = {June}
}

@InProceedings{Grauman2022Ego4D,
    author    = {Grauman, Kristen and Westbury, Andrew and Byrne, Eugene and Chavis, Zachary and Furnari, Antonino and Girdhar, Rohit and Hamburger, Jackson and Jiang, Hao and Liu, Miao and Liu, Xingyu and Martin, Miguel and Nagarajan, Tushar and Radosavovic, Ilija and Ramakrishnan, Santhosh Kumar and Ryan, Fiona and Sharma, Jayant and Wray, Michael and Xu, Mengmeng and Xu, Eric Zhongcong and Zhao, Chen and Bansal, Siddhant and Batra, Dhruv and Cartillier, Vincent and Crane, Sean and Do, Tien and Doulaty, Morrie and Erapalli, Akshay and Feichtenhofer, Christoph and Fragomeni, Adriano and Fu, Qichen and Gebreselasie, Abrham and Gonz\'alez, Cristina and Hillis, James and Huang, Xuhua and Huang, Yifei and Jia, Wenqi and Khoo, Weslie and Kol\'a\v{r}, J\'achym and Kottur, Satwik and Kumar, Anurag and Landini, Federico and Li, Chao and Li, Yanghao and Li, Zhenqiang and Mangalam, Karttikeya and Modhugu, Raghava and Munro, Jonathan and Murrell, Tullie and Nishiyasu, Takumi and Price, Will and Ruiz, Paola and Ramazanova, Merey and Sari, Leda and Somasundaram, Kiran and Southerland, Audrey and Sugano, Yusuke and Tao, Ruijie and Vo, Minh and Wang, Yuchen and Wu, Xindi and Yagi, Takuma and Zhao, Ziwei and Zhu, Yunyi and Arbel\'aez, Pablo and Crandall, David and Damen, Dima and Farinella, Giovanni Maria and Fuegen, Christian and Ghanem, Bernard and Ithapu, Vamsi Krishna and Jawahar, C. V. and Joo, Hanbyul and Kitani, Kris and Li, Haizhou and Newcombe, Richard and Oliva, Aude and Park, Hyun Soo and Rehg, James M. and Sato, Yoichi and Shi, Jianbo and Shou, Mike Zheng and Torralba, Antonio and Torresani, Lorenzo and Yan, Mingfei and Malik, Jitendra},
    title     = {Ego4D: Around the World in 3,000 Hours of Egocentric Video},
    booktitle = {Proceedings of the IEEE/CVF Conference on Computer Vision and Pattern Recognition (CVPR)},
    month     = {June},
    year      = {2022},
    pages     = {18995-19012}
}

@misc{Hoque2025egodex,
      title={EgoDex: Learning Dexterous Manipulation from Large-Scale Egocentric Video}, 
      author={Ryan Hoque and Peide Huang and David J. Yoon and Mouli Sivapurapu and Jian Zhang},
      year={2025},
      eprint={2505.11709},
      archivePrefix={arXiv},
      primaryClass={cs.CV},
      url={https://arxiv.org/abs/2505.11709}, 
}

@InProceedings{foundationposewen2024,
author        = {Bowen Wen and Wei Yang and Jan Kautz and Stan Birchfield},
title         = {{FoundationPose}: Unified 6D Pose Estimation and Tracking of Novel Objects},
booktitle     = {CVPR},
year          = {2024},
}

@InProceedings{bundlesdfwen2023,
author        = {Bowen Wen and Jonathan Tremblay and Valts Blukis and Stephen Tyree and Thomas M\"{u}ller and Alex Evans and Dieter Fox and Jan Kautz and Stan Birchfield},
title         = {{BundleSDF}: {N}eural 6-{DoF} Tracking and {3D} Reconstruction of Unknown Objects},
booktitle     = {CVPR},
year          = {2023},
}

@InProceedings{Liu_2022_hoi4d,
    author    = {Liu, Yunze and Liu, Yun and Jiang, Che and Lyu, Kangbo and Wan, Weikang and Shen, Hao and Liang, Boqiang and Fu, Zhoujie and Wang, He and Yi, Li},
    title     = {HOI4D: A 4D Egocentric Dataset for Category-Level Human-Object Interaction},
    booktitle = {Proceedings of the IEEE/CVF Conference on Computer Vision and Pattern Recognition (CVPR)},
    month     = {June},
    year      = {2022},
    pages     = {21013-21022}
}

@article{liu2023groundingdino,
  title={Grounding dino: Marrying dino with grounded pre-training for open-set object detection},
  author={Liu, Shilong and Zeng, Zhaoyang and Ren, Tianhe and Li, Feng and Zhang, Hao and Yang, Jie and Li, Chunyuan and Yang, Jianwei and Su, Hang and Zhu, Jun and others},
  journal={arXiv preprint arXiv:2303.05499},
  year={2023}
}

@article{ravi2024sam2,
  title={SAM 2: Segment Anything in Images and Videos},
  author={Ravi, Nikhila and Gabeur, Valentin and Hu, Yuan-Ting and Hu, Ronghang and Ryali, Chaitanya and Ma, Tengyu and Khedr, Haitham and R{\"a}dle, Roman and Rolland, Chloe and Gustafson, Laura and Mintun, Eric and Pan, Junting and Alwala, Kalyan Vasudev and Carion, Nicolas and Wu, Chao-Yuan and Girshick, Ross and Doll{\'a}r, Piotr and Feichtenhofer, Christoph},
  journal={arXiv preprint arXiv:2408.00714},
  url={https://arxiv.org/abs/2408.00714},
  year={2024}
}

@inproceedings{labbe2022megapose,
    title     = {MegaPose: 6D Pose Estimation of Novel Objects via Render \& Compare},
    author    = {Labb\'e, Yann and Manuelli, Lucas and Mousavian, Arsalan and Tyree, Stephen and Birchfield, Stan and Tremblay, Jonathan and Carpentier, Justin and Aubry, Mathieu and Fox, Dieter and Sivic, Josef},
    booktitle = {Proceedings of the 6th Conference on Robot Learning (CoRL)},
    year      = {2022},
  }

@article{
  ornek2024foundpose,
  author    = {{\"O}rnek, Evin P{\i}nar and Labb\'e, Yann and Tekin, Bugra and Ma, Lingni and Keskin, Cem and Forster, Christian and Hoda{\v{n}}, Tom{\'a}{\v{s}}},
  title     = {FoundPose: Unseen Object Pose Estimation with Foundation Features}, 
  journal   = {European Conference on Computer Vision (ECCV)},
  year      = {2024},
}

@inproceedings{ponimatkin2025freepose,
      title={{{6D}} {{Object}} {{Pose}} {{Tracking}} in {{Internet}} {{Videos}} for {{Robotic}} {{Manipulation}}},
      author={Georgy Ponimatkin and Martin C{\'\i}fka and Tom\'{a}\v{s} Sou\v{c}ek and M{\'e}d{\'e}ric Fourmy and Yann Labb{\'e} and Vladimir Petrik and Josef Sivic},
      booktitle={The Thirteenth International Conference on Learning Representations},
      year={2025},
}

@InProceedings{Wang_2025_unihope,
    author    = {Wang, Yinqiao and Xu, Hao and Heng, Pheng-Ann and Fu, Chi-Wing},
    title     = {UniHOPE: A Unified Approach for Hand-Only and Hand-Object Pose Estimation},
    booktitle = {Proceedings of the IEEE/CVF Conference on Computer Vision and Pattern Recognition (CVPR)},
    month     = {June},
    year      = {2025},
    pages     = {12231-12241}
}

@INPROCEEDINGS{hampali2020ho3dv3,
	      title={HOnnotate: A method for 3D Annotation of Hand and Object Poses},
          author={Shreyas Hampali and Mahdi Rad and Markus Oberweger and Vincent Lepetit},
          booktitle = {CVPR},
      year = {2020}
         }

@INPROCEEDINGS{chao2021dexycb,
  author    = {Yu-Wei Chao and Wei Yang and Yu Xiang and Pavlo Molchanov and Ankur Handa and Jonathan Tremblay and Yashraj S. Narang and Karl {Van Wyk} and Umar Iqbal and Stan Birchfield and Jan Kautz and Dieter Fox},
  booktitle = {IEEE/CVF Conference on Computer Vision and Pattern Recognition (CVPR)},
  title     = {{DexYCB}: A Benchmark for Capturing Hand Grasping of Objects},
  year      = {2021},
}

@article{wen2025foundationstereo,
  title={FoundationStereo: Zero-Shot Stereo Matching},
  author={Bowen Wen and Matthew Trepte and Joseph Aribido and Jan Kautz and Orazio Gallo and Stan Birchfield},
  journal={arXiv},
  year={2025}
}

@inproceedings{lee2025any6d,
    title     = {{Any6D}: Model-free 6D Pose Estimation of Novel Objects},
    author    = {Lee, Taeyeop and Wen, Bowen and Kang, Minjun and Kang, Gyuree and Kweon, In So and Yoon, Kuk-Jin},
    booktitle = {Proceedings of the Computer Vision and Pattern Recognition Conference (CVPR)},
    year      = {2025},
}

@InProceedings{Moon_2024_genflow,
    author    = {Moon, Sungphill and Son, Hyeontae and Hur, Dongcheol and Kim, Sangwook},
    title     = {GenFlow: Generalizable Recurrent Flow for 6D Pose Refinement of Novel Objects},
    booktitle = {Proceedings of the IEEE/CVF Conference on Computer Vision and Pattern Recognition (CVPR)},
    month     = {June},
    year      = {2024},
    pages     = {10039-10049}
}

@misc{zhu2026wuji_mint,
  title  = {MINT: A Unified Model for World-Space Camera and Hand Motion
            Estimation from Scalable Egocentric Pipeline Supervision},
  author = {Zhu, Zijie and Cai, Weiren and Wang, Yizhou and Yang, Zhenjie and
            Liu, Yide and Chen, Jiahao and He, Guanqi},
  year   = {2026},
  note   = {Manuscript under review}
}

@inproceedings{xiang2018posecnn,
    Author = {Xiang, Yu and Schmidt, Tanner and Narayanan, Venkatraman and Fox, Dieter},
    Title = {PoseCNN: A Convolutional Neural Network for 6D Object Pose Estimation in Cluttered Scenes},
    Journal   = {Robotics: Science and Systems (RSS)},
    Year = {2018}
}

@article{wang2019densefusion,
  title={DenseFusion: 6D Object Pose Estimation by Iterative Dense Fusion},
  author={Wang, Chen and Xu, Danfei and Zhu, Yuke and Mart{\'\i}n-Mart{\'\i}n, Roberto and Lu, Cewu and Fei-Fei, Li and Savarese, Silvio},
  booktitle={Computer Vision and Pattern Recognition (CVPR)},
  year={2019}
}

@InProceedings{Wang_2021_GDRN,
    title     = {{GDR-Net}: Geometry-Guided Direct Regression Network for Monocular 6D Object Pose Estimation},
    author    = {Wang, Gu and Manhardt, Fabian and Tombari, Federico and Ji, Xiangyang},
    booktitle = {IEEE/CVF Conference on Computer Vision and Pattern Recognition (CVPR)},
    month     = {June},
    year      = {2021},
    pages     = {16611-16621}
}

@inproceedings{peng2019pvnet,
  title={PVNet: Pixel-wise Voting Network for 6DoF Pose Estimation},
  author={Peng, Sida and Liu, Yuan and Huang, Qixing and Zhou, Xiaowei and Bao, Hujun},
  booktitle={CVPR},
  year={2019}
}

@inproceedings{bansal2026hopformer,
  title     = {Towards in-the-wild Egocentric 3D Hand-Object Pose Estimation},
  author    = {Bansal, Siddhant and Zhu, Zhifan and Tripathi, Shashank and
               Zhao, Jiahe and Black, Michael J. and Damen, Dima},
  booktitle = {European Conference on Computer Vision (ECCV)},
  year      = {2026}
}

@article{wang20196pack,
  title={6-PACK: Category-level 6D Pose Tracker with Anchor-Based Keypoints},
  author={Wang, Chen and Mart{\'\i}n-Mart{\'\i}n, Roberto and Xu, Danfei and Lv, Jun and Lu, Cewu and Fei-Fei, Li and Savarese, Silvio and Zhu, Yuke},
  booktitle={International Conference on Robotics and Automation (ICRA)},
  year={2020}
}

@article{shin2026compose,
  author    = {Shin, Jisu and Lee, Junoh and Lee, JunGyu and Bae, Inhwan and Lee, Dohyeon and Im, Hokyun and Lee, Youngwoon and Jeon, Hae-Gon},
  title     = {ComPose: When to Trust Hands for Object Pose Tracking},
  journal   = {arXiv preprint arXiv:2605.23523},
  year      = {2026},
}

@InProceedings{Zhu_2026_rohit,
    author    = {Zhu, Zhifan and Bansal, Siddhant and Tripathi, Shashank and Damen, Dima},
    title     = {Reconstructing Objects along Hand Interaction Timelines in Egocentric Video},
    booktitle = {Proceedings of the IEEE/CVF Conference on Computer Vision and Pattern Recognition (CVPR) Workshops},
    month     = {June},
    year      = {2026},
    pages     = {3955-3965}
}

@InProceedings{He_2020_pvn3d,
author = {He, Yisheng and Sun, Wei and Huang, Haibin and Liu, Jianran and Fan, Haoqiang and Sun, Jian},
title = {PVN3D: A Deep Point-Wise 3D Keypoints Voting Network for 6DoF Pose Estimation},
booktitle = {IEEE/CVF Conference on Computer Vision and Pattern Recognition (CVPR)},
month = {June},
year = {2020}
}

@article{liu2024taco,
      title={TACO: Benchmarking Generalizable Bimanual Tool-ACtion-Object Understanding},
      author={Liu, Yun and Yang, Haolin and Si, Xu and Liu, Ling and Li, Zipeng and Zhang, Yuxiang and Liu, Yebin and Yi, Li},
      journal={arXiv preprint arXiv:2401.08399},
      year={2024}
    }

@article{banerjee2025hot3d,
  title={{HOT3D}: Hand and Object Tracking in {3D} from Egocentric Multi-View Videos},
  author={Banerjee, Prithviraj and Shkodrani, Sindi and Moulon, Pierre and Hampali, Shreyas and Han, Shangchen and Zhang, Fan and Zhang, Linguang and Fountain, Jade and Miller, Edward and Basol, Selen and Newcombe, Richard and Wang, Robert and Engel, Jakob Julian and Hodan, Tomas},
  journal={CVPR},
  year={2025}
}

@inproceedings{YangCVPR2022OakInk,
  author    = {Yang, Lixin and Li, Kailin and Zhan, Xinyu and Wu, Fei and Xu, Anran and Liu, Liu and Lu, Cewu},
  title     = {{OakInk}: A Large-Scale Knowledge Repository for Understanding Hand-Object Interaction},
  booktitle = {IEEE/CVF Conference on Computer Vision and Pattern Recognition (CVPR)},
  year      = {2022},
}

@InProceedings{Kwon_2021_h2o,
author = {Kwon, Taein and Tekin, Bugra and St\"uhmer, Jan and Bogo, Federica and Pollefeys, Marc},
title = {H2O: Two Hands Manipulating Objects for First Person Interaction Recognition},
booktitle = {Proceedings of the IEEE/CVF International Conference on Computer Vision (ICCV)},
month = {October},
year = {2021},
pages = {10138-10148}
}

@inproceedings{li2018deepim,
title     = {DeepIM: Deep Iterative Matching for 6D Pose Estimation},
author    = {Yi Li and Gu Wang and Xiangyang Ji and Yu Xiang and Dieter Fox},
booktitle = {European Conference on Computer Vision (ECCV)},
year      = {2018}
}

@inproceedings{labbe2020cosypose,
title= {CosyPose: Consistent multi-view multi-object 6D pose estimation},
author={Y. {Labbe} and J. {Carpentier} and M. {Aubry} and J. {Sivic}},
booktitle={Proceedings of the European Conference on Computer Vision (ECCV)},
year={2020}}

@misc{potamias2024wilor,
      title={WiLoR: End-to-end 3D Hand Localization and Reconstruction in-the-wild},
      author={Rolandos Alexandros Potamias and Jinglei Zhang and Jiankang Deng and Stefanos Zafeiriou},
      year={2024},
      eprint={2409.12259},
      archivePrefix={arXiv},
      primaryClass={cs.CV}
  }

@misc{carion2025sam3segmentconcepts,
      title={SAM 3: Segment Anything with Concepts},
      author={Nicolas Carion and Laura Gustafson and Yuan-Ting Hu and Shoubhik Debnath and Ronghang Hu and Didac Suris and Chaitanya Ryali and Kalyan Vasudev Alwala and Haitham Khedr and Andrew Huang and Jie Lei and Tengyu Ma and Baishan Guo and Arpit Kalla and Markus Marks and Joseph Greer and Meng Wang and Peize Sun and Roman Rädle and Triantafyllos Afouras and Effrosyni Mavroudi and Katherine Xu and Tsung-Han Wu and Yu Zhou and Liliane Momeni and Rishi Hazra and Shuangrui Ding and Sagar Vaze and Francois Porcher and Feng Li and Siyuan Li and Aishwarya Kamath and Ho Kei Cheng and Piotr Dollár and Nikhila Ravi and Kate Saenko and Pengchuan Zhang and Christoph Feichtenhofer},
      year={2025},
      eprint={2511.16719},
      archivePrefix={arXiv},
      primaryClass={cs.CV},
      url={https://arxiv.org/abs/2511.16719},
}
}

\clearpage
\setcounter{page}{1}
\clearpage
\onecolumn
\setcounter{page}{1}

\begin{center}
    \Large\textbf{\thetitle}\\
    \vspace{0.5em}
    Supplementary Material
\end{center}
\vspace{1em}
% \section{Project Website}
% Our project website is available at: \url{https://tool-pose-tracking.github.io/}.

\section{Additional Quantitative Results}
\label{sec:sup_additional_quantitative}

\subsection{DexYCB}

\begin{figure}[htbp]
    \centering
    \begin{minipage}{\textwidth}
        \centering

        % Time arrow
        \noindent
        \hspace*{0.05\linewidth}%
        \begin{minipage}{0.90\linewidth}
            \centering
            \begin{tikzpicture}
                \draw[->, line width=1.2pt]
                    (0,0) -- node[midway, above] {\small time}
                    (0.9\linewidth,0);
            \end{tikzpicture}%
        \end{minipage}\par

        \vspace{0.1cm}

        \dexycbgridrow{Ours\\(w/ silhouette loss)}{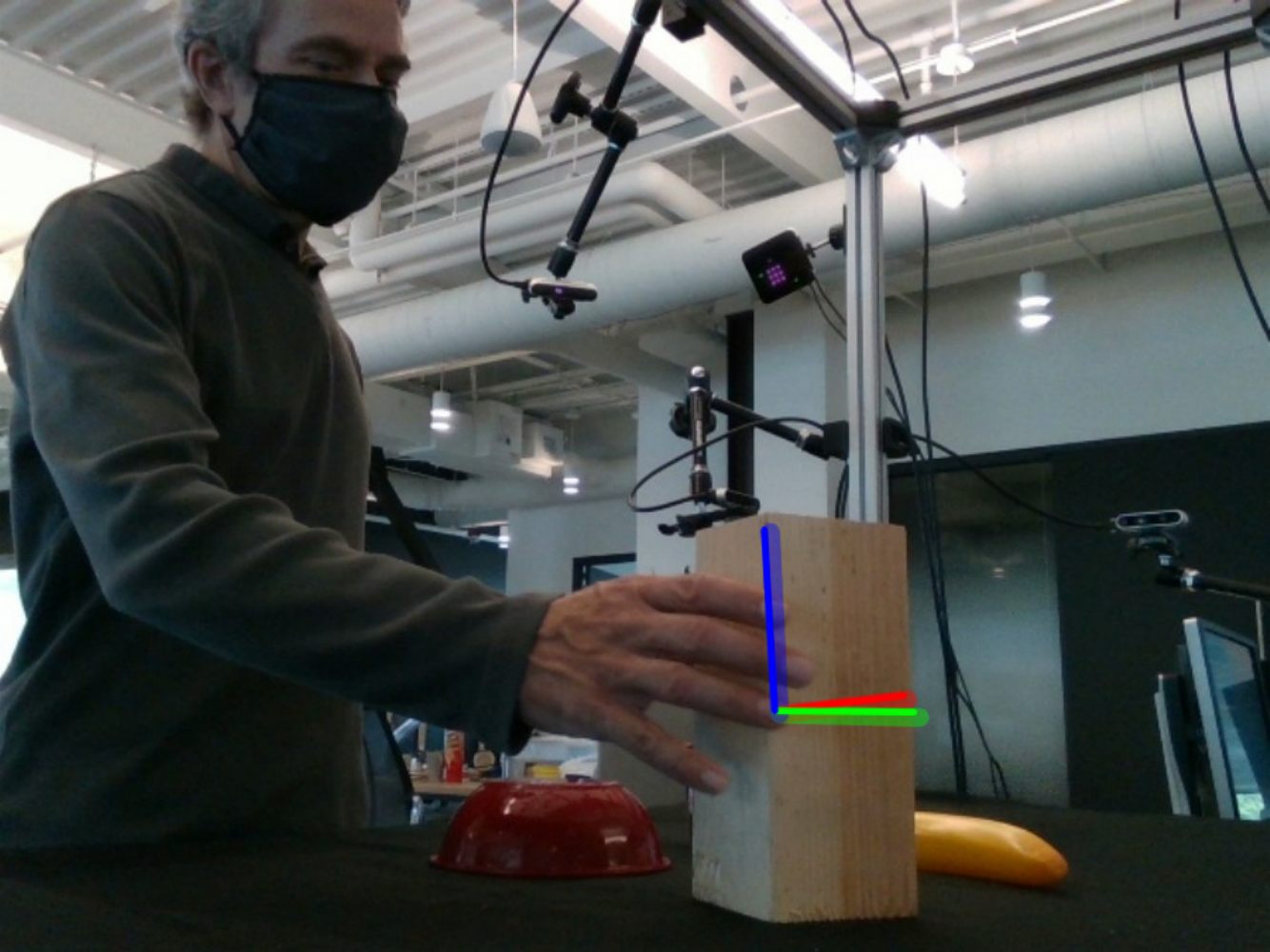}
        \vspace{0.02cm}

        \dexycbgridrow{FoundationPose\\Pretrained}{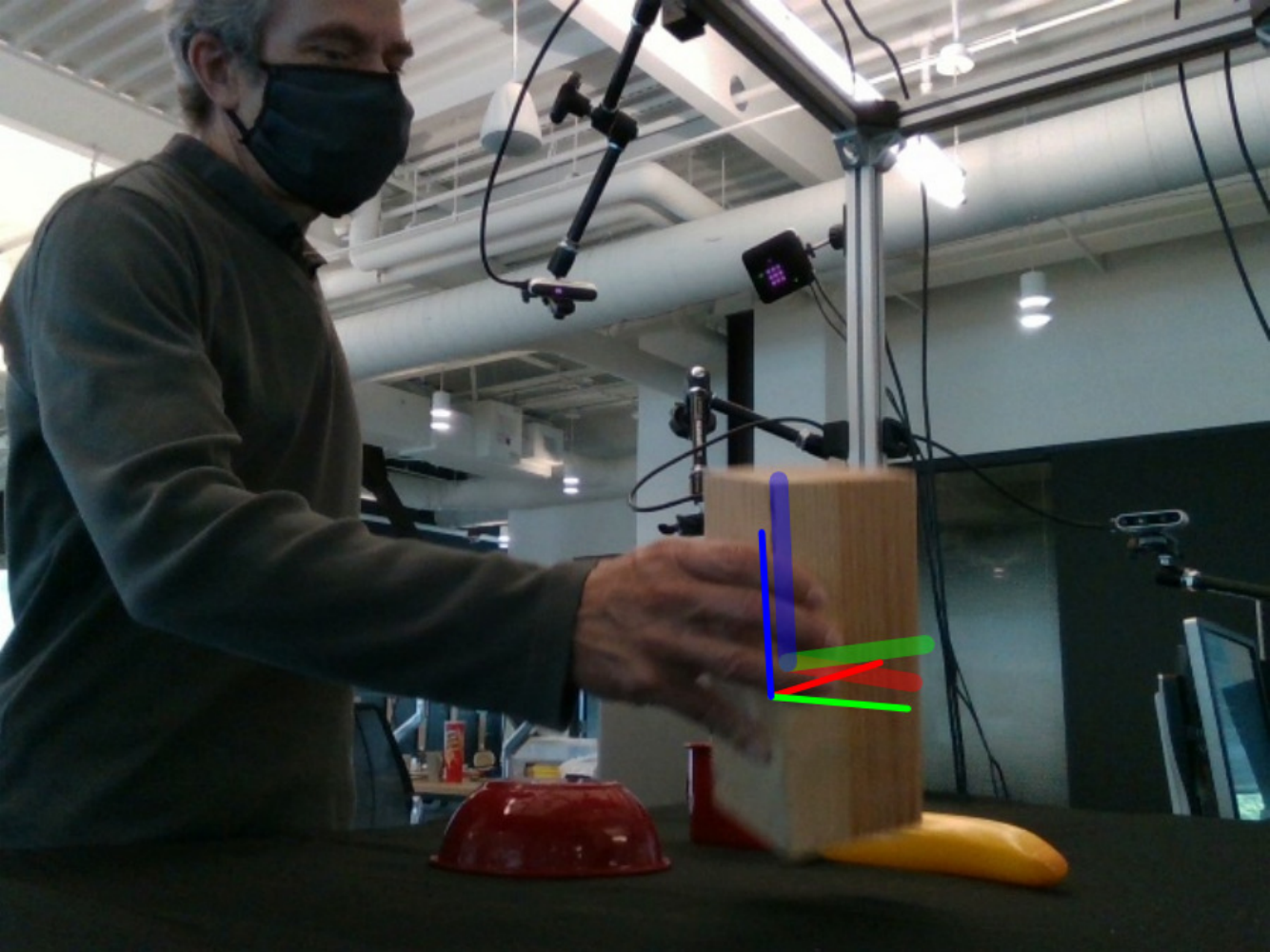}
        \vspace{0.02cm}

        \dexycbgridrow{FoundationPose\\Fine-tuned}{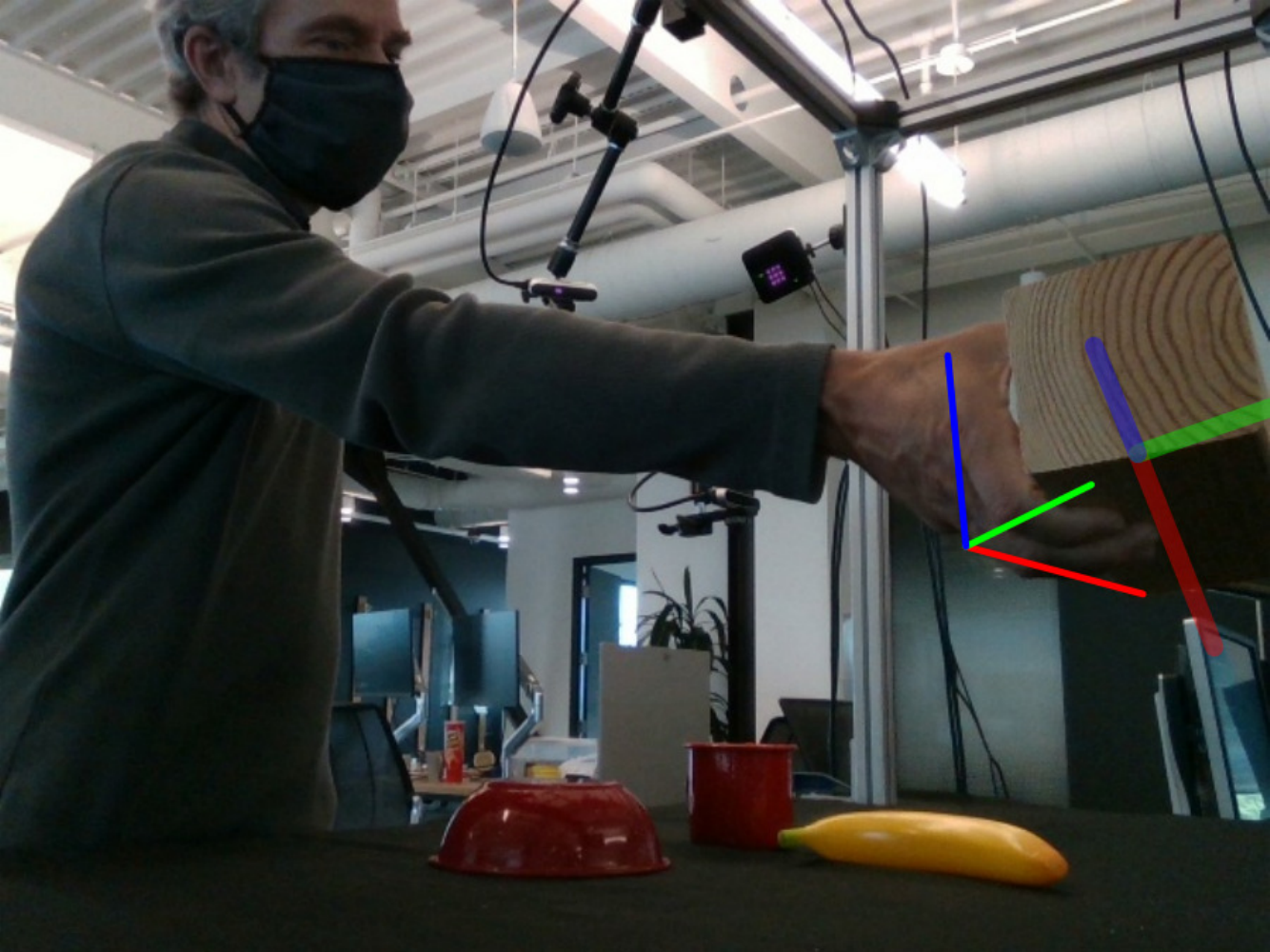}
        \vspace{0.02cm}

        \dexycbgridrow{MegaPose}{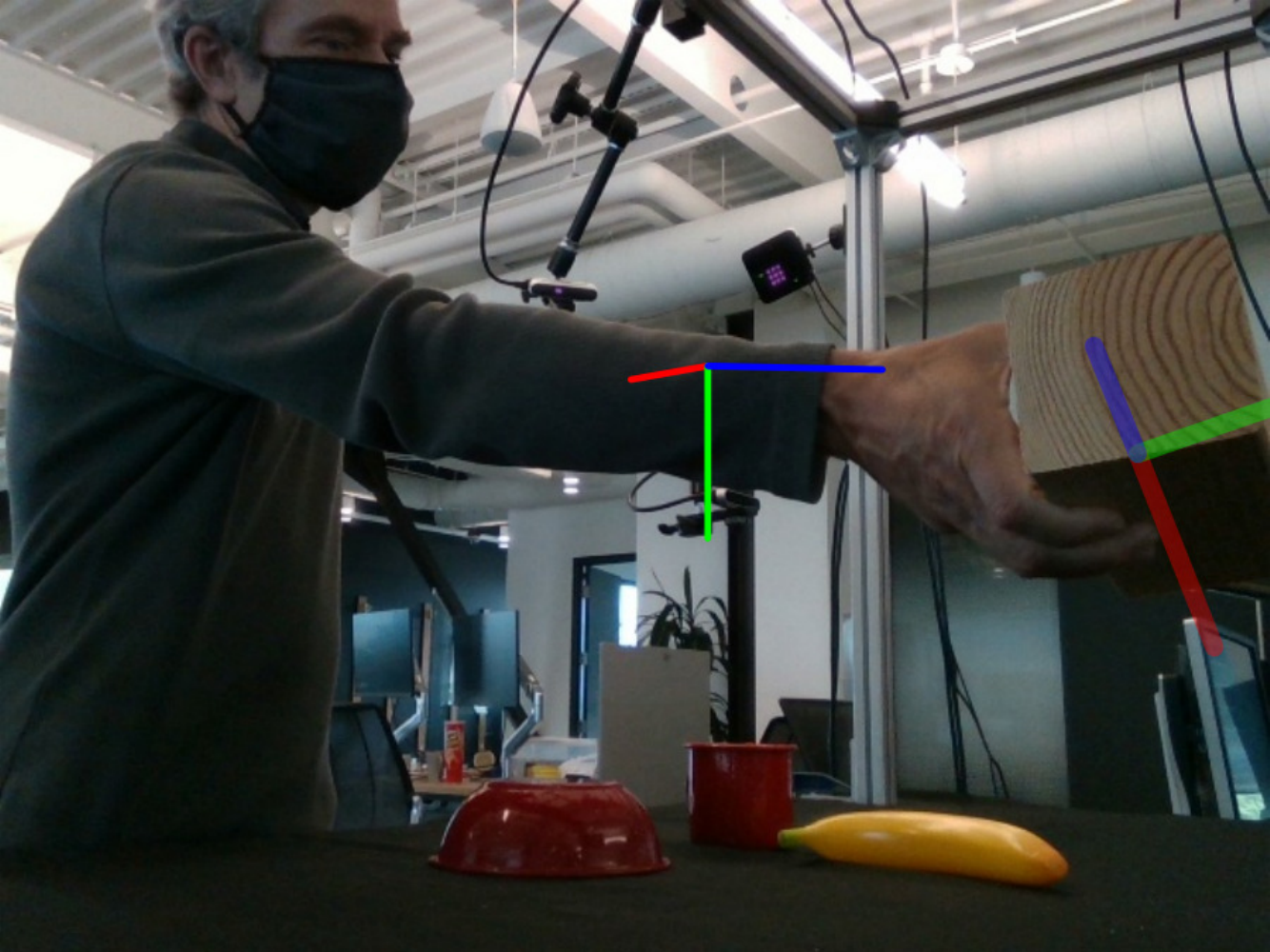}
        \vspace{0.02cm}

        \dexycbgridrow{FoundPose}{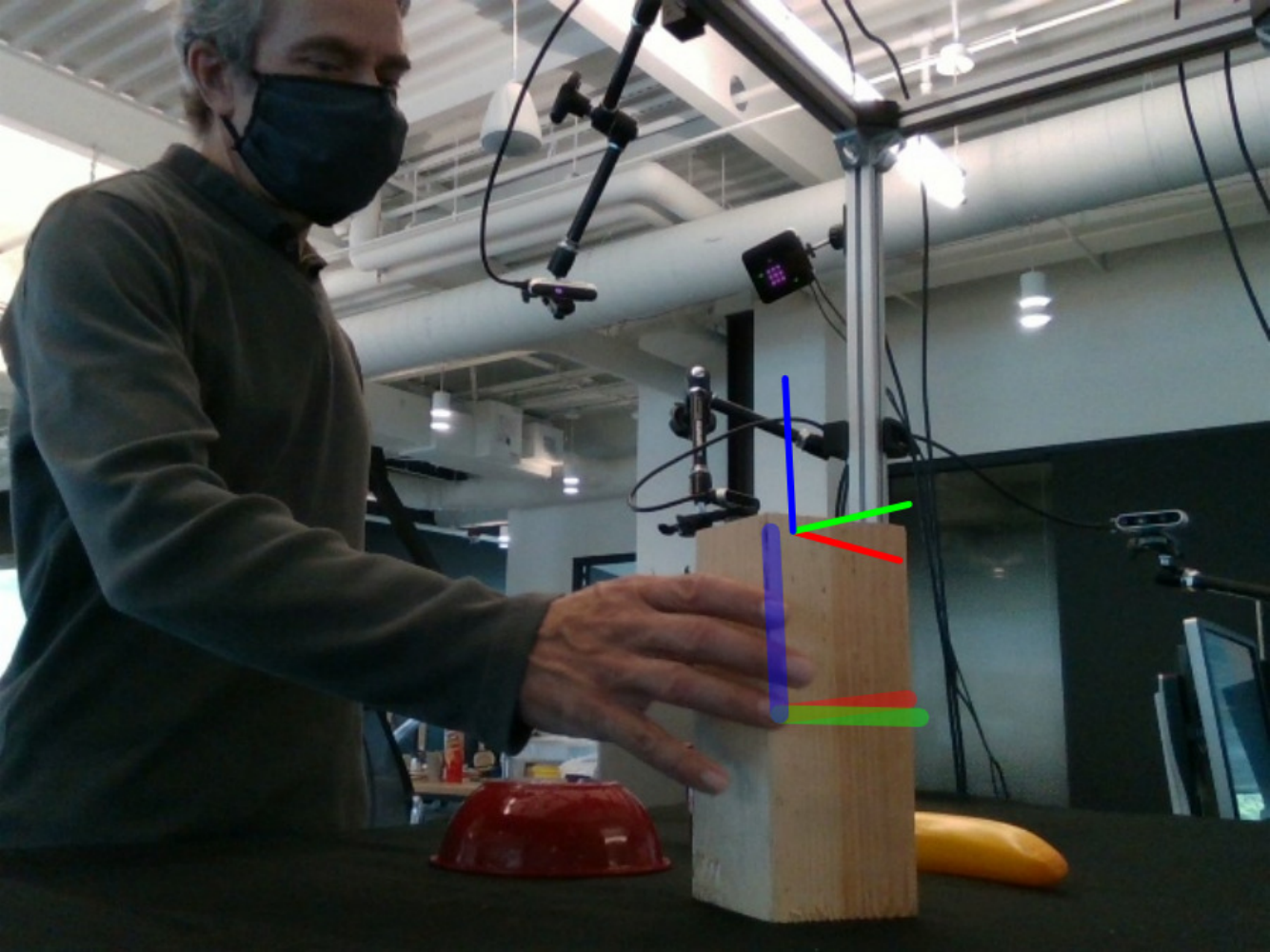}
        \vspace{0.02cm}

        \dexycbgridrow{FreePose}{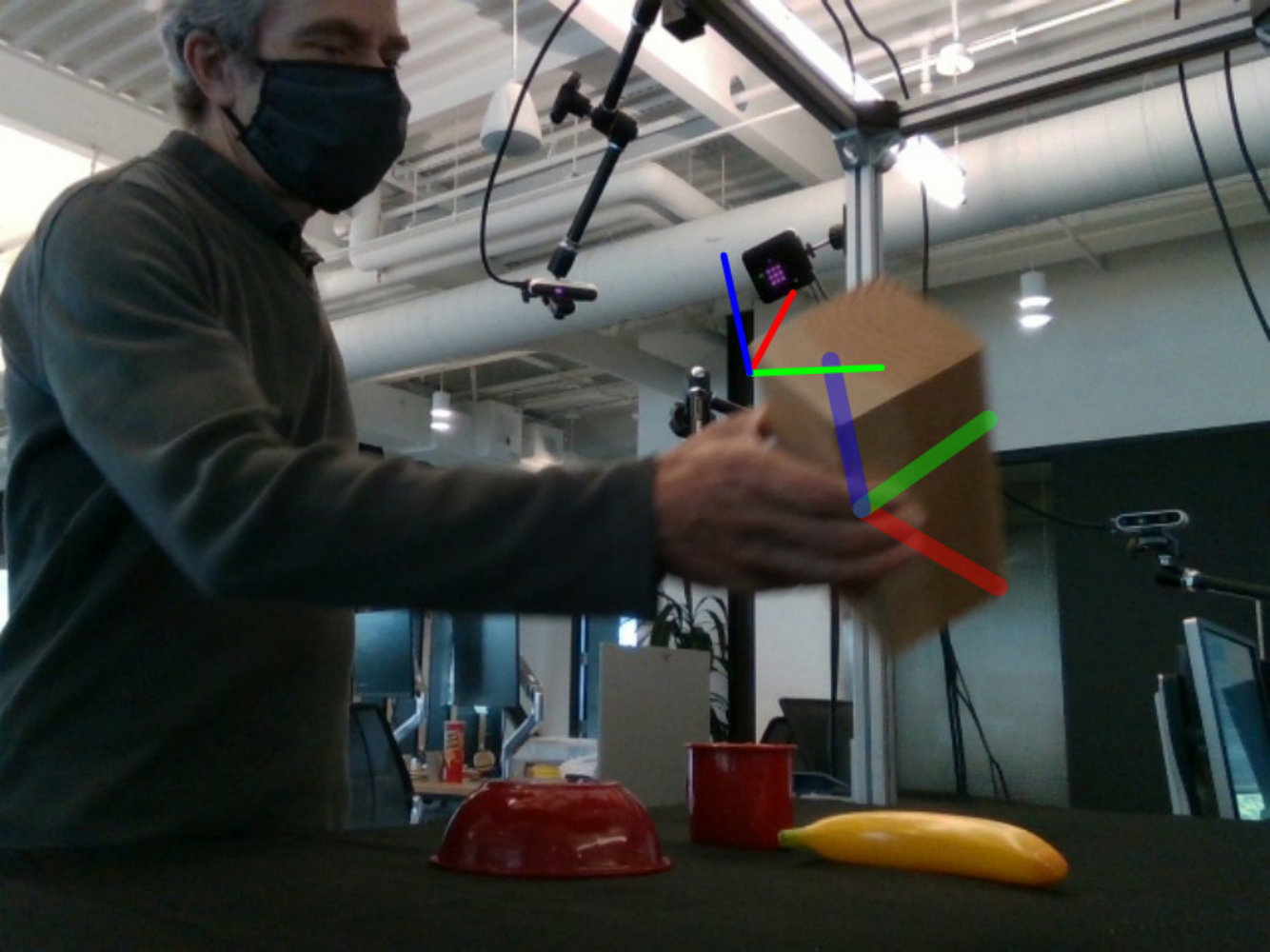}
        \vspace{0.02cm}

        \dexycbgridrow{UniHOPE}{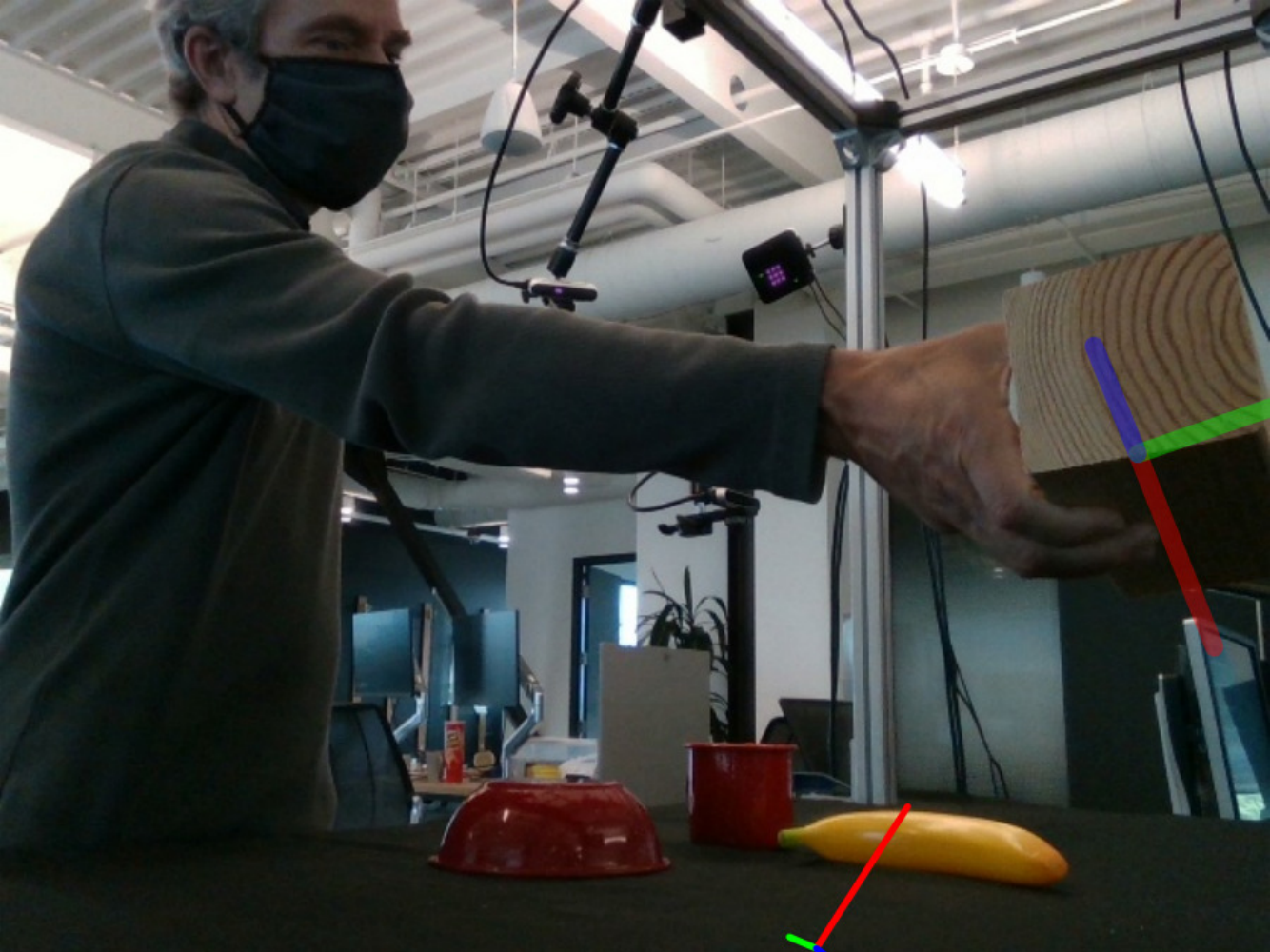}
        \vspace{0.02cm}

        \caption{Qualitative results on DexYCB across a video sequence.}
        \label{fig:sup-dexycb-qualitative-grid}

    \end{minipage}
\end{figure}

\clearpage

\subsection{HO3Dv3}

\begin{figure}[htbp]
    \centering
    \begin{minipage}{\textwidth}
        \centering

        % Time arrow
        \noindent
        \hspace*{0.05\linewidth}%
        \begin{minipage}{0.90\linewidth}
            \centering
            \begin{tikzpicture}
                \draw[->, line width=1.2pt]
                    (0,0) -- node[midway, above] {\small time}
                    (0.9\linewidth,0);
            \end{tikzpicture}%
        \end{minipage}\par

        \vspace{0.1cm}

        \hogridrow{Ours\\(w/ silhouette loss)}{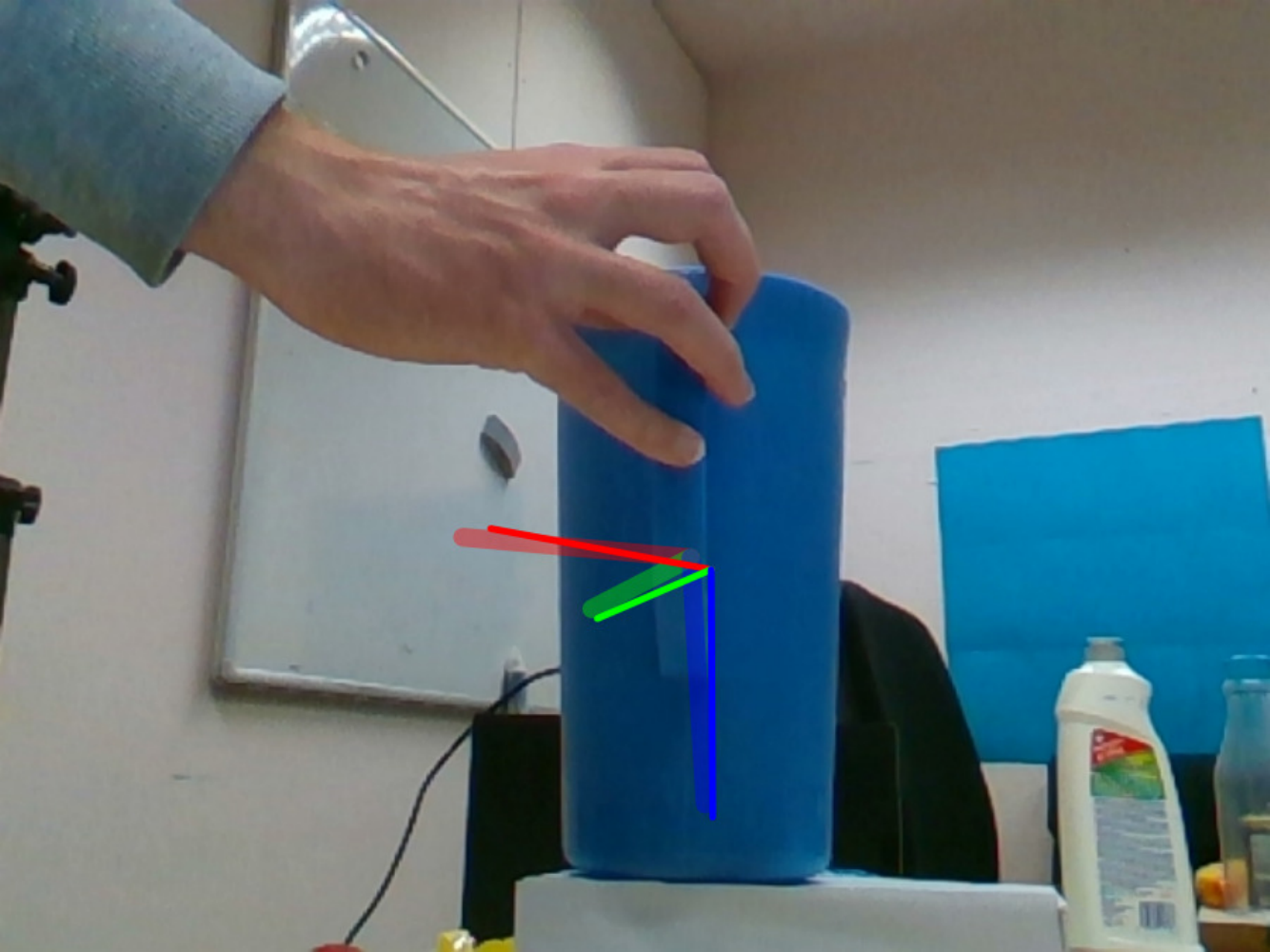}
        \vspace{0.02cm}

        \hogridrow{FoundationPose\\Pretrained}{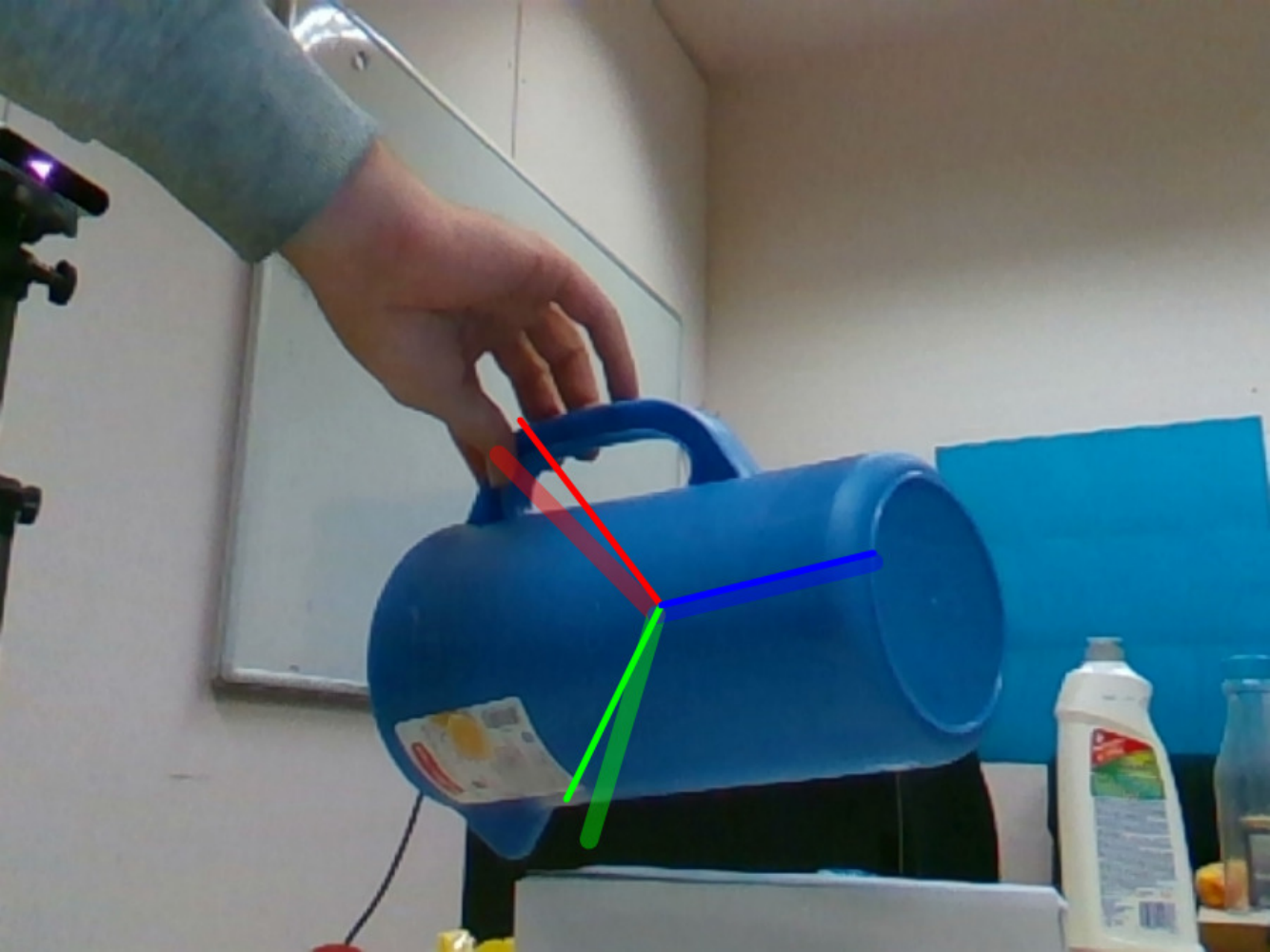}
        \vspace{0.02cm}

        \hogridrow{FoundationPose\\Fine-tuned}{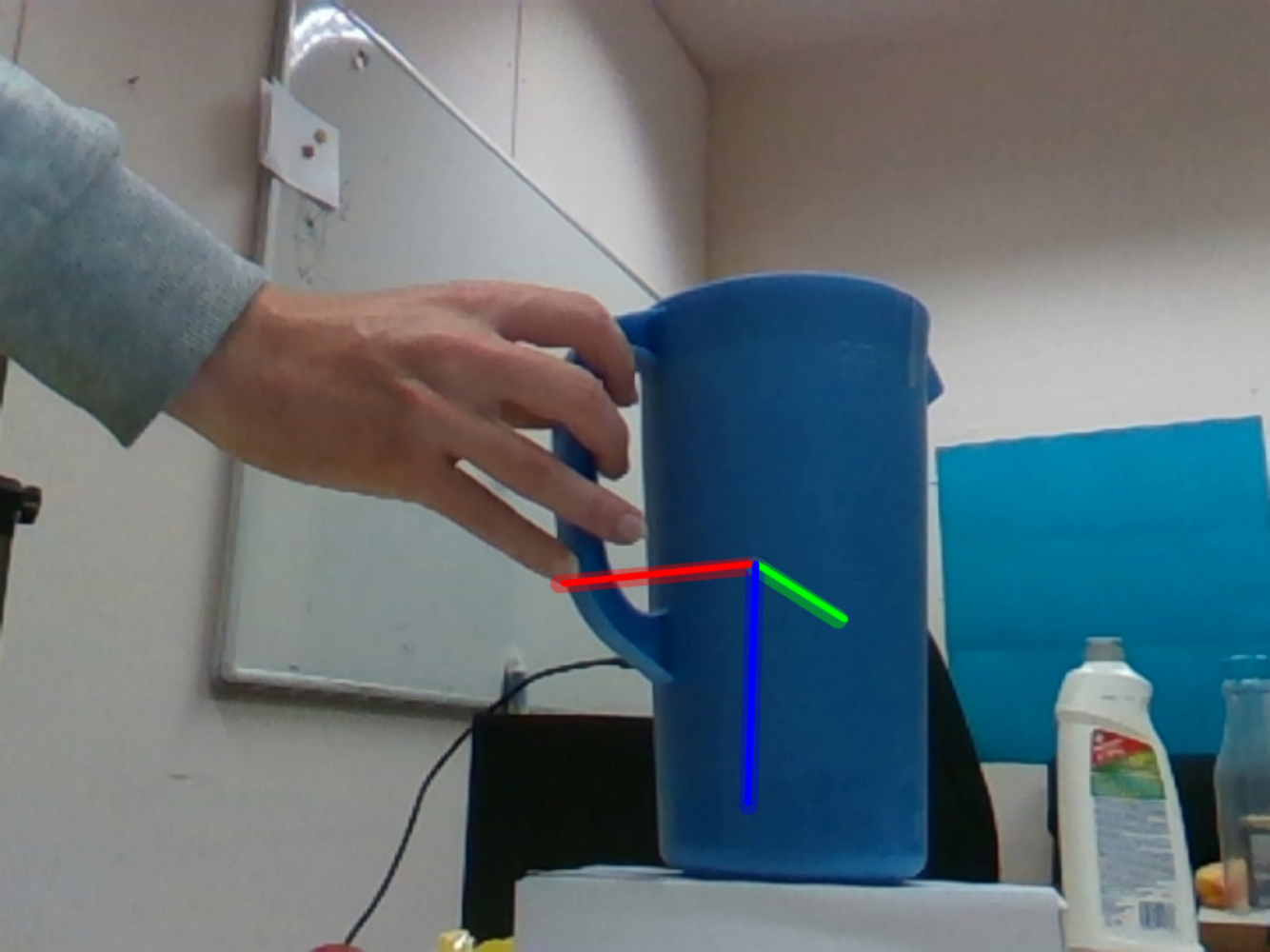}
        \vspace{0.02cm}

        \hogridrow{MegaPose}{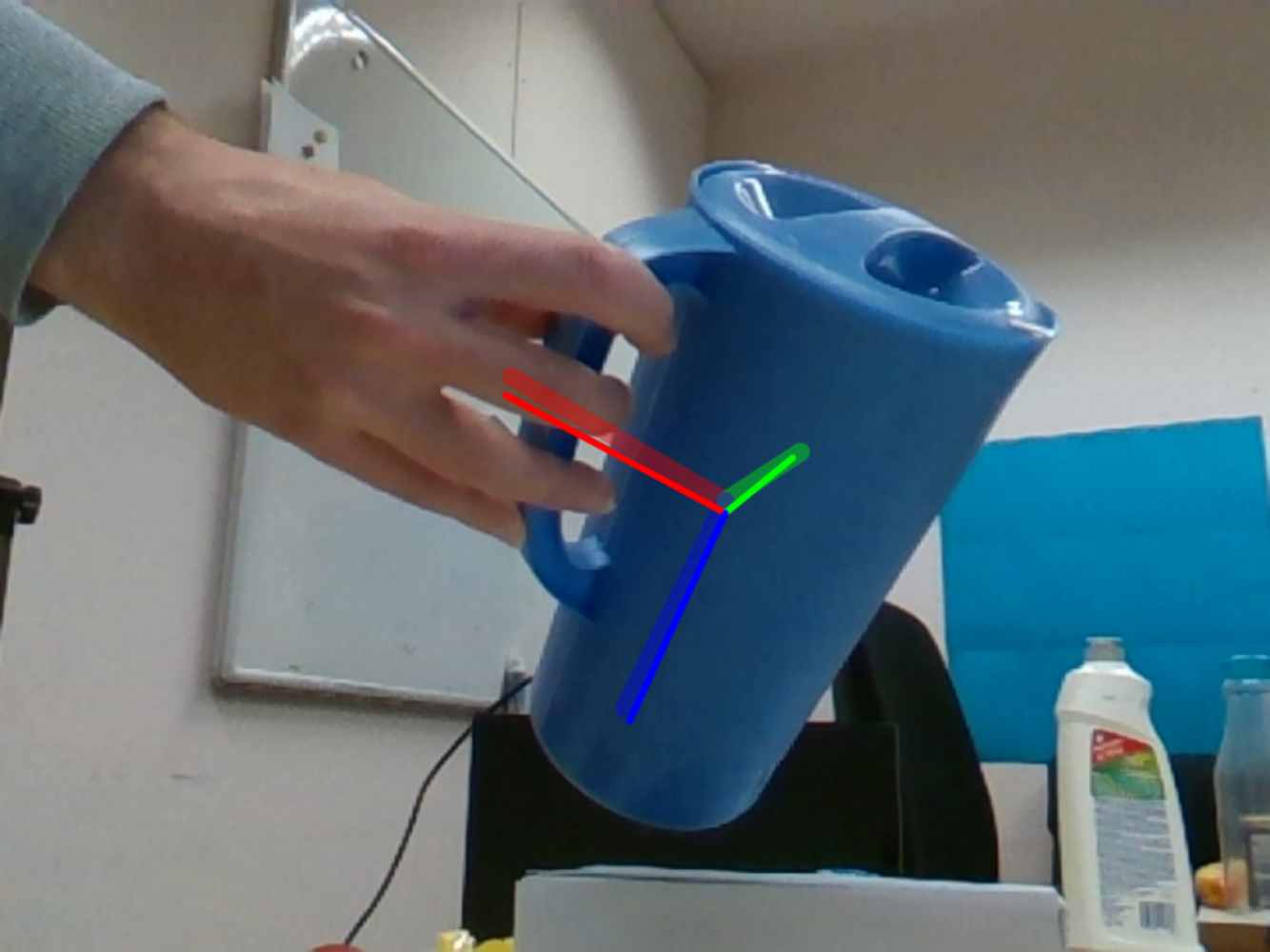}
        \vspace{0.02cm}

        \hogridrow{FoundPose}{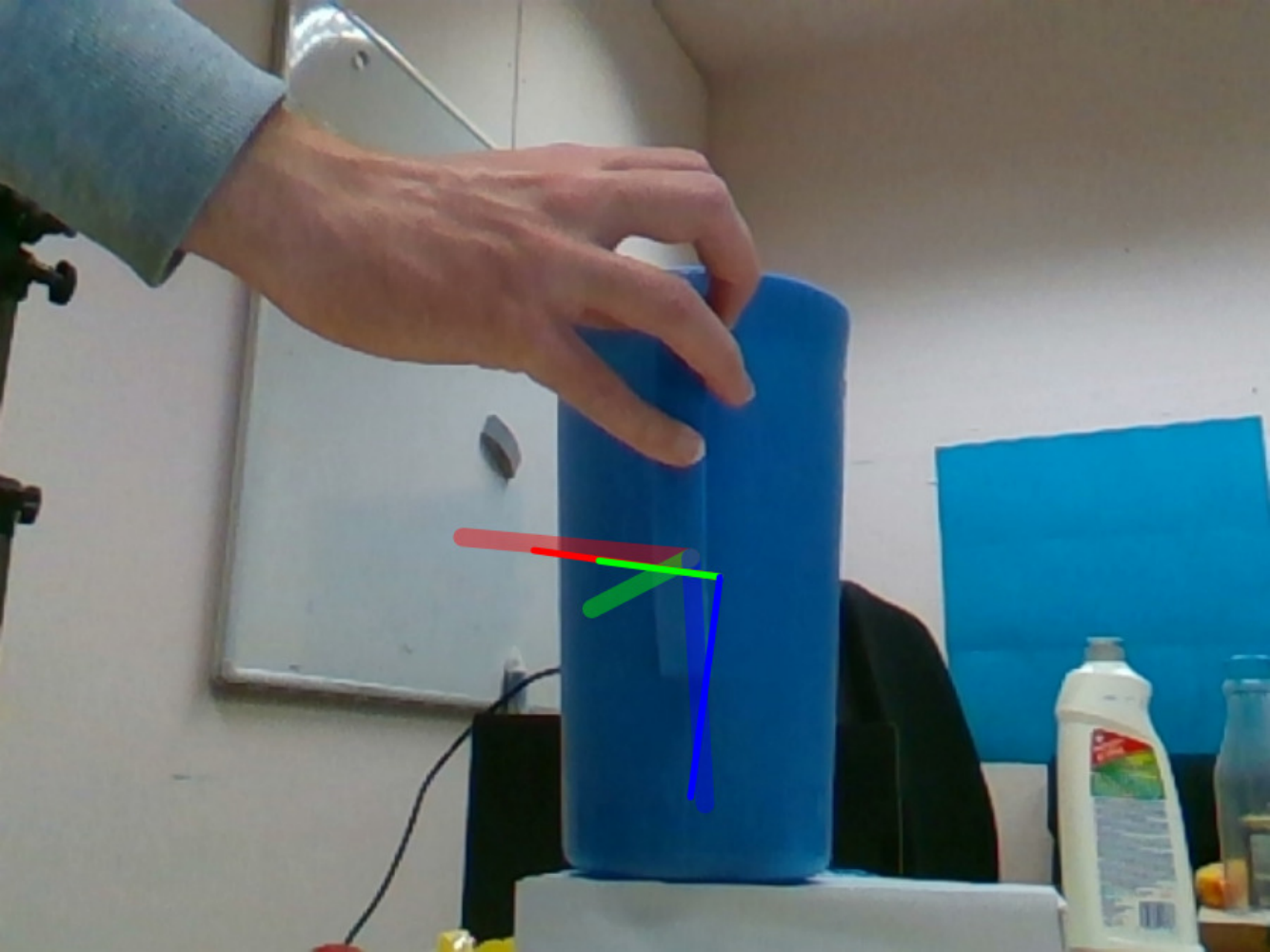}
        \vspace{0.02cm}

        \hogridrow{FreePose}{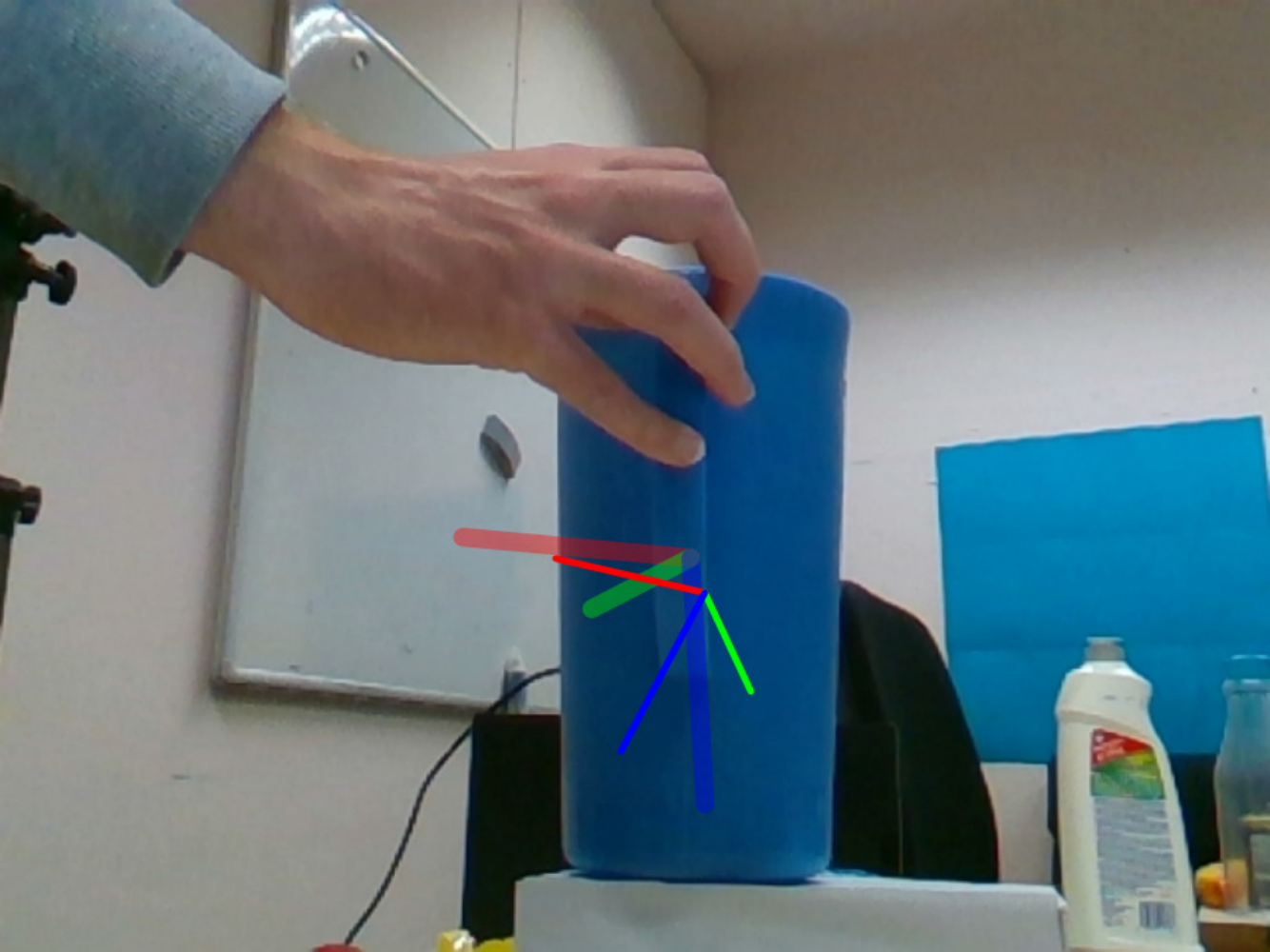}
        \vspace{0.02cm}

        \hogridrow{UniHOPE}{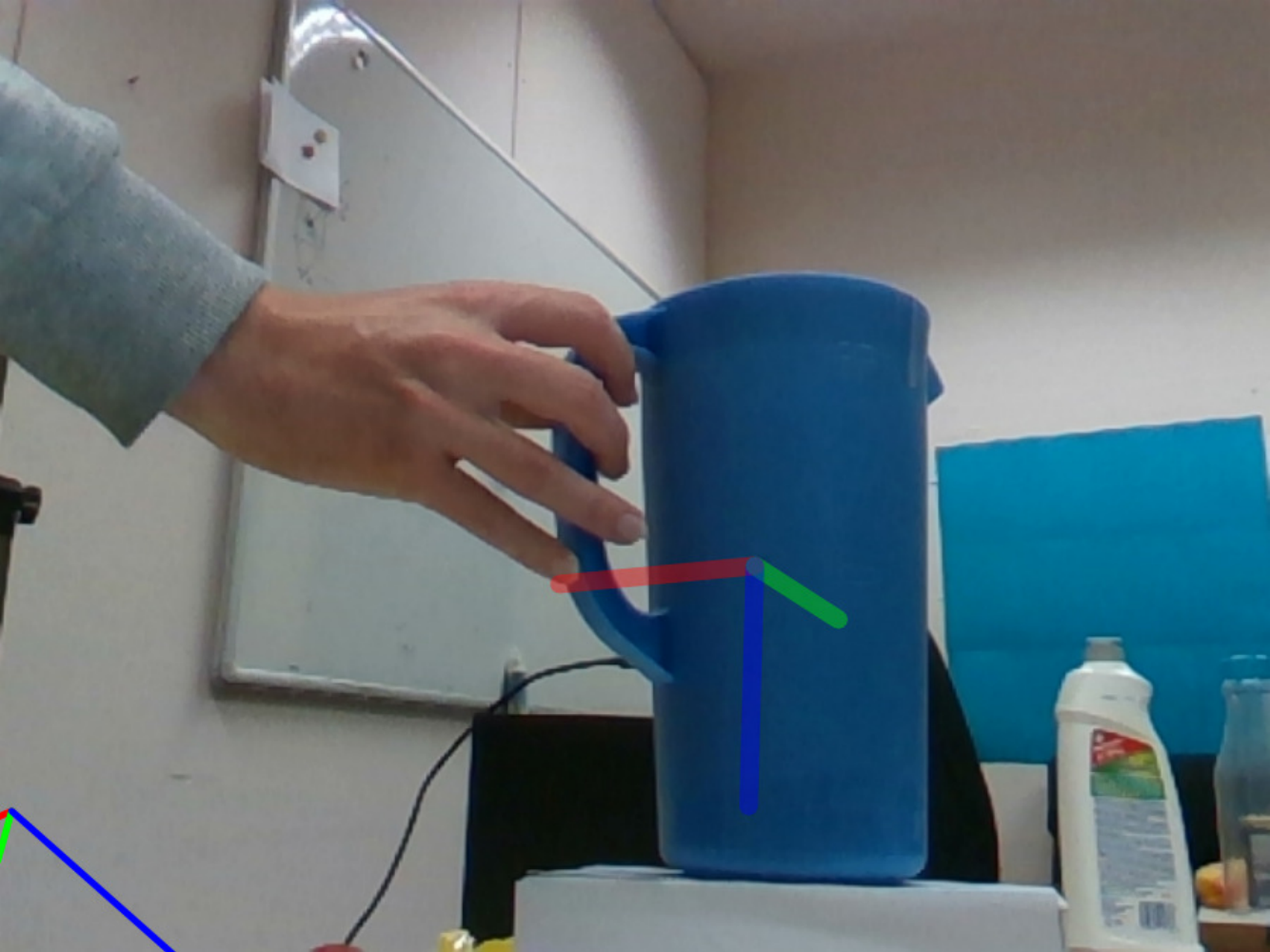}
        \vspace{0.02cm}

        \caption{Qualitative results on HO3Dv3 across a video sequence.}
        \label{fig:sup-ho3dv3-qualitative-grid}

    \end{minipage}
\end{figure}

\clearpage

\subsection{Random Frame Sampling Across All Datasets}

\begin{figure}[htbp]
    \centering
    \begin{minipage}{\textwidth}
        \centering

        % Time arrow
        \noindent
        \hspace*{0.05\linewidth}%
        \begin{minipage}{0.90\linewidth}
            \centering
            \begin{tikzpicture}
                \draw[->, line width=1.2pt]
                    (0,0) -- node[midway, above] {\small time}
                    (0.9\linewidth,0);
            \end{tikzpicture}%
        \end{minipage}\par

        \vspace{0.1cm}

        \randomgridrow{Ours\\(w/ silhouette loss)}{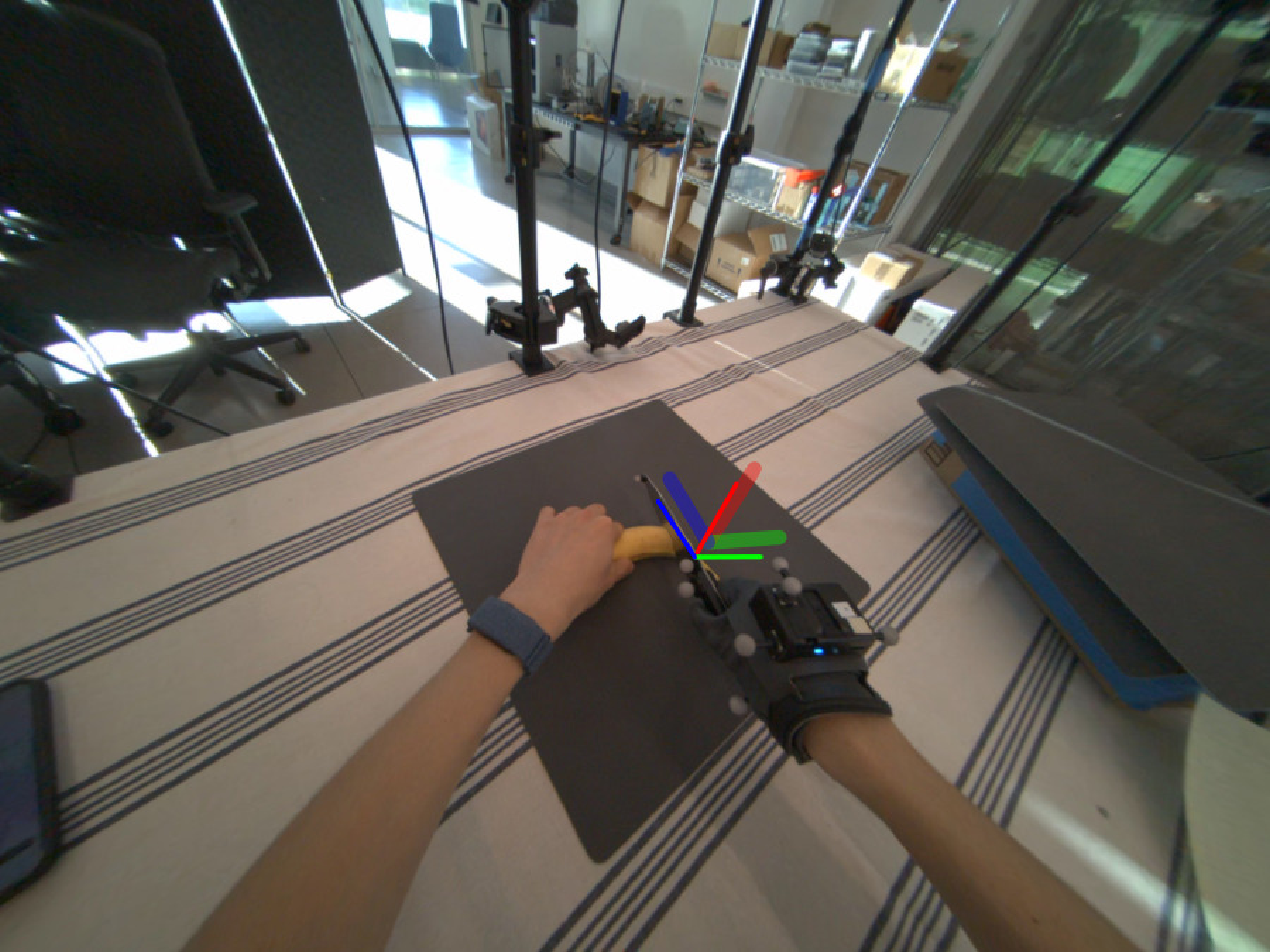}
        \vspace{0.02cm}

        \randomgridrow{FoundationPose\\Pretrained}{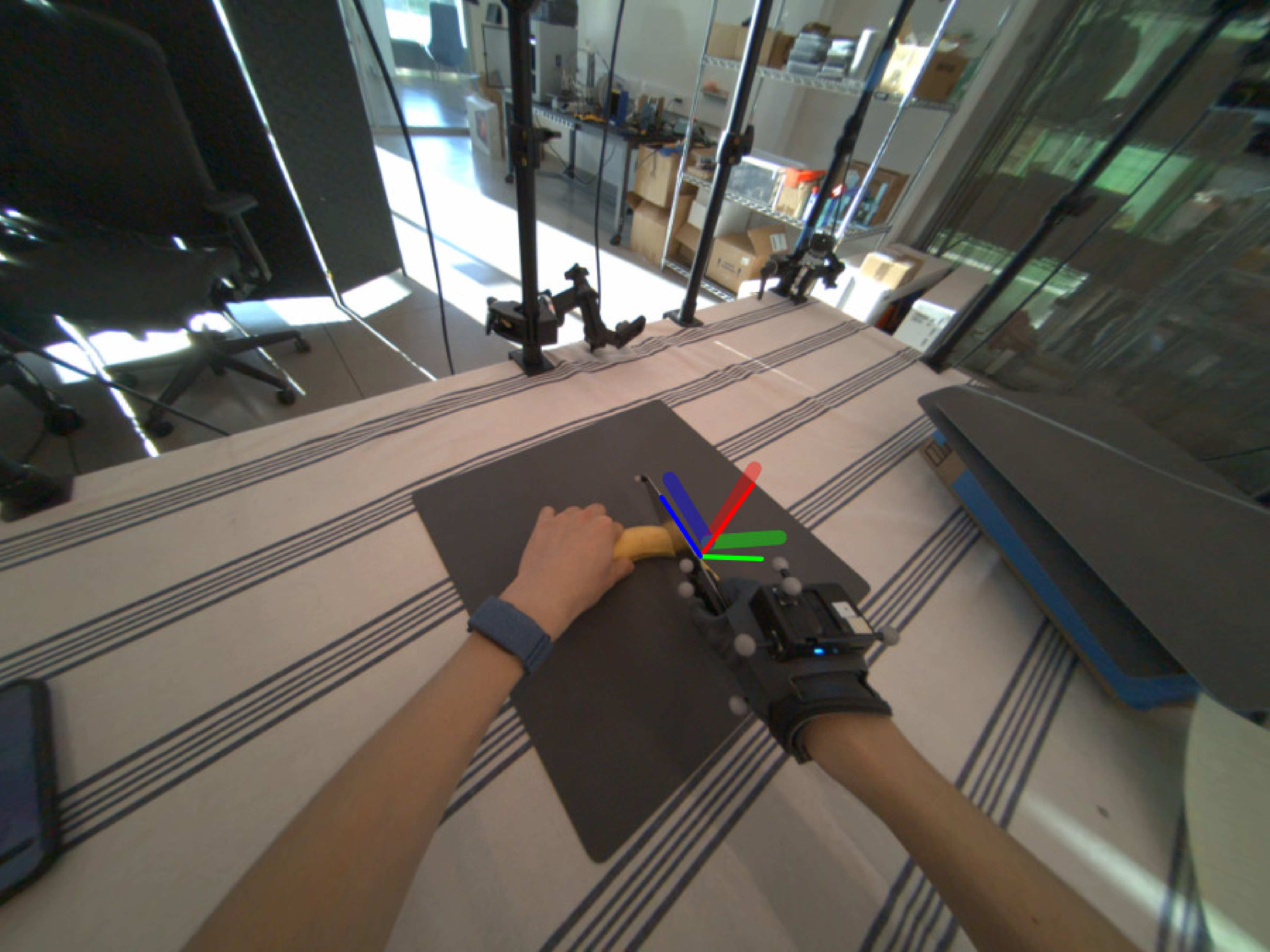}
        \vspace{0.02cm}

        \randomgridrow{FoundationPose\\Fine-tuned}{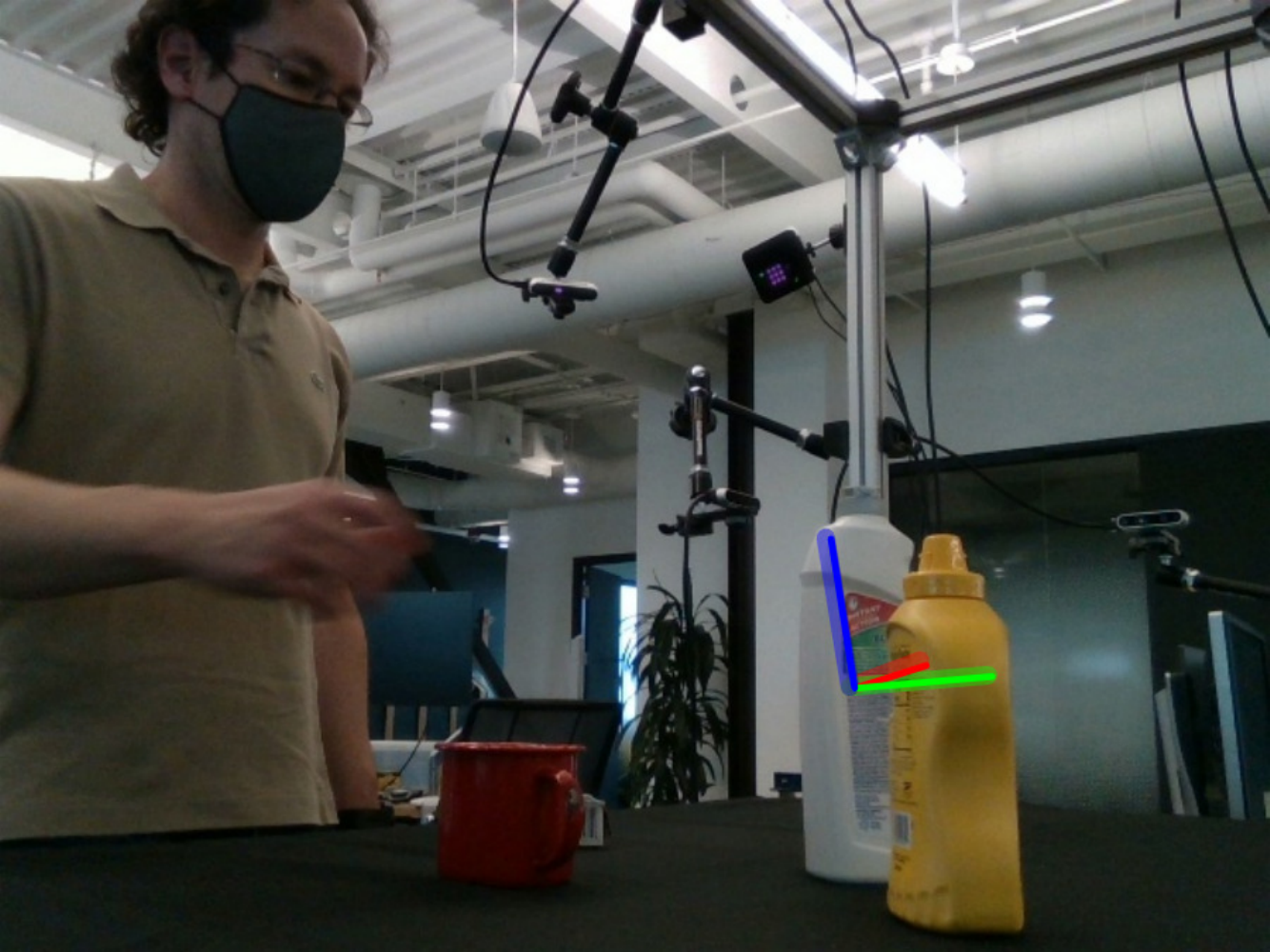}
        \vspace{0.02cm}

        \randomgridrow{MegaPose}{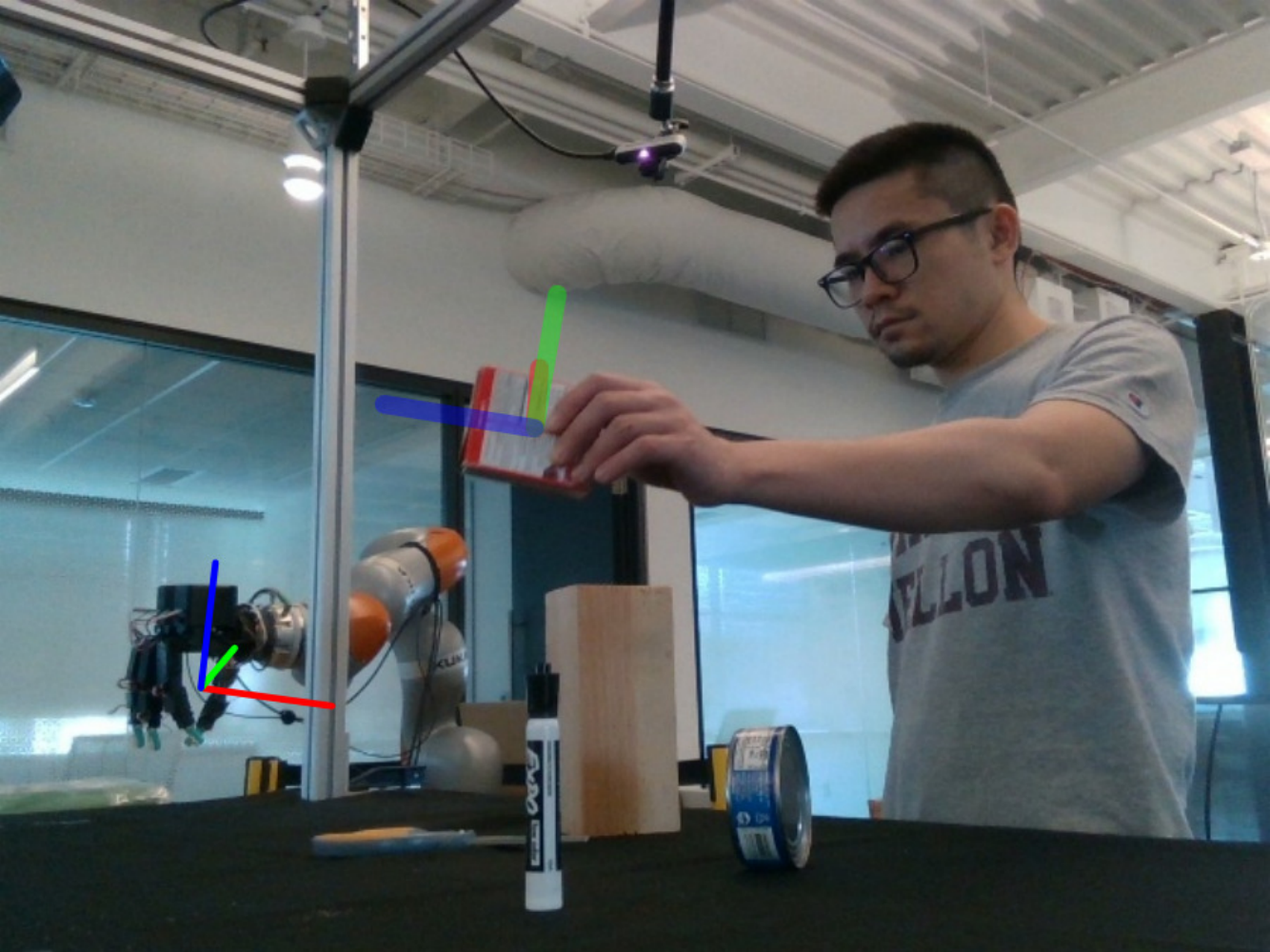}
        \vspace{0.02cm}

        \randomgridrow{FoundPose}{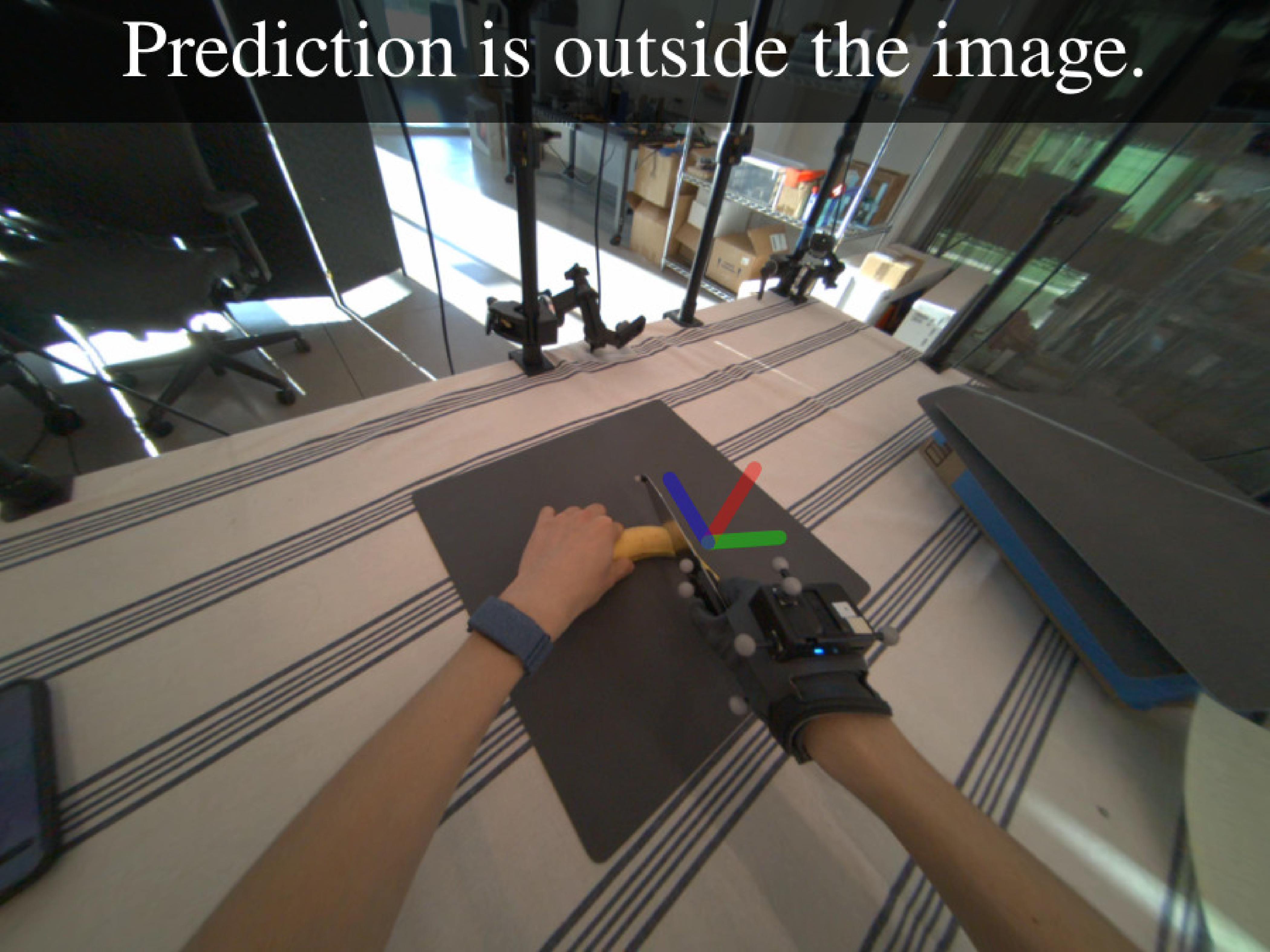}
        \vspace{0.02cm}

        \randomgridrow{FreePose}{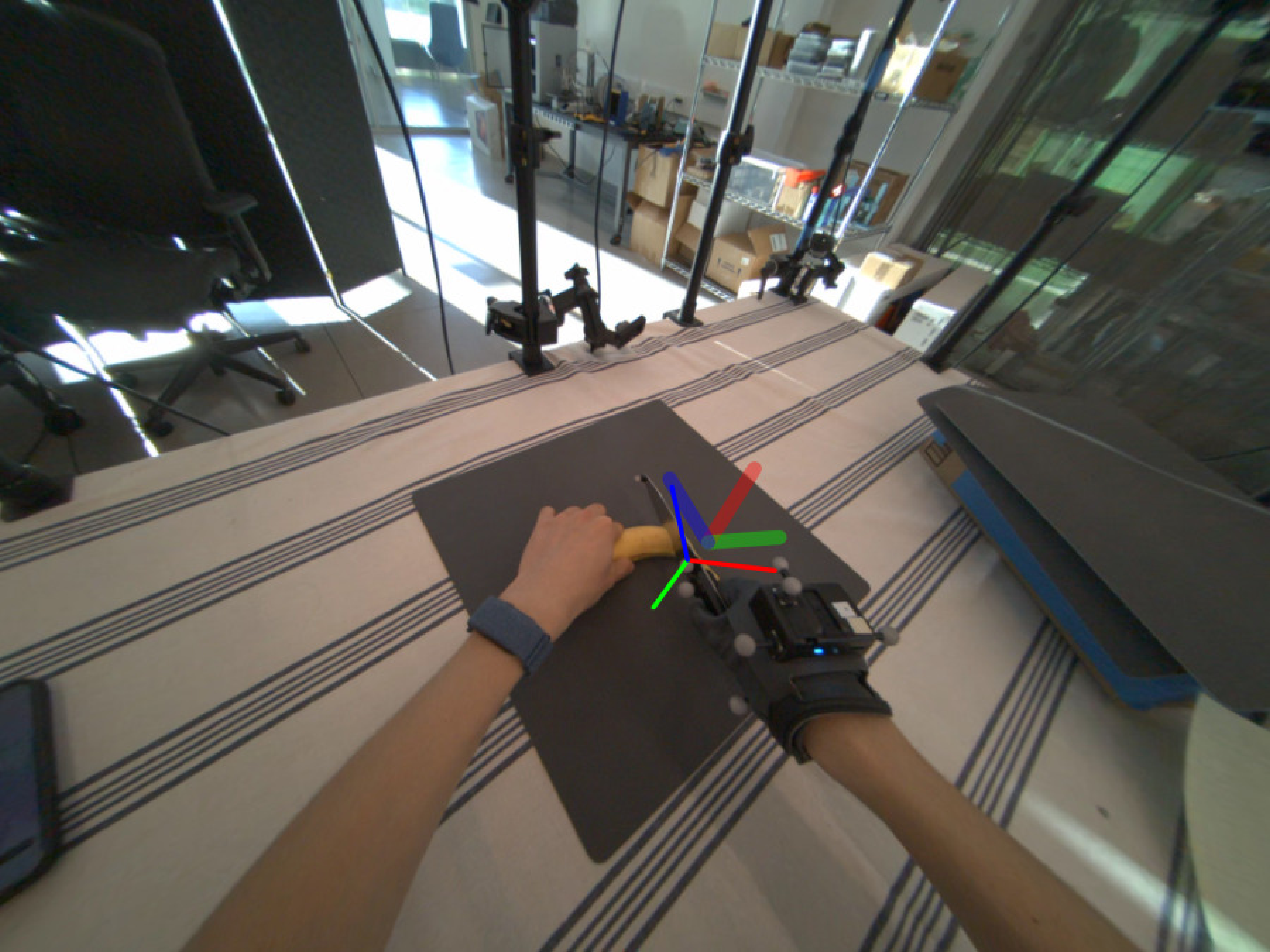}
        \vspace{0.02cm}

        \randomgridrow{UniHOPE}{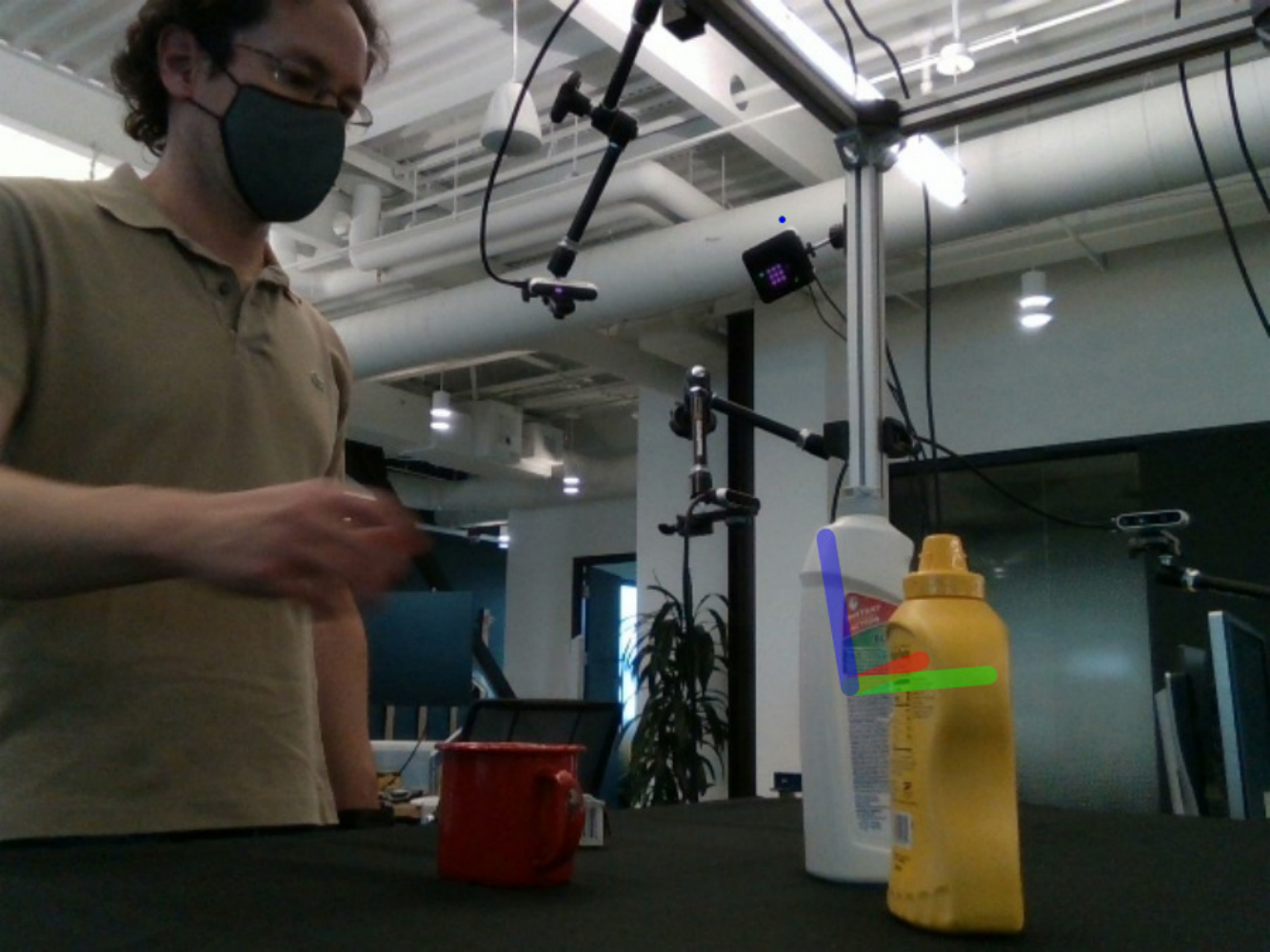}
        \vspace{0.02cm}

        \caption{Qualitative results on randomly sampled frames from video sequences in DexYCB, HO3Dv3, and MoCap Tools.}
        \label{fig:sup-random-qualitative-grid}

    \end{minipage}
\end{figure}

\clearpage
\onecolumn

\section{Robustness to Input Variations}
% We evaluate the robustness of our method to variations in input sources while keeping all other settings fixed. Specifically, we investigate the effects of different hand segmentation methods and reconstructed versus ground-truth object meshes. 

\subsection{Hand Segmentation Methods}
\label{sec:sup_robustness}
\vspace{0.5em}
\begin{table}[htbp]
\renewcommand{\arraystretch}{0.95}
\begin{adjustbox}{max width=\textwidth}
\begin{tabular}{@{}ll*{6}{c}@{\hspace{6pt}}*{6}{c}@{}}
\toprule
& & \multicolumn{6}{c}{\textbf{Consecutive ($i$ to $i+1$)}}
  & \multicolumn{6}{c}{\textbf{Initial-to-subsequent ($0$ to $i$)}} \\
\cmidrule(lr){3-8}\cmidrule(lr){9-14}
& & \multicolumn{2}{c}{Mean ($\downarrow$)}
  & \multicolumn{2}{c}{P90 ($\downarrow$)}
  & \multicolumn{2}{c}{AUC ($\uparrow$)}
  & \multicolumn{2}{c}{Mean ($\downarrow$)}
  & \multicolumn{2}{c}{P90 ($\downarrow$)}
  & \multicolumn{2}{c}{AUC ($\uparrow$)} \\
\cmidrule(lr){3-4}\cmidrule(lr){5-6}\cmidrule(lr){7-8}
\cmidrule(lr){9-10}\cmidrule(lr){11-12}\cmidrule(lr){13-14}
Dataset & Method
  & RRE & RTE & RRE & RTE & ADD & ADD-S
  & RRE & RTE & RRE & RTE & ADD & ADD-S \\
\midrule

DexYCB & FP (pretrained)
  & \posecell{000}{3.07} & \posecell{0000}{8.3}
  & \posecell{000}{7.24} & \posecell{0000}{24.1}
  & \posecell{00}{91.31} & \posecell{00}{95.12}
  & \posecell{000}{25.17} & \posecell{0000}{77.6}
  & \posecell{000}{90.24} & \posecell{0000}{327.8}
  & \posecell{00}{68.97} & \posecell{00}{74.67} \\
& FP (fine-tuned)
  & \posecell{000}{2.24} & \posecell{0000}{6.0}
  & \posecell{000}{5.26} & \posecell{0000}{17.1}
  & \posecell{00}{93.60} & \posecell{00}{96.32}
  & \posecell{000}{22.68} & \posecell{0000}{58.5}
  & \posecell{000}{80.95} & \posecell{0000}{200.1}
  & \posecell{00}{72.54} & \posecell{00}{79.73} \\
\cdashline{2-14}[3pt/2pt]
\noalign{\vskip 0.5ex}
& Ours (train, eval: SAM 3)
  & \posecell[\textbf]{000}{2.19} & \posecell[\underline]{0000}{4.0}
  & \posecell[\underline]{000}{4.96} & \posecell[\underline]{0000}{8.3} 
  & \posecell{00}{95.54} & \posecell[\underline]{00}{97.24}
  & \posecell{000}{18.88} & \posecell{0000}{30.0}
  & \posecell{000}{57.36} & \posecell{0000}{43.6} 
  & \posecell{00}{79.23} & \posecell{00}{88.26}\\
& Ours (train: SAM 3, eval: GDINO+SAM 2)
  & \posecell[\underline]{000}{2.20} & \posecell[\textbf]{0000}{3.9}
  & \posecell[\underline]{000}{4.96} & \posecell[\textbf]{0000}{8.2} 
  & \posecell[\underline]{00}{95.58} & \posecell[\textbf]{00}{97.26}
  & \posecell[\textbf]{000}{18.69} & \posecell[\underline]{0000}{28.6}
  & \posecell[\textbf]{000}{56.32} & \posecell{0000}{42.7} 
  & \posecell[\underline]{00}{79.38} & \posecell[\underline]{00}{88.46}\\
& Ours (train, eval: GDINO+SAM 2)
  & \posecell[\underline]{000}{2.20} & \posecell[\textbf]{0000}{3.9}
  & \posecell{000}{4.97} & \posecell[\underline]{0000}{8.3}
  & \posecell[\textbf]{00}{95.59} & \posecell[\textbf]{00}{97.26}
  & \posecell[\underline]{000}{18.78} & \posecell[\textbf]{0000}{28.5}
  & \posecell[\underline]{000}{56.81} & \posecell[\textbf]{0000}{40.6}
  & \posecell[\textbf]{00}{79.43} & \posecell[\textbf]{00}{88.47} \\

& Ours (train: GDINO+SAM 2, eval: SAM 3)
  & \posecell[\underline]{000}{2.20} & \posecell[\underline]{0000}{4.0}
  & \posecell[\textbf]{000}{4.94} & \posecell[\underline]{0000}{8.3}
  & \posecell{00}{95.55} & \posecell[\underline]{00}{97.24}
  & \posecell[\underline]{000}{18.78} & \posecell{0000}{29.6}
  & \posecell{000}{57.77} & \posecell[\underline]{0000}{41.6}
  & \posecell{00}{79.37} & \posecell{00}{88.38} \\
\midrule

HO3Dv3 & FP (pretrained)
  & \posecell{000}{1.25} & \posecell[\underline]{0000}{1.4}
  & \posecell{000}{2.43} & \posecell{0000}{2.0}
  & \posecell{00}{98.18} & \posecell{00}{98.65}
  & \posecell[\textbf]{000}{16.55} & \posecell{0000}{14.1}
  & \posecell{000}{86.40} & \posecell{0000}{27.9}
  & \posecell{00}{82.24} & \posecell{00}{91.01} \\
& FP (fine-tuned)
  & \posecell[\textbf]{000}{0.94} & \posecell{0000}{10.7}
  & \posecell{000}{1.93} & \posecell{0000}{2.0}
  & \posecell{00}{93.57} & \posecell{00}{94.71}
  & \posecell{000}{18.28} & \posecell{0000}{128.5}
  & \posecell{000}{82.28} & \posecell{0000}{26.3}
  & \posecell{00}{80.39} & \posecell{00}{88.18} \\
\cdashline{2-14}[3pt/2pt] \noalign{\vskip 0.5ex}
& Ours (train, eval: SAM 3)
  & \posecell{000}{0.99} & \posecell[\textbf]{0000}{1.0}
  & \posecell{000}{1.91} & \posecell[\underline]{0000}{1.8} 
  & \posecell[\underline]{00}{98.63} & \posecell{00}{98.85}
  & \posecell{000}{18.17} & \posecell[\textbf]{0000}{10.0}
  & \posecell[\underline]{000}{79.39} & \posecell[\underline]{0000}{23.9} 
  & \posecell[\textbf]{00}{82.74} & \posecell[\underline]{00}{93.92} \\
& Ours (train: SAM 3, eval: GDINO+SAM 2)
  & \posecell{000}{0.99} & \posecell[\textbf]{0000}{1.0}
  & \posecell[\textbf]{000}{1.85} & \posecell[\textbf]{0000}{1.7} 
  & \posecell[\textbf]{00}{98.66} & \posecell[\textbf]{00}{98.87}
  & \posecell{000}{18.93} & \posecell{0000}{10.3}
  & \posecell[\textbf]{000}{79.01} & \posecell{0000}{24.6} 
  & \posecell[\underline]{00}{82.63} & \posecell[\textbf]{00}{93.94} \\
& Ours (train, eval: GDINO+SAM 2)
  & \posecell{000}{0.99} & \posecell[\textbf]{0000}{1.0}
  & \posecell[\underline]{000}{1.87} & \posecell[\underline]{0000}{1.8}
  & \posecell[\underline]{00}{98.63} & \posecell{00}{98.85}
  & \posecell{000}{18.52} & \posecell[\underline]{0000}{10.1}
  & \posecell{000}{80.20} & \posecell[\textbf]{0000}{23.7}
  & \posecell{00}{82.56} & \posecell{00}{93.91} \\

& Ours (train: GDINO+SAM 2, eval: SAM 3)
  & \posecell[\underline]{000}{0.98} & \posecell[\textbf]{0000}{1.0}
  & \posecell[\textbf]{000}{1.85} & \posecell[\underline]{0000}{1.8}
  & \posecell[\textbf]{00}{98.66} & \posecell[\underline]{00}{98.86}
  & \posecell[\underline]{000}{17.38} & \posecell{0000}{10.6}
  & \posecell{000}{80.61} & \posecell{0000}{25.9}
  & \posecell{00}{82.57} & \posecell{00}{93.78} \\
\midrule

MoCap Tools & FP (pretrained)
  & \posecell{000}{3.68} & \posecell{0000}{5.8}
  & \posecell{000}{7.83} & \posecell{0000}{14.6} & -- & --
  & \posecell{000}{44.26} & \posecell{0000}{55.0}
  & \posecell{000}{149.98} & \posecell{0000}{176.6} & -- & -- \\
& FP (fine-tuned)
  & \posecell{000}{2.58} & \posecell{0000}{3.7}
  & \posecell{000}{5.67} & \posecell{0000}{7.8} & -- & --
  & \posecell{000}{42.04} & \posecell{0000}{32.5}
  & \posecell{000}{147.48} & \posecell{0000}{101.8} & -- & -- \\
\cdashline{2-14}[3pt/2pt] \noalign{\vskip 0.5ex}
& Ours (train, eval: SAM 3)
  & \posecell[\textbf]{000}{2.34} & \posecell[\underline]{0000}{3.3}
  & \posecell[\textbf]{000}{4.94} & \posecell[\underline]{0000}{6.7} 
  & -- & --
  & \posecell[\underline]{000}{39.88} & \posecell[\underline]{0000}{29.4}
  & \posecell[\underline]{000}{133.98} & \posecell{0000}{84.9} 
  & -- & -- \\
& Ours (train: SAM 3, eval: GDINO+SAM 2)
  & \posecell{000}{2.44} & \posecell{0000}{3.4}
  & \posecell{000}{5.13} & \posecell{0000}{6.9} 
  & -- & --
  & \posecell{000}{43.40} & \posecell{0000}{30.3}
  & \posecell{000}{143.02} & \posecell{0000}{89.0} 
  & -- & -- \\
& Ours (train, eval: GDINO+SAM 2)
  & \posecell[\underline]{000}{2.38} & \posecell[\textbf]{0000}{3.2}
  & \posecell{000}{5.07} & \posecell[\textbf]{0000}{6.6} & -- & --
  & \posecell[\textbf]{000}{31.56} & \posecell[\textbf]{0000}{27.6}
  & \posecell[\textbf]{000}{82.77} & \posecell[\textbf]{0000}{80.9} & -- & -- \\

& Ours (train: GDINO+SAM 2, eval: SAM 3)
  & \posecell[\underline]{000}{2.38} & \posecell{0000}{3.4}
  & \posecell[\underline]{000}{4.98} & \posecell{0000}{7.1} & -- & --
  & \posecell{000}{48.35} & \posecell{0000}{29.8}
  & \posecell{000}{144.93} & \posecell[\underline]{0000}{84.4} & -- & -- \\
\bottomrule
\end{tabular}
\end{adjustbox}
\caption{We evaluate four combinations of training and evaluation hand masks generated by Grounding DINO+SAM 2 and SAM 3. For each dataset, results above the dashed lines are reproduced from Table~\ref{tab:pose_estimation}, while results below the dashed lines report the four configurations of our method. Our method achieves comparable performance across segmentation configurations and outperforms both FoundationPose variants on most metrics. FP denotes FoundationPose, and GDINO denotes Grounding DINO. Bold and underlined values indicate best and second-best performance, respectively.}
\label{tab:sup_hand_segmentation_methods}
\end{table}
To assess sensitivity to hand segmentation, we train an additional model using hand masks generated by SAM 3~\cite{carion2025sam3segmentconcepts}, instead of Grounding DINO + SAM 2, while keeping all other settings unchanged. We also generate evaluation masks using SAM 3, yielding four training--evaluation combinations. Table~\ref{tab:sup_hand_segmentation_methods} reports these results below the dashed lines for each dataset, alongside the corresponding baseline results reproduced from Table~\ref{tab:pose_estimation} above the dashed lines for comparison. 
\par\vspace{0.5em}\noindent
Our method achieves comparable performance across segmentation configurations and outperforms both pretrained and fine-tuned FoundationPose on most metrics. The exceptions are in mean RRE on HO3Dv3 under both temporal evaluation settings and on MoCap Tools in the initial-to-subsequent setting. These findings are overall consistent with the main results in Table~\ref{tab:pose_estimation}, indicating that our method remains effective across different hand segmentation pipelines.

\clearpage
\onecolumn

\subsection{Object Mesh Quality}
\label{sec:sup_hand_mask_segmentation_methods}

\begin{table}[htbp]
\centering
\footnotesize
\setlength{\tabcolsep}{2pt}
\renewcommand{\arraystretch}{0.95}
\begin{adjustbox}{max width=\textwidth}
\begin{tabular}{@{}ll*{6}{c}@{\hspace{6pt}}*{6}{c}@{}}
\toprule
& & \multicolumn{6}{c}{\textbf{Consecutive ($i$ to $i+1$)}}
  & \multicolumn{6}{c}{\textbf{Initial-to-subsequent ($0$ to $i$)}} \\
\cmidrule(lr){3-8}\cmidrule(lr){9-14}
& & \multicolumn{2}{c}{Mean ($\downarrow$)}
  & \multicolumn{2}{c}{P90 ($\downarrow$)}
  & \multicolumn{2}{c}{AUC ($\uparrow$)}
  & \multicolumn{2}{c}{Mean ($\downarrow$)}
  & \multicolumn{2}{c}{P90 ($\downarrow$)}
  & \multicolumn{2}{c}{AUC ($\uparrow$)} \\
\cmidrule(lr){3-4}\cmidrule(lr){5-6}\cmidrule(lr){7-8}
\cmidrule(lr){9-10}\cmidrule(lr){11-12}\cmidrule(lr){13-14}
Dataset & Method
  & RRE & RTE & RRE & RTE & ADD & ADD-S
  & RRE & RTE & RRE & RTE & ADD & ADD-S \\
\midrule

DexYCB & MegaPose
  & \posecell{000}{5.81} & \posecell{0000}{21.6}
  & \posecell{000}{12.55} & \posecell{0000}{32.2}
  & \posecell{00}{87.96} & \posecell{00}{92.36}
  & \posecell{000}{30.36} & \posecell{0000}{97.2}
  & \posecell{000}{123.36} & \posecell{0000}{277.3}
  & \posecell{00}{70.31} & \posecell{00}{78.65} \\
& FP (pretrained)
  & \posecell{000}{1.76} & \posecell{0000}{4.0}
  & \posecell{000}{4.02} & \posecell[\underline]{0000}{8.9}
  & \posecell{00}{95.70} & \posecell{00}{97.38}
  & \posecell[\underline]{000}{11.87} & \posecell[\underline]{0000}{38.5}
  & \posecell[\underline]{000}{26.33} & \posecell[\underline]{0000}{73.2}
  & \posecell[\underline]{00}{83.24} & \posecell[\underline]{00}{86.63} \\
& FP (fine-tuned)
  & \posecell[\underline]{000}{1.60} & \posecell[\underline]{0000}{3.9}
  & \posecell[\underline]{000}{3.88} & \posecell{0000}{9.5}
  & \posecell[\underline]{00}{95.75} & \posecell[\underline]{00}{97.44}
  & \posecell{000}{13.16} & \posecell{0000}{39.8}
  & \posecell{000}{32.12} & \posecell{0000}{76.2}
  & \posecell{00}{81.22} & \posecell{00}{86.14} \\
& FoundPose
  & \posecell{000}{55.86} & \posecell{0000}{149.1}
  & \posecell{000}{171.52} & \posecell{0000}{369.7}
  & \posecell{00}{36.28} & \posecell{00}{57.71}
  & \posecell{000}{64.63} & \posecell{0000}{193.6}
  & \posecell{000}{173.45} & \posecell{0000}{344.5}
  & \posecell{00}{31.64} & \posecell{00}{56.34} \\
& FreePose
  & \posecell{000}{2.04} & \posecell{0000}{21.3}
  & \posecell{000}{5.55} & \posecell{0000}{37.1}
  & \posecell{00}{85.85} & \posecell{00}{91.07}
  & \posecell{000}{29.53} & \posecell{0000}{141.3}
  & \posecell{000}{93.31} & \posecell{0000}{306.2}
  & \posecell{00}{44.50} & \posecell{00}{61.02} \\
& UniHOPE
  & \posecell{000}{53.08} & \posecell{0000}{367.2}
  & \posecell{000}{149.95} & \posecell{0000}{1000.0}
  & \posecell{00}{23.34} & \posecell{00}{30.63}
  & \posecell{000}{108.43} & \posecell{0000}{775.0}
  & \posecell{000}{170.76} & \posecell{0000}{2000.0}
  & \posecell{00}{9.17} & \posecell{00}{11.41} \\
\rowcolor{oursblue}
& Ours
  & \posecell[\textbf]{000}{1.54} & \posecell[\textbf]{0000}{3.0}
  & \posecell[\textbf]{000}{3.58} & \posecell[\textbf]{0000}{5.7}
  & \posecell[\textbf]{00}{96.70} & \posecell[\textbf]{00}{97.90}
  & \posecell[\textbf]{000}{10.66} & \posecell[\textbf]{0000}{24.8}
  & \posecell[\textbf]{000}{17.67} & \posecell[\textbf]{0000}{27.3}
  & \posecell[\textbf]{00}{85.24} & \posecell[\textbf]{00}{90.62} \\
\bottomrule
\end{tabular}
\end{adjustbox}

\caption{We evaluate object pose estimation and tracking performance on DexYCB using ground-truth meshes. Our method outperforms all baselines under both temporal evaluation settings, using either ground-truth or BundleSDF-reconstructed meshes (Table~\ref{tab:pose_estimation}).}
\label{tab:sup_groundtruth_meshes}
\end{table}

% unihope numbers reused because it uses pretrained weights by dataset
% freepose has two rows - retrieved mesh and gt mesh
% this shows that our method can perform well consistently, no matter the quality of mesh inputs (given either 3d reconstructed meshes or gt meshes)
To assess sensitivity to object mesh quality, we also evaluate the methods using ground-truth meshes instead of reconstructed meshes, while keeping all other settings unchanged. In Table~\ref{tab:sup_groundtruth_meshes}, we reproduce the UniHOPE results from Table~\ref{tab:pose_estimation}, because the UniHOPE checkpoints are already trained on ground-truth object meshes. For FreePose, we provide the ground-truth meshes directly rather than retrieving the closest matches from a mesh bank. 
\par\vspace{0.5em}\noindent
Our method outperforms all baselines across both temporal settings on DexYCB, using either reconstructed or ground-truth meshes. These results indicate that our method remains effective across different levels of object mesh quality.

\clearpage
\onecolumn

\section{MoCap Tools Dataset}
\label{sec:sup_mocap_tools}
In this section, we provide additional details about the MoCap Tools dataset, including data collection and dataset statistics.  

\subsection{Data Collection}
\label{sec:sup_data_collection}
\begin{figure}[H]
    \centering
    \begin{minipage}{0.75\textwidth}
        \centering
        \includegraphics[width=\linewidth]{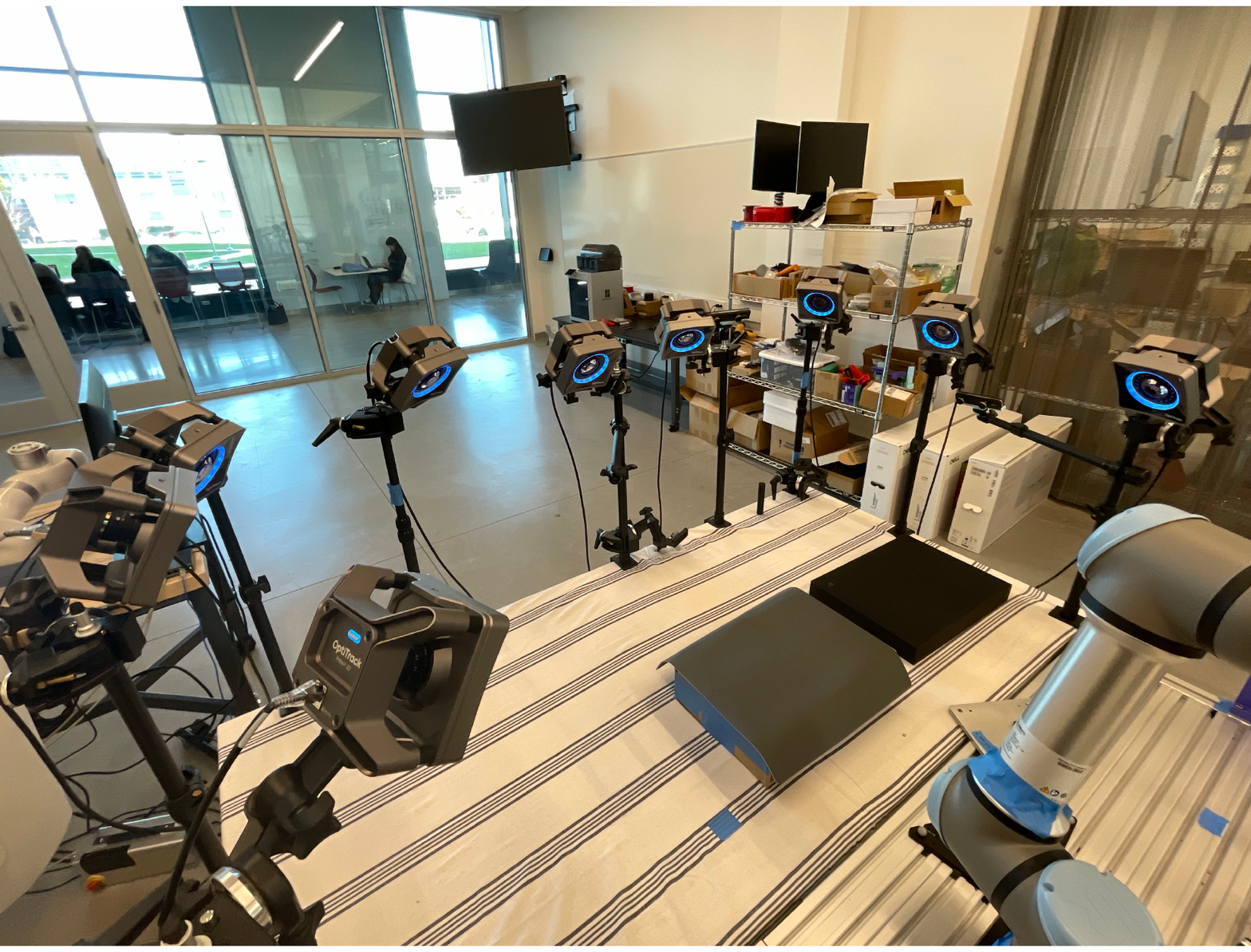}
        \captionsetup{justification=raggedright,singlelinecheck=false}
        \caption{OptiTrack setup.}
        \label{fig:sup_optitrack}
    \end{minipage}
\end{figure}

We capture egocentric tool-use sequences using Meta Project Aria Gen~2 glasses and an OptiTrack motion capture (MoCap) system. The OptiTrack system consists of nine synchronized cameras (Figure~\ref{fig:sup_optitrack}) operating at 60\,Hz, providing ground-truth 6-DoF poses of the manipulated tools and headset. We attach three to five rigid markers to each of the manipulated objects and the headset for tracking. The Aria glasses record $2016\times1512$ fisheye RGB images at 30\,fps, which we rectify to perspective images at the same resolution using Aria's factory-calibrated camera model.

\subsection{Data Statistics}
\label{sec:sup_data_statistics}
The dataset contains 11 manipulation sequences involving 11 everyday tools: brush, fork, hammer, knife, marker, mug, peeler, scsissors, spatula, stapler, and tongs. These tools interact with six target objects: whiteboard, cardboard, plastic block, banana, apple, and cup. Each sequence lasts approximately 15 seconds, containing 434--447 RGB frames at 30\,fps. The dataset contains a total of 4,875 RGB-D frames with ground-truth 6-DoF tool and target object poses.

\end{document}